\documentclass[preprint,5p,times,twocolumn]{elsarticle}

\usepackage{amssymb}
\usepackage{amsmath}

\usepackage{amsmath,amsfonts}
\usepackage{algorithmic}
\usepackage{algorithm}
\usepackage{adjustbox} 
\usepackage{array}
\usepackage{textcomp}
\usepackage{stfloats}
\usepackage{url}
\usepackage{verbatim}
\usepackage{graphicx}
\usepackage{amssymb}
\usepackage{booktabs}
\graphicspath{{./figs/}}
\graphicspath{{./authors/}}
\DeclareGraphicsExtensions{.pdf,.jpg,.png}
\usepackage{bbm}
\usepackage{subfig}
\usepackage{multirow} 
\usepackage{threeparttable}
\usepackage{makecell}
\usepackage[table]{xcolor} 
\usepackage{mathrsfs}
\usepackage{diagbox}
\usepackage{arydshln} 

\newcommand{\myPara}[1]{\vspace{0.05in}\noindent\textbf{#1}}
\newcommand{\tabincell}[2]{\begin{tabular}{@{}#1@{}}#2\end{tabular}}

\def\ie{{\em i.e.}}
\def\eg{{\em e.g.}}
\def\etal{{\em et al.}}
\def\etc{{\em etc}}

\makeatletter

\newcommand{\Rmnum}[1]{\expandafter\@slowromancap\romannumeral #1@}
\makeatother

\usepackage{booktabs}
\usepackage{amssymb}
\usepackage{bbding}
\usepackage{pifont}
\usepackage{wasysym}
\usepackage{utfsym}
\usepackage{wrapfig}  
\usepackage{circledsteps}
\definecolor{darkgreen}{rgb}{0.0, 0.5, 0.0}  

\newcommand{\bio}[2]{
    \noindent 
    \begin{minipage}[t]{0.12\textwidth} 
        \vspace{0pt} 
        \includegraphics[width=\linewidth]{#1} 
    \end{minipage}%
    \hfill 
    \begin{minipage}[t]{0.35\textwidth} 
        \vspace{0pt} 
        #2 
    \end{minipage}
    \vspace{2em} 
}

\usepackage{hyperref} 

\usepackage{xcolor} 
\definecolor{ifblue}{RGB}{0, 114, 189} 
\hypersetup{
    colorlinks=true,
    linkcolor=ifblue!80,     
    citecolor=ifblue!60,     
    urlcolor=ifblue,         
    pdfborder={0 0 0}
}

\journal{Elsevier}

\begin{document}

\begin{frontmatter}



\title{COST: Contrastive One-Stage Transformer for Vision-Language Small Object Tracking}


\author[1,2,3]{{Chunhui Zhang}}

\affiliation[1]{organization={Cooperative Medianet Innovation Center, Shanghai Jiao Tong University},
            city={Shanghai},
            postcode={200240}, 
            country={China}}

\author[2]{{Li Liu}}
\cortext[1]{Corresponding author: Li Liu.}
\ead{avrillliu@hkust-gz.edu.cn}
\affiliation[2]{organization={The Hong Kong University of Science and Technology (Guangzhou)},
            city={Guangzhou},
            postcode={511458}, 
            country={China}}

\author[1]{{Jialin Gao}}
\affiliation[3]{organization={CloudWalk Technology Co., Ltd},
            city={Shanghai},
            postcode={201203}, 
            country={China}}

\author[1]{{Xin Sun}}
\affiliation[4]{organization={Institute of Information Engineering, Chinese Academy of Sciences},
            city={Beijing},
            postcode={100085}, 
            country={China}}

\affiliation[5]{organization={Shanghai Artificial Intelligence Laboratory},
            city={Shanghai},
            postcode={200032}, 
            country={China}}

\author[3]{{Hao Wen}}

\author[3]{{Xi Zhou}}

\author[4]{{Shiming Ge}}

\author[1,5]{{Yanfeng Wang}}

\begin{abstract} Transformer has recently demonstrated great potential in improving vision-language (VL) tracking algorithms. However, most of the existing VL trackers rely on carefully designed mechanisms to perform the multi-stage multi-modal fusion. Additionally, direct multi-modal fusion without alignment ignores distribution discrepancy between modalities in feature space, potentially leading to suboptimal representations. In this work, we propose COST, a contrastive one-stage transformer fusion framework for VL tracking, aiming to learn semantically consistent and unified VL representations. Specifically, we introduce a contrastive alignment strategy that maximizes mutual information (MI) between a video and its corresponding language description. This enables effective cross-modal alignment, yielding semantically consistent features in the representation space. By leveraging a visual-linguistic transformer, we establish an efficient multi-modal fusion and reasoning mechanism, empirically demonstrating that a simple stack of transformer encoders effectively enables unified VL representations. Moreover, we contribute a newly collected VL tracking benchmark dataset for small object tracking, named VL-SOT500, with bounding boxes and language descriptions. Our dataset comprises two challenging subsets, VL-SOT230 and VL-SOT270, dedicated to evaluating \emph{generic} and \emph{high-speed} small object tracking, respectively. Small object tracking is notoriously challenging due to weak appearance and limited features, and this dataset is, to the best of our knowledge, the first to explore the usage of language cues to enhance visual representation for small object tracking. Extensive experiments demonstrate that COST achieves state-of-the-art performance on five existing VL tracking datasets, as well as on our proposed VL-SOT500 dataset. Source codes and dataset will be made publicly available at \href{https://github.com/983632847/Awesome-Multimodal-Object-Tracking}{\color{magenta}{here}}.

\end{abstract}







\end{frontmatter}


\section{Introduction}
\label{sec:intro}
\begin{figure}[ht]
  \centering
  \includegraphics[width=1.0\linewidth]{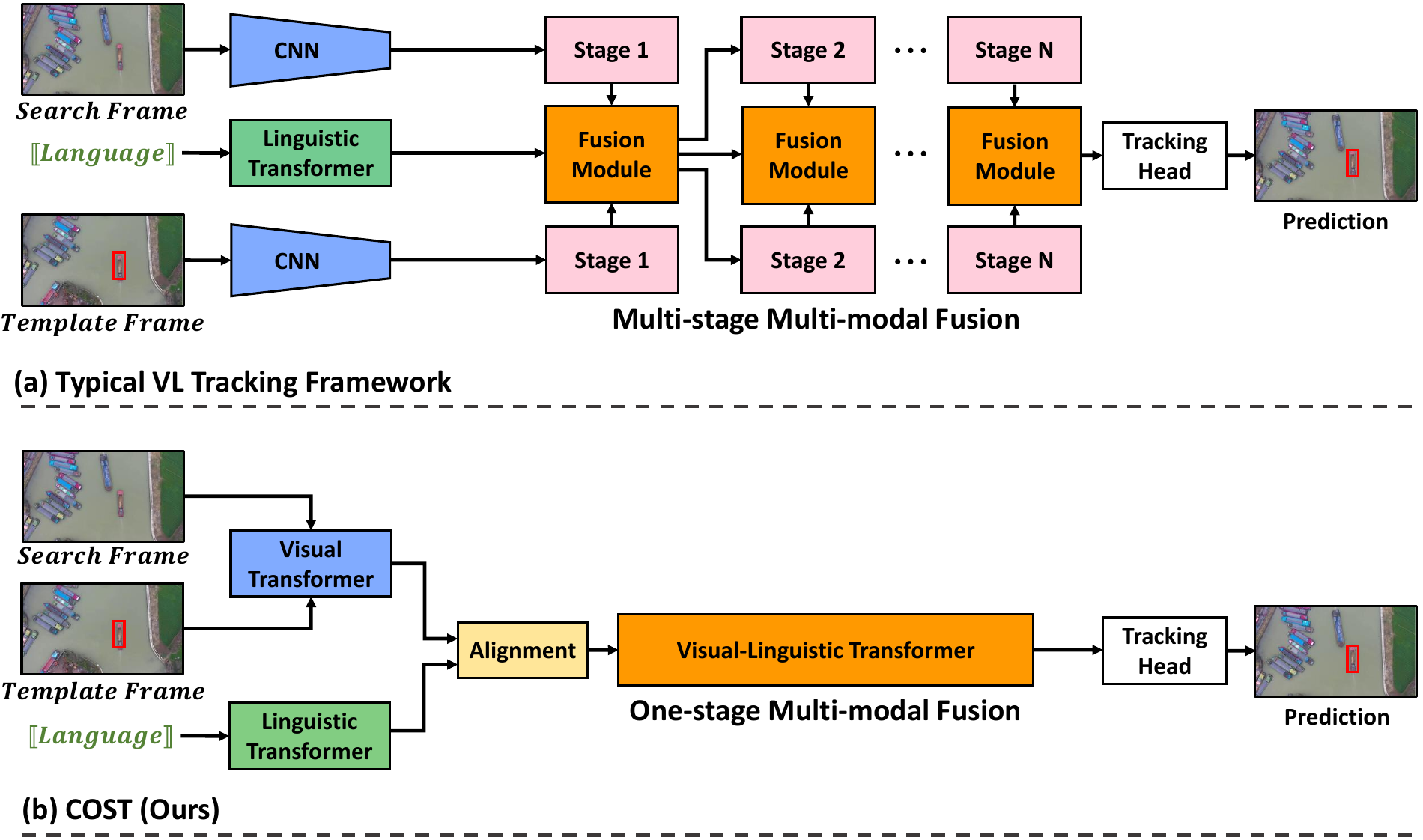}
  \caption{Comparison of VL tracking pipelines. (a) The typical VL tracking framework aggregates CNN and Transformer features heterogeneously using \emph{multi-stage multi-modal fusion}. (b) Our COST performs \emph{one-stage multi-modal fusion} with a contrastive transformer fusion framework in a homogeneous way and predicts the object location by a tracking head.}
   \label{fig:Motivation}
\end{figure}

Vision-language (VL) tracking refers to the task of sequentially locating a moving object in a video sequence based on an initial bounding box and a language description~\citep{li2017tracking,wang2021towards,feng2021siamese,fan2019lasot}. This is one of the fundamental yet open problems in computer vision (CV), and it has a wide range of real applications, such as transportation surveillance~\citep{wang2021towards}, aerial photography~\citep{zhang2022webuav}, and intelligent agriculture~\citep{li2017tracking}. In the past decade, the two dominating tracking paradigms are Siamese networks~\citep{FENG2024102492,zhang2020ocean,guo2020siamcar} and deep discriminative correlation filters~\citep{xu2019learning,danelljan2019atom,danelljan2020probabilistic,ge2020CD,zhang2020accurate,ge2020cascaded}. Inspired by the huge success of the transformer~\citep{vaswani2017attention} in various vision and language tasks, there is a surging interest in exploring transformer-based trackers~\citep{FENG2024102492,guo2022divert,chen2021transformer,wang2021transformer,WANG2025102940}. Nevertheless, existing VL trackers~\citep{feng2021siamese,wang2021towards} heavily rely on a highly customized and meticulously designed multi-stage multi-modal fusion module to heterogeneously model interactions between visual features (\textit{e.g.}, extracted by convolutional neural network (CNN)~\citep{he2016deep}) and language features (\textit{e.g.}, extracted by linguistic transformer~\citep{devlin2018bert}) as shown in Fig.~\ref{fig:Motivation}(a).

Existing works demonstrate that the core problems of VL tracking are \emph{multi-modal fusion and reasoning}~\citep{li2017tracking,feng2021siamese,LIU2025102941,ma2021capsule,zhang2024awesome,zhang2024webuot,CAI2023101816}. Mainstream VL trackers attempt to explore adaptive interactions of multi-modality, where the key insight is to apply a \emph{carefully-designed} fusion encoder to perform multi-stage multi-modal fusion to learn joint representations~\citep{feng2021siamese,guo2022divert}. In~\citep{feng2021siamese}, a dynamic aggregation module was proposed to combine predictions from both visual and language modalities based on the entropy of predictions. The recent tracker VLT\_TT~\citep{guo2022divert} was proposed to learn VL representations with a ModaMixer from shallow to deep layers of the asymmetrical ConvNet. Despite their advanced performance, these \emph{highly customized} multi-stage multi-modal fusion methods suffer from the problem that vision and language modalities have huge distribution discrepancies in the feature space (\textcolor{black}{\ie, vision is spatial redundancy and semantic sparse, while language is highly semantic and information-dense})~\citep{He2022MAE,guo2024divert}, which leads to significant learning inefficiency in multi-modal fusion. \textbf{\emph{Thus, can we achieve efficient multi-modal fusion and reasoning for VL tracking using a unified one-stage fusion architecture?}}

To answer the above question, we propose a contrastive one-stage multi-modal fusion framework based on the transformer, namely COST, for VL tracking. The core idea is to design a homogeneous contrastive visual-linguistic fusion (CVLF) module, which achieves both cross-modal alignment and relation reasoning simultaneously, that is, learning VL representations in a unified transformer architecture~\citep{vaswani2017attention}. In this way, instead of using carefully designed multi-stage fusion networks (\eg, Siamese natural language region proposal network in~\citep{feng2021siamese}, or ModaMixer in~\citep{guo2022divert}), the visual and language signals are embedded into a shared and unified semantic space with a simple CVLF module. As shown in Fig.~\ref{fig:Motivation}(b), we first feed the video sequence (\ie, search frame and template frame) and language description into visual and linguistic branches. The visual transformer and linguistic transformer are applied in these two branches to model the global cues in vision and language domains, respectively. To handle the huge distribution discrepancy between modalities, we introduce a contrastive alignment (CoA) to pull the embeddings of matched video-language pairs together while pushing those of non-matched pairs apart by maximizing global mutual information (MI)~\citep{oord2018representation} between matched video and language. The CoA forces the learned visual and language features to align well in embedding spaces via contrastive learning (CL)~\citep{yang2022vision}, ensuring the preservation of semantically consistent information. Then, the aligned visual and language features are fused via a visual-linguistic transformer to promote cross-modal relation reasoning. Note that this work primarily focuses on multi-modal fusion, but directly facilitates multi-modal reasoning (\ie, target position estimation) through modalities fusion. Finally, the object's location is predicted by a tracking head. Our VL tracking framework has several appealing advantages: 1) achieving homogeneous one-stage multi-modal fusion; 2) learning representations that are semantically meaningful for cross-modal video-language pairs; 3) replacing complex fusion modules with a simple stack of basic transformer encoders~\citep{vaswani2017attention}.

Recently, several VL tracking datasets~\citep{fan2021lasot,wang2021towards,zhang2022webuav,zhang2024webuot} have been proposed, greatly advancing the development of this field. However, there are still many challenging issues that remain unresolved. For instance, small objects are commonly encountered in many scenarios, \eg, unmanned aerial vehicles (UAVs), sports, remote sensing, and autonomous driving, where the small size of the objects leads to weak appearance and features, posing significant challenges for trackers. To this end, we propose the first multi-modal small object tracking dataset to explore language-enhanced small object tracking, called VL-SOT500. However, one more critical issue is that small objects often move at high speeds~\citep{cheng2023towards}, resulting in severe motion blur in the captured video sequences and abrupt changes in motion direction. Unfortunately, tracking high-speed small objects remains largely unexplored~\citep{zhang2022tracking,zhang2022webuav}, with a lack of publicly available large-scale benchmark datasets. Therefore, we construct VL-SOT500 into two subsets: VL-SOT230 and VL-SOT270, dedicated to evaluating \emph{generic} and \emph{high-speed} small object tracking, respectively.

The main contributions are summarized as follows:
\begin{itemize}
\item We propose a simple yet efficient contrastive one-stage transformer fusion framework for VL tracking to learn feature representations in a homogeneous manner. 

\item We frame cross-modal alignment as a CL problem and achieve the alignment of visual and language features in the feature space by CoA. The CoA delivers a novel explicit cross-modal alignment for VL tracking.

\item We propose VL-SOT500, the first large-scale multi-modal small object tracking dataset with bounding boxes and language descriptions. The dataset includes two challenging subsets, VL-SOT230 and VL-SOT270, designed for developing language-enhanced generic and high-speed small object tracking algorithms.

\item We conduct comprehensive experiments to validate the merits of our method and show significantly improved results on five existing VL tracking benchmarks and the newly proposed VL-SOT500. Through in-depth analysis and discussion, we derive numerous valuable observations and insights in the field of VL tracking.
\end{itemize}

\begin{table*}[t]
\footnotesize
  \centering
  \captionsetup{font={small}}
  \caption{Comparison of VL-SOT500 with existing generic object tracking and small object tracking datasets. VL-SOT500 includes two subsets (\ie, VL-SOT230 and VL-SOT270) designed for benchmarking \emph{generic} and \emph{high-speed} small object tracking, respectively.}
  \footnotesize
  \begin{threeparttable}
  \setlength{\tabcolsep}{2.2mm}{
  \begin{tabular}{lccccccccccc}
    \Xhline{0.75pt}  
    {Datasets}  &  Videos &  Classes & Attributes  & \tabincell{c}{ Min \\ frame} & \tabincell{c}{Mean \\frame} & \tabincell{c}{Max \\frame} &  \tabincell{c}{Total \\frames} &  \tabincell{c}{Average\\ size ($\downarrow$) } &  \tabincell{c}{Average relative \\ speed ($\uparrow$)} & \tabincell{c}{Absent \\ labels} & \tabincell{c}{Language \\descriptions} \\
    
    \midrule

    OTB100~\citep{wu2015otb}  & 100  & 16 &  11 & 71  &  590 &   3,878   &  59 K & 67.6  & 0.440   &  \ding{55}  &   \ding{55}      \\

    VOT2018~\citep{kristan2018sixth}  & 60  & 24 &  5 & 41  &  356 &   1,500  &   21 K & 300.9  & 0.815  &  \ding{55}  &     \ding{55}      \\

    GOT-10k~\citep{huang2019got} & 10,000 & 563 &  6 & 29 &  149 &   1,418    &  1.5 M  & 299.8  & 0.566 & \ding{51} &    \ding{55}      \\

    LaSOT~\citep{fan2019lasot}  & 1,400 & 70 &  14 & 1,000 &  2,506 &  11,397     & 3.52 M  & 179.5  & 0.584  &  \ding{51}  &    \ding{51} \\
    
    TNL2K~\citep{wang2021towards}  & 2,000 & - &  17  & 21  &  622 &   18,488  &  1.24 M & 181.4  & 0.473  &  \ding{51}  &    \ding{51} \\
    \hline
    
    Small90~\citep{liu2019aggregation}  & 90  & 15 &  11 & 34  &  439 &   2,738  &   39.5 K & 37.2  & 0.543  &  \ding{55}  &   \ding{55}    \\

    TSFMO$^{\star}$~\citep{zhang2022tracking}  & 250  & 26 &  12  & 16  &  196 &   887  &   49 K &  22.6  &  -  &  \ding{55}  &   \ding{55}    \\

    LaTOT~\citep{zhu2023tiny}  & 434  & 48 &  12  & 21  &  501 &   4,632  &   217.7 K & 14.0  & 0.700  &  \ding{55}  &   \ding{55}    \\
    
    \hline
    
    {VL-SOT230 (Ours)} &  {230} & {50} &  {17} & {47}  &  {1,002} &   {4,632}  & {230.4 K}  & {13.8}  & {0.755} &  \ding{51} &  \ding{51}  \\
    
    {VL-SOT270 (Ours)} &  {270} & {46} &  {17} & {7}  &  {82} &   {578}  & {22.3  K}  & {14.3}  & {3.930} &  \ding{51} &  \ding{51}  \\

    \textbf{VL-SOT500 (Ours)} &  \textbf{500} & \textbf{84} &  \textbf{17} & \textbf{7}  &  \textbf{505} &   \textbf{4,632}  & \textbf{252.7 K}  & \textbf{14.1}  & \textbf{2.469} &  \ding{52} &  \ding{52}  \\
   \Xhline{0.75pt}  
  \end{tabular}
  }

    \begin{tablenotes}
        \footnotesize
        \item[$\star$] This dataset was not publicly available until the submission of our paper.
    \end{tablenotes}
\end{threeparttable}
\label{tab:datasets_comparison}
\end{table*}

\section{Related Works}
\label{sec:relatedworks}
\subsection{Vision-Language Tracking for Small Object}

Recently, VL tracking has received extensive attention~\citep{feng2021siamese,guo2022divert,wang2021towards,zhang2022webuav}. There are many recent algorithms on this topic, which are not limited to adaptive tracking and grounding switch based on a local-global-search scheme~\citep{wang2021towards}, Siamese natural language region proposal network~\citep{feng2021siamese}, capsule-based tracking network~\citep{ma2021capsule}, dynamic filter generating and attention model~\citep{li2017tracking}, LSTM-based tracking~\citep{feng2020real}, structure-aware local search, global proposal generation~\citep{wang2018describe}, and learning adaptive VL representations with ModaMixer~\citep{guo2022divert}. Some advanced VL tracking methods combine several techniques, \eg, visual grounding~\citep{zhou2023joint} and neural architecture search~\citep{guo2022divert}. In addition, the unified model~\citep{zhang2023all}, sequence-to-sequence model~\citep{zheng2023towards} and the Mamba model~\citep{zhang2024mambatrack,liu2024mambavlt} have also demonstrated significant advantages in VL tracking. However, existing VL tracking methods mainly focus on tracking objects of normal size, overlooking small object tracking~\citep{liu2019aggregation,zhang2022tracking,zhu2023tiny}, which is prevalent in the real world and presents greater challenges, such as extremely low resolution, fast motion, weak visual information, and more noise. To address this gap, we propose the first VL tracking benchmark dedicated to small object tracking and a simple yet effective baseline method.

\textbf{Challenges of small object detection/tracking} include low resolution, poor visibility, susceptibility to occlusion, and difficulty in maintaining accurate localization due to their limited size and noisy feature representation~\citep{cheng2023towards,YANG2025103007}. In image-based detection, information loss is exacerbated by the down-sampling operations in deep neural networks, making it difficult to retain discriminative features~\citep{cheng2023towards}. Moreover, small objects exhibit low tolerance to bounding box perturbations, significantly affecting localization accuracy. For video tracking, additional complexities arise from motion blur, temporal inconsistency, and the need for continuous feature association across frames. Small objects are more vulnerable to occlusions and background noise, further complicating their tracking. Recent studies have proposed solutions such as coarse-to-fine proposal generation to improve localization precision~\citep{yuan2023small}, optimization of the effective receptive field to enhance feature extraction and reduce noise~\citep{kim2023dead}, and ensemble fusion techniques that integrate multi-scale or multi-frame predictions to improve robustness in dynamic scenarios~\citep{hou2023ensemble}. In this work, we propose a VL tracking approach aimed at alleviating various challenges in small object tracking—particularly the insufficiency of visual information caused by small target sizes—from \emph{a novel semantic-enhanced perspective}~\citep{zhang2022webuav,fan2021lasot}. To validate our method, we construct a large-scale small object tracking dataset with language descriptions.

\textbf{Compared to existing trackers}, the proposed method exhibits the following differences: 1) Different from early VL trackers that use a heterogeneous fusion manner (\ie, CNN-Transformer~\citep{wang2021towards,feng2021siamese}), we investigate an efficient and homogeneous fusion manner (\ie, Transformer-Transformer) to enhance cross-modal relation reasoning by learning unified VL representations. In general, features from similar architectures can reduce the gap of multi-modal feature fusion~\citep{baltruvsaitis2018multimodal}. We experimentally verify the advantage of the homogeneous fusion manner in Section~\ref{sec:Further_Discussions}. 2) To the best of our knowledge, most of the existing VL tracking methods ignore visual-linguistic alignment in the feature space. In this work, we suggest using CL for explicit visual-linguistic feature alignment to promote multi-modal fusion and improve tracking performance. 3) Following the spirit of ``align before fusion''~\citep{li2021align,duan2022multi,khan2022single}, we achieve contrastive one-stage multi-modal fusion without using carefully-designed multi-stage multi-modal fusion~\citep{feng2021siamese,guo2022divert} and complex post-processing modules (\eg, temporal modeling module~\citep{zhou2023joint}) for the VL tracking task.

\subsection{Transformer in Vision and Language Tasks}
Transformer is a type of deep neural network mainly based on the self-attention mechanism~\citep{devlin2018bert,vaswani2017attention,dosovitskiy2020image}. Since the pioneering work~\citep{vaswani2017attention}, the transformer has brought significant advances in the field of natural language processing (NLP)~\citep{devlin2018bert,brown2020language,raffel2020exploring,yang2019xlnet}, \eg, BERT~\citep{devlin2018bert} and GPT-3~\citep{brown2020language}. Motivated by the prominent success of transformer in natural language processing tasks, researchers have recently applied transformer to different CV tasks~\citep{dosovitskiy2020image,carion2020end,liu2021swin,chen2020generative,chen2021empirical,ELHARROUSS2025102951}. ViT~\citep{dosovitskiy2020image} and follow-up vision transformer works focus on pixel prediction~\citep{chen2020generative}, set-based prediction and bipartite matching~\citep{carion2020end}, shifted window-based self-attention~\citep{liu2021swin}, deformable attention~\citep{xia2022vision}, self-supervised learning~\citep{chen2021empirical,He2022MAE}, \etc. Besides, researchers also investigate vision-language pre-training~\citep{yang2022vision,li2020oscar}, vision-language navigation~\citep{wang2019reinforced}, visual grounding~\citep{deng2021transvg}, text-to-image generation~\citep{ramesh2021zero,ding2021cogview}, and cross-modal retrieval~\citep{salvador2021revamping}. In this work, we develop a visual-linguistic transformer to enhance relationships between vision and language modalities for VL tracking. The proposed visual-linguistic transformer is the core structure of our one-stage multi-modal fusion framework.

\begin{figure*}[t]
  \centering
  \includegraphics[width=1.0\linewidth]{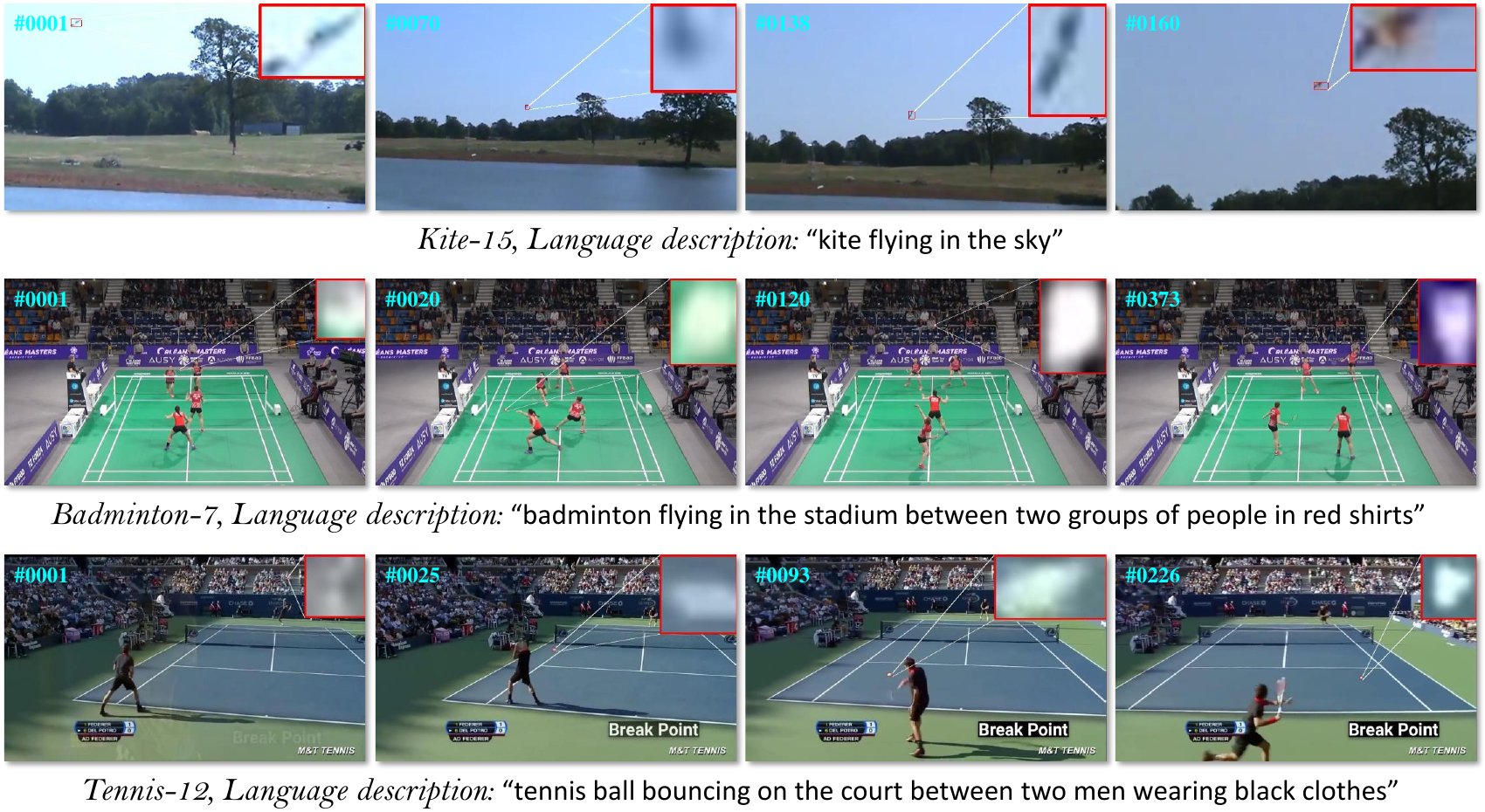} 
  \caption{Some representative samples in the proposed VL-SOT500 dataset. We annotate each video sequence with bounding boxes and a language description. Small objects pose significant challenges to tracking due to less effective visual information, high-speed motion, \etc. Best viewed by zooming in.}
   \label{fig:Example_Sequences}
\end{figure*}

\subsection{Contrastive Learning} 
Self-supervised CL~
\citep{oord2018representation,he2020momentum,chen2020simple,tian2020makes,caron2020unsupervised} has shown striking performance on many downstream tasks, including CV, NLP, and other domains. It aims at grouping similar positive samples closer and repelling negative samples via a standard loss function, \ie, Noise-Contrastive Estimation loss (InfoNCE)~\citep{oord2018representation}. Most of recent CL approaches are focused on studying effective contrastive loss, generation of positive and negative pairs, and sampling methods~\citep{tian2020makes,chen2020simple,he2020momentum,caron2020unsupervised}. For example, MoCo~\citep{he2020momentum}  builds a dynamic dictionary with a queue and a moving-averaged encoder. SimCLR~\citep{chen2020simple} is a simple framework for CL of visual representations with strong data augmentations and a large training batch size. Recently, several efforts have been made to further relieve the requirement of negatives and simplify the conventional CL framework, including BYOL~\citep{grill2020bootstrap}, SimSiam~\citep{chen2021exploring}, and BarlowTwins~\citep{zbontar2021barlow}. From a different perspective rather than using CL as a pre-training strategy for vision and language representation learning~\citep{li2021align,khan2022single}, we explore multi-modal fusion with the contrastive alignment to achieve SOTA performance for the VL tracking task.

\section{VL-SOT500 Dataset}

Before introducing our constructed VL-SOT500 dataset, we first answer a fundamental question: \textbf{\emph{what is the definition of a small object in tracking domain?}} 

Given that the size of the target is $\sqrt{wh}$, where $w$ and $h$ represent the width and height of the target box, respectively. Due to the significant variation in video resolutions, it is unreasonable to determine whether an object is small solely based on the absolute size of its region~\citep{liu2019aggregation}. For instance, an object with an area of $25\times25$ pixels may be considered a relatively large target in a video with a resolution of $256\times256$, but it could be regarded as a small object in a video with a resolution of $4096\times4096$. Following~\citep{zhu2023tiny,zhang2022tracking}, we adopt both the average relative size and the average absolute size to define the size of the object. Specifically, in our work, the small object is defined as having an average relative size smaller than a threshold $s$ (\ie, $1\%$) and an average absolute size smaller than $\sqrt{k \times k}$ (\ie, $\sqrt{22\times 22}$) pixels. Videos containing objects that satisfy both conditions will be selected as candidate videos for our dataset.

Next, we address another question: \textbf{\emph{how to accurately measure the high-speed motion of the small object?}} 

\begin{figure}[t]
  \centering
  \includegraphics[width=1.0\linewidth]{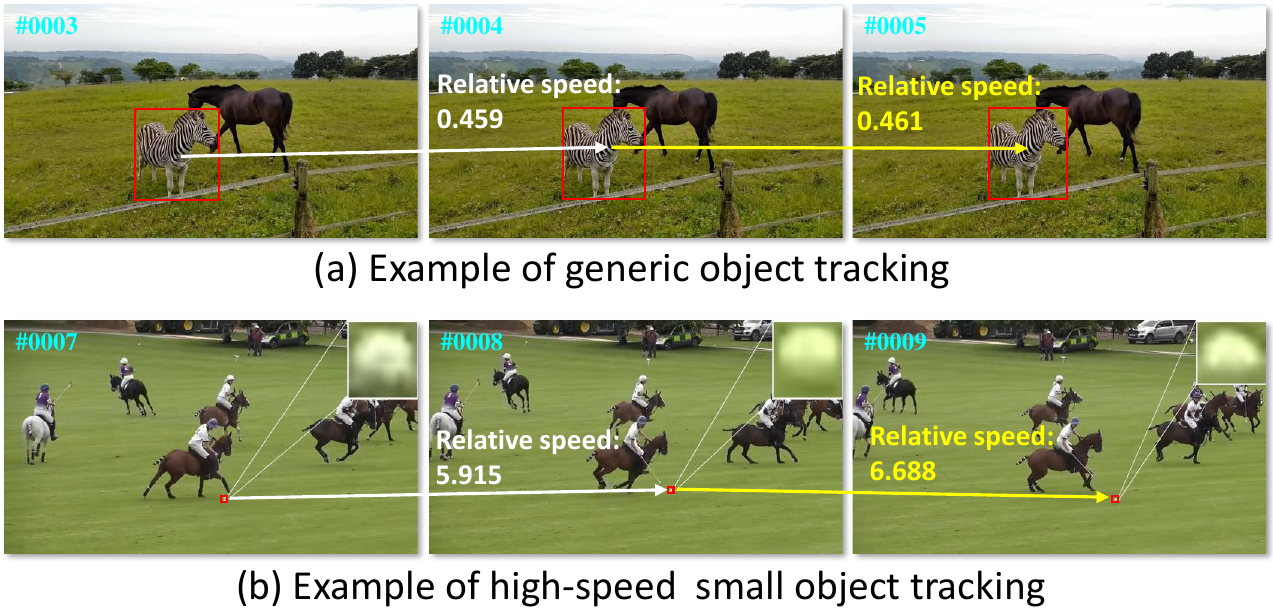}
  \caption{Comparison of (a) generic object tracking and (b) high-speed small object tracking. The latter poses considerably greater challenges, mainly due to the object exhibiting a reduced visual scale and increased relative speeds.}
   \label{fig:Example_Sequences_VLSOT270}
\end{figure}

Small objects inherently contain less effective visual information~\citep{zhu2023tiny}, and if high-speed motion occurs simultaneously, tracking small objects becomes significantly challenging (see Fig.~\ref{fig:Example_Sequences_VLSOT270}). Existing tracking methods typically assume that the target's bounding boxes have only a small displacement between consecutive frames~\citep{huang2020globaltrack,kou2023zoomtrack,zhang2022webuav}, making them unsuitable for high-speed motion. To accurately measure the motion speed of the target, we adopt the relative speed~\citep{valmadre2018long}. Specifically, the target's relative speed in the \( t \)th frame, relative to its size, is defined as follows:
\begin{equation}
\Delta_{t} = \frac{1}{\sqrt{s_{t-1}s_{t}}}\frac{||p_{t}-p_{t-1}||_2}{T_{t}-T_{t-1}},
\label{eq:highspeed}
\end{equation}
where $s_t\!=\!\sqrt{w_t h_t}$ represents the target size, $p_{t}\!=\!(x_t, y_t)$ denotes the target's center coordinate, and $T_t$ indicates the timestamp of frame $t$. Accordingly, we can compute the average relative speed of the target over the entire video/dataset. 

Last but not least, we aim to answer: \textbf{\emph{why existing small object tracking datasets are insufficient and how our dataset uniquely bridges these gaps?}} 

As presented in Tab.~\ref{tab:datasets_comparison} and Fig.~\ref{fig:Example_Sequences}, current small object tracking datasets face several critical limitations that hinder the development and evaluation of tracking algorithms: \textbf{1)} Lack of large-scale, publicly available benchmarks. For instance, LaTOT—the largest existing small object tracking dataset—contains only 165 test videos, while TSFMO remains non-public. \textbf{2)} Absence of language descriptions. All current small-object tracking datasets are vision-only, restricting progress in multi-modal small object tracking research. \textbf{3)} Limited target categories and scenarios (see Tab.~\ref{tab:datasets_comparison}). \textbf{4)} Limited challenges and comprehensiveness. For example, most datasets overlook challenging high-speed motion scenarios, making them inadequate for evaluating cutting-edge methods in such demanding conditions. To overcome these limitations, we introduce a large-scale, multi-modal small object tracking dataset (incorporating both visual and language modalities) with diverse target categories and scenarios covered. It further includes two challenging subsets for generic small object tracking and high-speed small object tracking to comprehensively address the gaps in existing datasets~\citep{fan2019lasot,huang2019got,wang2021towards}.

\subsection{Data Collection and Annotation}
\label{sec:data_collection_and_annotation}

To build a large-scale small object tracking dataset, we follow the common practices of data collection including extensive Internet search\footnote{Raw videos are downloaded from public video websites (\eg, https://www.youtube.com/ and https://www.bilibili.com/) under the Creative Commons 4.0 license, strictly for academic research purposes.} and scientific literature mining~\citep{fan2019lasot,zhang2022webuav,zhang2024webuot,zhang2024towards}. To make our dataset more challenging, we incorporated 165 videos from the LaTOT test set~\citep{zhu2023tiny} as part of our dataset and re-annotated them with language descriptions and attribute annotations. We rigorously check the videos to ensure that they contain rich object categories, scenes, and are suitable for the tracking task. From Tab.~\ref{tab:datasets_comparison} and Fig.~\ref{fig:Example_Sequences}, we can see that our dataset contains the largest number of video sequences and object categories and covers a wide range of real and complex environments, compared to existing small object tracking datasets~\citep{liu2019aggregation,zhang2022tracking,zhu2023tiny}.

After video collection, we perform data annotation, which mainly includes bounding box annotation, attribute annotation, and language annotation. Annotators select eligible targets from the videos and manually label a bounding box $[x, y, w, h]$ for each frame, where $(x, y)$ represents the top-left corner of the target and $(w, h)$ represents the width and height of the target. To provide rich information for precise tracking, we also provide missing labels for each frame (see Tab.~\ref{tab:datasets_comparison}). Following~\citep{wu2015otb,zhang2022webuav}, we annotate 17 challenging tracking attributes, with detailed definitions provided in Section~\ref{sec:attribute_definition}. Following the language annotation practices~\citep{fan2021lasot,fan2019lasot,zhang2022webuav,zhang2024webuot}, we label a language description for each video to describe the target class, color, behavior, attributes, and surroundings of the target to enhance small object tracking with the language modality. Some representative examples are shown in Fig.~\ref{fig:Example_Sequences}.

\begin{table}[t]
{
   \footnotesize
   \caption{\footnotesize Definition of 17 attributes in the VL-SOT500 dataset.}
    \label{tab:Attribute_defination}
	\begin{center}
		\setlength{\tabcolsep}{0.5mm}{
			\begin{tabular}{l|l}
				\Xhline{0.75pt} 
                
				\textbf{Attributes} & \textbf{Definition} \\
				\hline
                    \textbf{01. CM} &  Abrupt motion of the camera. \\
                    
                    \textbf{02. VC} &  Viewpoint affects target appearance significantly. \\

			    \textbf{03. PO} & The target is partially occluded in the sequence. \\
			    
                \textbf{04. FO} &  The target is fully occluded in the sequence. \\
                
                \textbf{05. OV} &  The target completely leaves the video frame. \\

                \textbf{06. ROT} & The target rotates in the video sequence. \\
                
                \textbf{07. DEF} &  The target is deformable during tracking. \\

                \textbf{08. SD} & There is a similar object or background near the target object.\\

                \textbf{09. IV} &  The illumination in the target region changes. \\
                
                \textbf{10. MB} & \tabincell{l}{The target region is blurred due to the target or camera motion.} \\

                \textbf{11. NAO} & The type of the target object is a natural or artificial object. \\

                \textbf{12. PTI} & Only part of the target information is visible in the initial frame.\\ 
                
                \textbf{13. BRI} & \tabincell{l}{The average brightness ($b$) of the video sequence is low ($b \!\leq \!83$),\\ medium ($83 \!<\! b \!\leq\! 119$), or high ($b \!>\!119$).}\\

			\textbf{14. FM} &  \tabincell{l}{The motion of the object is larger than its size.} \\
			    
                \textbf{15. SV} &  The ratio of the bounding box is outside the range [0.5, 2]. \\
                
                \textbf{16. ARV} &  \tabincell{l}{The ratio of bounding box aspect ratio is outside the rage [0.5, 2].} \\
                
                \textbf{17. LEN} & \tabincell{l}{The length ($l$) of current video is short ($l \!\leq \!600$ frames, 20s for \\30 fps), or medium ($600 \!<\! l \!\leq\! 1800$ frames, 60s for 30 fps), \\or long ($l \!>\!1800$ frames).}\\
                
			\Xhline{0.75pt} 
			\end{tabular}
		}
	\end{center}
    }
\end{table}

\subsection{Attribute Definition}
\label{sec:attribute_definition}

To comprehensively evaluate the performance of trackers under various conditions, we define 17 challenging tracking attributes, \eg, deformation (DEF), similar distractors (SD), illumination variations (IV), motion blur (MB), partial target information (PTI), brightness (BRI), and fast motion (FM). The detailed attribute definitions are summarized in Tab.~\ref{tab:Attribute_defination}. Most of the tracking attributes are referenced from popular tracking benchmarks~\citep{fan2019lasot,zhang2024webuot} to ensure they are reasonable. In some complex cases, one video may have multiple attributes. Note that BRI is a tracking attribute newly defined in our work. We found that the average brightness of common nighttime UAV tracking datasets~\citep{li2022all,ye2022unsupervised} is $83$, while the average brightness of common daytime tracking datasets~\citep{huang2019got,fan2019lasot} is $119$. Therefore, we define the average brightness ($b$) of the video sequence as follows: low ($b \leq 83$), medium ($ 83 < b \leq 119 $), or high ($ b > 119 $).

\subsection{Statistics and Analysis}

\begin{figure}[t]
  \centering
  \includegraphics[width=1.0\linewidth]{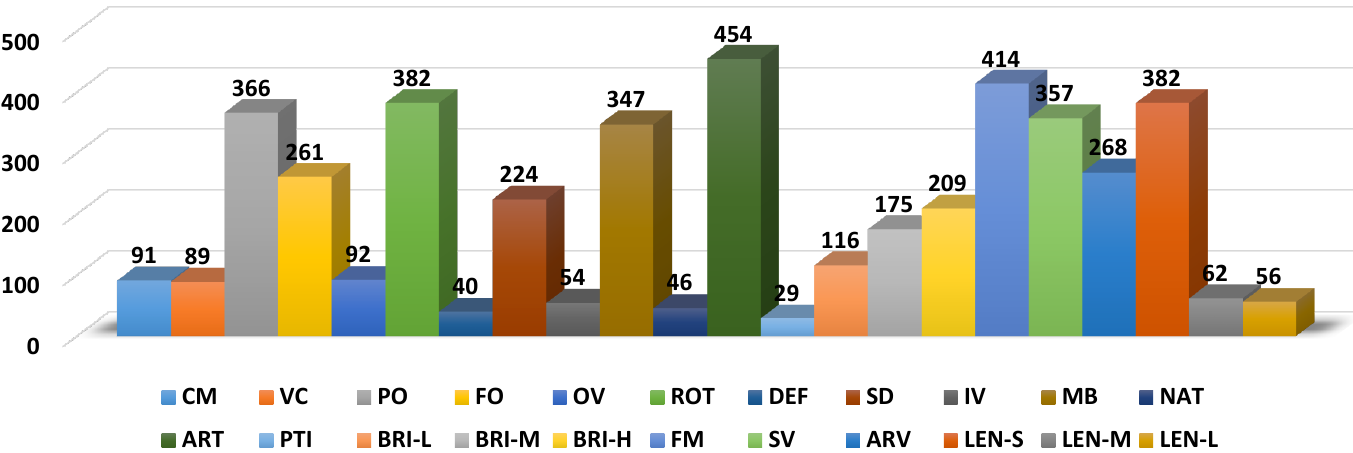}
  \caption{Distribution of each attribute in VL-SOT500.}
   \label{fig:vl_sot230_attribute}
\end{figure}

\begin{figure}[t]
  \centering
  \includegraphics[width=1.0\linewidth]{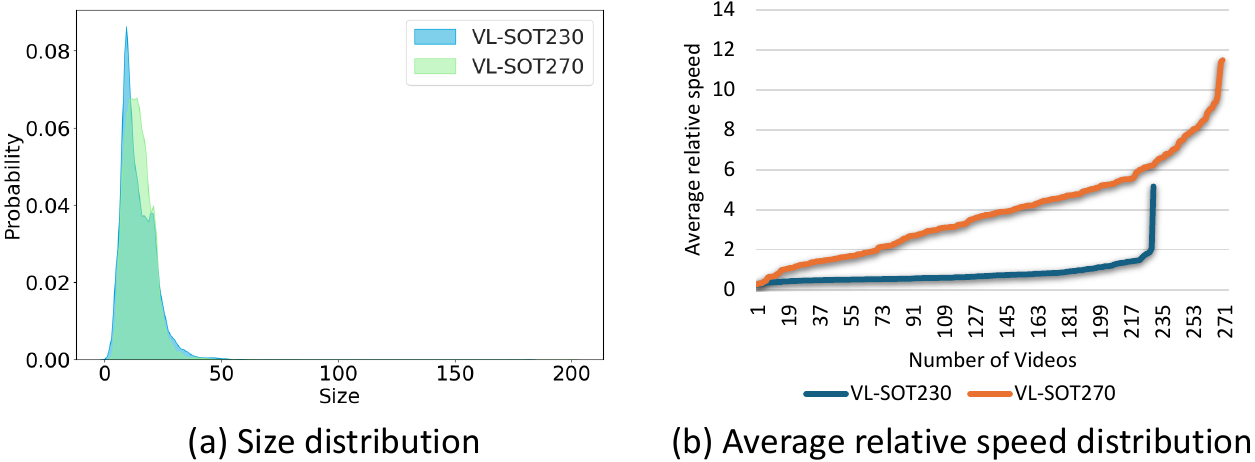}
  \caption{Target size, and average relative speed distributions in VL-SOT500. Best viewed in color and zoomed in.}
   \label{fig:Size_and_Speed_Distribution}
\end{figure}

As shown in Tab.~\ref{tab:datasets_comparison} and Fig.~\ref{fig:Example_Sequences}, we have ultimately constructed a large-scale multi-modal small object tracking dataset, VL-SOT500, with 84 object categories, containing precise bounding boxes and language description annotations. Our dataset consists of 500 video sequences with 252.7 K frames, and the average video length is 505 frames. In addition to the generic small object tracking subset VL-SOT230, we construct a high-speed small object subset VL-SOT270. As shown in Fig.~\ref{fig:Example_Sequences_VLSOT270}, high-speed small object tracking is more challenging than generic object tracking due to the smaller target size and faster motion. From Fig.~\ref{fig:vl_sot230_attribute}, we can observe that our dataset contains diverse tracking attributes, which can facilitate a comprehensive and in-depth evaluation of existing tracking algorithms. Fig.~\ref{fig:Size_and_Speed_Distribution} illustrates the distributions of size and average relative speed in VL-SOT500. The size of the targets varies dramatically across the dataset, ranging from 0 to 200 pixels. The two subsets, VL-SOT230 and VL-SOT270, exhibit similar size distributions, with average target sizes of only 13.8 and 14.3 pixels, respectively, highlighting the challenges posed by their limited dimensions. The average relative speed of the VL-SOT270 subset is significantly higher than that of VL-SOT230, indicating that the former will present greater challenges and substantial opportunities for small object tracking.

\textbf{Compared to existing tracking datasets}, our VL-SOT500 has the following differences: \textbf{1)} Unlike generic object tracking datasets~\citep{fan2019lasot,huang2019got,wang2021towards}, VL-SOT500 is tailored for the challenging small object tracking, making it a valuable testbed for many important real-world applications, \eg, UAV, sports and autonomous driving. \textbf{2)} To the best of our knowledge, our VL-SOT500 is currently the largest and most comprehensive dataset for small object tracking. Specifically, we introduced the VL-SOT230 subset for generic small object tracking, which contains a total of 230.4 K frames, with a mean frame count (\ie, 1002) significantly surpassing that of LaTOT (\ie, 501). Additionally, we proposed the VL-SOT270 subset for high-speed small object tracking, where the average relative speed is 3.5 times higher than that of the previous small object tracking dataset~\citep{zhu2023tiny}. \textbf{3)} Compared to popular small object tracking datasets~\citep{liu2019aggregation,zhang2022tracking,zhu2023tiny}, our dataset includes a richer set of object categories, tracking attributes, and video frames. Notably, the number of our total frames (252.7 K) exceeds that of the previous largest small object tracking dataset, LaTOT (217.7 K)~\citep{zhu2023tiny}. This is because our dataset includes more challenging long videos. \textbf{4)} While TSFMO is primarily used for tracking small and fast-moving objects, especially in sports scenarios, our VL-SOT500 focuses on a wider range of environments (\eg, traffic, river, sky, sports, and indoor). \textbf{5)} Compared to LaTOT, we annotate language descriptions and construct the first multi-modal small object tracking dataset, with more comprehensive experimental evaluations. \textbf{6)} As shown in Tab.~\ref{tab:datasets_comparison}, the objects in VL-SOT500 have an extremely small average target size (\ie, 14.1) and the fastest average relative speed (\ie, 2.469) compared to existing small object tracking datasets, indicating that our dataset is more challenging.

\section{Proposed Method}
\begin{figure*}[t]
  \centering
  \includegraphics[width=1.0\linewidth]{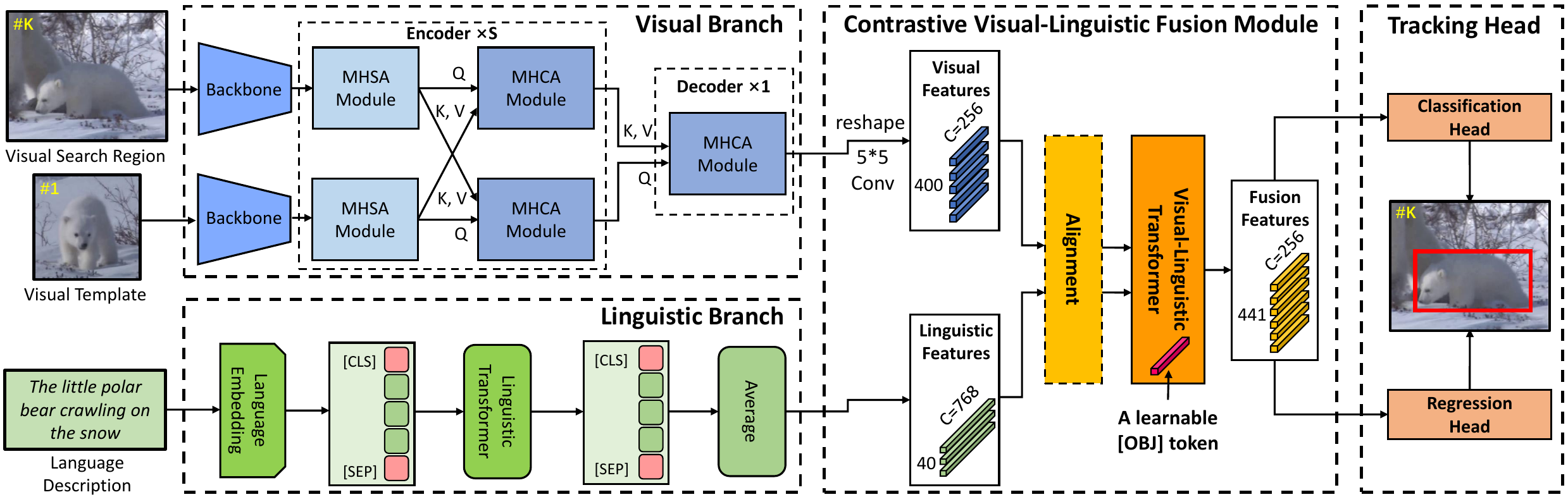}
  \caption{Overview of the proposed COST framework, which contains a visual branch, a linguistic branch, a contrastive visual-linguistic fusion module, and a tracking head to predict object location. The transformer-based visual and language features are extracted by two branches and then fed into the contrastive alignment and the visual-linguistic transformer to learn semantically consistent and unified VL representations in a homogeneous manner. The contrastive alignment learning occurs exclusively during the training phase. For simplicity, the linear projections are omitted.}
   \label{fig:TransVLT}
\end{figure*}

An overview of our COST is shown in Fig.~\ref{fig:TransVLT}, which mainly contains a visual branch for learning visual features, a linguistic branch for learning language features, and a contrastive visual-linguistic transformer with contrastive alignment to achieve a one-stage multi-modal fusion. These components are mainly based on transformers~\citep{vaswani2017attention,devlin2018bert} that enable our method to learn homogeneous VL representations. In addition, an efficient tracking head performs binary classification and bounding box regression based on the advanced multi-modal features to predict the object location. We detail each component in the following subsections. 

\subsection{Preliminary}
In this subsection, we give a brief review of the conventional transformer~\citep{vaswani2017attention}. The fundamental component of the transformer is the attention mechanism. Given input tokens $\mathbf{X}\in \mathbb{R}^{L_{x} \times d}$, they are first linearly projected to the query embedding $\mathbf{Q}$, key embedding $\mathbf{K}$, and value embedding $\mathbf{V}$ using projection matrices, \ie, $(\mathbf{Q}, \mathbf{K}, \mathbf{V})=(\mathbf{X}\mathbf{W}^{Q}, \mathbf{X}\mathbf{W}^{K}, \mathbf{X}\mathbf{W}^{V})$, where $L_{x}$ and $d$ are the length and dimension of tokens $\mathbf{X}$. $\mathbf{W}^{Q/K/V}\in \mathbb{R}^{d \times d_{m}}$ represents the projection matrix for query, key, and value embeddings, respectively, $d_{m}$ denotes the dimension of embeddings. Then, to extract the semantic dependencies between each part, a dot product attention scaled and normalized with a softmax layer is performed. The sequences of values are then weighted by a single-head attention layer computed as ${\rm Attn}(\mathbf{Q},\mathbf{K},\mathbf{V})={\rm softmax}(\frac{\mathbf{Q} \mathbf{K}^{\top}}{\sqrt{d_{k}}})\cdot\mathbf{V}$, where $d_{k}$ is the dimension of the key. This self-attention operation is repeated $h$ times to formulate the multi-head self-attention (MHSA) layer~\citep{vaswani2017attention}, where $h$ is the number of heads. Finally, the output features of the $h$ heads are concatenated along the channel dimension to produce the output of the MHSA layer as follows:
\begin{equation}
 {\rm MultiHead}(\mathbf{Q}, \mathbf{K}, \mathbf{V})={\rm Concat}(\mathbf{H}_{1},...,\mathbf{H}_{h})\mathbf{W}^{O},
\label{eq:MHSA}
\end{equation}
\begin{equation}
\mathbf{H}_{i}={\rm Attn}(\mathbf{QW}_{i}^{Q}, \mathbf{KW}_{i}^{K}, \mathbf{VW}_{i}^{V}),
\label{eq:head}
\end{equation}
where $\mathbf{W}^{O} \in \mathbb{R}^{d\times d_{m}}$ is a projection matrix. Combining an MHSA layer with a simple feed-forward network (FFN), we can obtain the structure of MHSA as shown in Fig.~\ref{fig:Transformer_Encoders}(a). FFN is an MLP composed of fully connected layers and ReLU activation layers. In the MHSA module, each sub-layer is in the form of the residual connection, where layer normalization (LN) is followed by the residual block. Specifically, the sine spatial position encodings $\mathbf{P}$ are first added to input tokens $\mathbf{X}$ to produce $\mathbf{X}_{0}=\mathbf{X}+\mathbf{P}$. Then, the procedure in the MHSA module can be formulated as follows:
\begin{equation}
\mathbf{X}_{l}'={\rm LN}(\mathbf{X}_{l}+{\rm MultiHead}(\mathbf{X}_{l})),
\label{eq:MHSA_Module_1}
\end{equation}
\begin{equation}
\mathbf{X}_{l+1}={\rm LN}(\mathbf{X}_{l}'+{\rm FFN}(\mathbf{X}_{l}')),
\label{eq:MHSA_Module_2}
\end{equation}
where $l$ is the index of the MHSA layer, ${\rm LN(\cdot)}$ is the layer normalization, and ${\rm FFN}(\cdot)$ denotes the feed-forward network. Similar to the MHSA module, the multi-head cross-modal attention (MHCA) module~\citep{vaswani2017attention} is defined as when query embedding, key embedding, and value embedding come from two different tokens $\mathbf{X}_{q}$ and $\mathbf{X}_{kv}$ (see Fig.~\ref{fig:Transformer_Encoders}(b)).

\subsection{Visual Branch}
The visual branch $\mathscr{V}$ consists of a backbone network and a visual transformer as shown in Fig.~\ref{fig:TransVLT}. Following TransT~\citep{chen2021transformer}, we employ a modified version of ResNet50 (V1)~\citep{he2016deep} as the backbone network. Concretely, we remove the last stage of the conventional ResNet50 and use the output of the fourth stage as the output of the backbone network. To increase the resolution of features, the stride of $3\times3$ convolution in the fourth stage is changed from 2 to 1. We further expand the receptive field of the network using a dilated convolution~\citep{yu2017dilated} with a stride of 2 in the fourth stage. As shown in Fig.~\ref{fig:TransVLT}, the visual transformer is composed of encoders and a decoder. Following~\citep{chen2021transformer}, we repeat the encoder $S=4$ times to enhance the learning of intra-modality visual features. There are two MHSA modules followed by two MHCA modules in each encoder. The decoder consists of an 8-head MHCA module for fusing the two feature maps from the last encoder.

Specifically, the visual branch takes the visual search region $x\in\mathbb{R}^{3\times H_{x0}\times W_{x0}}$ and visual template $z\in\mathbb{R}^{3\times H_{z0}\times W_{z0}}$ as the input of the backbone network. The backbone network processes the visual search region and visual template to obtain their features maps $\mathbf{F}_{x}\in\mathbb{R}^{C_{v}\times H_{x}\times W_{x}}$ and $\mathbf{F}_{z}\in\mathbb{R}^{C_{v}\times H_{z}\times W_{z}}$, where $H_{x}, W_{x}=\frac{H_{x0}}{8}, \frac{W_{x0}}{8}$, $H_{z}, W_{z}=\frac{H_{z0}}{8}, \frac{W_{z0}}{8}$, and $C_{v}=1024$. Then, we apply a $1\times1$ convolution to reduce the channel dimension of $\mathbf{F}_{x}$ and $\mathbf{F}_{z}$ to $C_{v}'$ (\ie, 256). Since the input of a transformer encoder is expected to be a sequence of 1D vectors, we further flatten $\mathbf{F}_{x}$ and $\mathbf{F}_{z}$ into $\mathbf{F}'_{x}\in\mathbb{R}^{C_{v}'\times N_{x}}$ and $\mathbf{F}'_{z}\in\mathbb{R}^{C_{v}'\times N_{z}}$, where $N_{x}=H_{x}\times W_{x}$ and $N_{z}=H_{z}\times W_{z}$. The $\mathbf{F}'_{x}$ and $\mathbf{F}'_{z}$ are fed into the visual transformer to generate a 1D vectors $\mathbf{F}_{xz}\in\mathbb{R}^{C_{v}'\times N_{v}}$ (we define $N_{v}=1024$ in this work). Finally, we leverage a reshaping operation and three $5\times 5$ convolution layers with the stride of 1 to obtain the visual features $\mathbf{F}^{v}_{0}\in\mathbb{R}^{C_{v}'\times N_{v}'}$, where $N_{v}'=400$.

\subsection{Linguistic Branch}

Intuitively, the linguistic branch $\mathscr{L}$ can be seen as a twin architecture of the visual branch. As shown in Fig.~\ref{fig:TransVLT}, the linguistic branch mainly contains a language embedding layer, a linguistic transformer, and an averaging operation. We intend to extract semantic information from the language description of the target to reduce ambiguity in the visual branch. The global contextual modeling capacity of BERT~\citep{devlin2018bert} perfectly fits our goal, therefore, the pre-trained BERT$\rm _{BASE}$ model is selected as the linguistic transformer of this branch. Concretely, we denote the number of linguistic transformer layers as 12, the hidden size as 768, and the number of self-attention heads as 12.

\begin{figure}[t]
  \centering
  \includegraphics[width=1.0\linewidth]{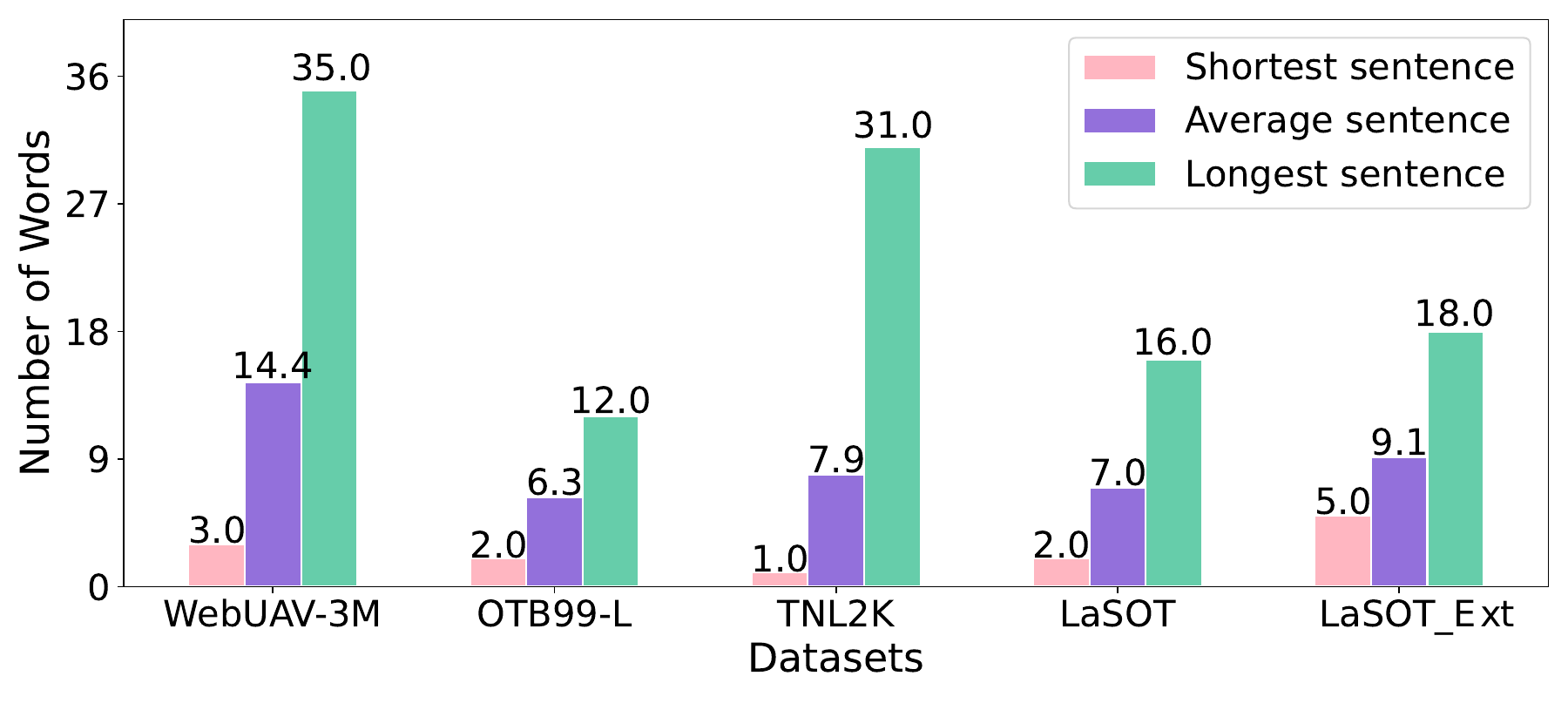} 
  \caption{Distribution of the number of words in sentences on existing VL tracking datasets.}
   \label{fig:number_of_words}
\end{figure}

Given a language description as the input of the linguistic branch, we first convert words into token embeddings and obtain the segmentation embeddings in the language embedding layer. Following~\citep{devlin2018bert}, a special classification token ($[{\rm CLS}]$) and a special separator token ($[{\rm SEP}]$) are added to the beginning and end of the tokenized language embedding, respectively. The maximum length of tokens is set to $K+2$, where $K=38$ is the maximum number of words. For the number of words in the sentence that is less than $K$, zero padding is performed~\citep{guo2022divert}. It should be pointed out that the length of words in most sentences (on existing VL tracking datasets~\citep{zhang2022webuav,fan2019lasot,wang2021towards,li2017tracking,fan2021lasot}, see Fig.~\ref{fig:number_of_words}) is much less than $K$. Similar to the visual branch, to improve the model's sensitivity to position, we also use position embeddings. Therefore, the output of the language embedding layer is the sum of the token embeddings, the segmentation embeddings, and the position embeddings. Then, we feed the language tokens into the linguistic transformer, and average the output of each layer to generate the advanced language features $\mathbf{F}^{l}_{0}\in\mathbb{R}^{C_{l}\times N_{l}}$, where $C_{l}=768$ is the output channel dimension of the linguistic transformer, and $N_{l}=40$ is the number of total language tokens. 

\begin{figure}[t]
  \centering
  \includegraphics[width=1.0\linewidth]{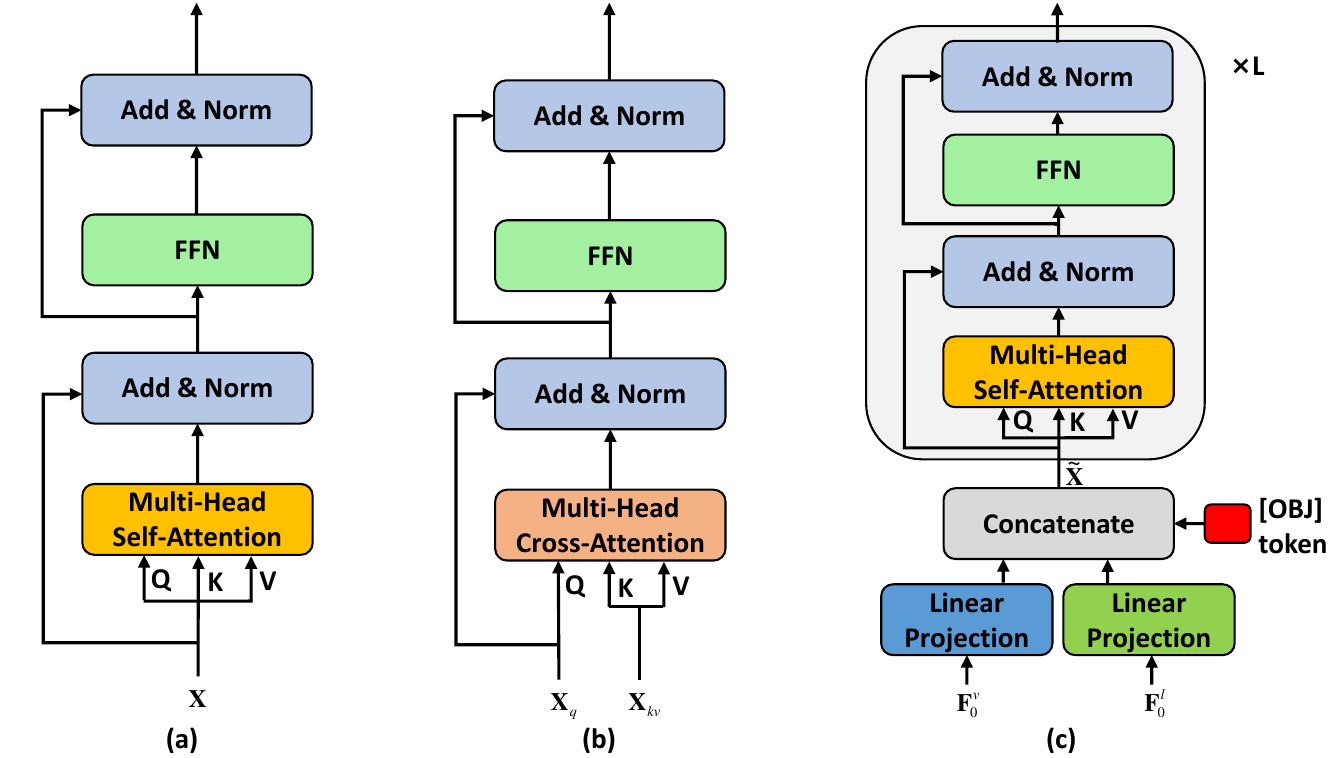} 
  \caption{(a) MHSA module. (b) MHCA module. (c) Our visual-linguistic transformer. Note that positional encodings are added with query and key embeddings, which are not illustrated here for simplicity.}
   \label{fig:Transformer_Encoders}
\end{figure}

\subsection{Contrastive Visual-Linguistic Fusion Module}

\myPara{Contrastive Alignment.} To achieve one-stage multi-modal fusion, we design the CoA to perform cross-modal alignment by pulling embeddings of matching video and language while pushing embeddings of mismatching pairs apart. The reason is that directly fusing high-level language information (\eg, the position, attribute, and behavior of the target) and sparse visual information is intractable, demanding additional designs (\eg, carefully-designed multi-stage multi-modal fusion modules~\citep{feng2021siamese,guo2022divert}). Following~\citep{oord2018representation,yang2022vision}, we adopt the CL to maximize the MI between the matched video and language pairs, which are assumed to contain the same semantic meaning. Given a batch size of $N$, we obtain $N$ language embeddings and $N$ visual embeddings from visual and linguistic branches, denoted as $\{\mathbf{F}^{v}_{0,i}, \mathbf{F}^{l}_{0,i}\}_{i=1}^{N}$. Specifically, we sample negative pairs from the mini-batch. The visual-language embeddings from the same video are treated as positive pairs, while for a given visual or language embedding, both visual and language embeddings from different videos are considered negative samples.

The proposed CoA includes two linear projections ($g_{v}, g_{l}$) for visual and language features. Please note that the linear projections are not shown in Fig.~\ref{fig:TransVLT} for the sake of simplicity. We denote the features of two modalities with the same dimension after projected as $\mathbf{F}_{i}^{v}\in\mathbb{R}^{C_{p}}$ and $\mathbf{F}_{i}^{l}\in \mathbb{R}^{C_{p}}$, respectively, where $C_{p}=256$. For a better understanding of our method, we provide the pseudo-code of the learning of CoA in Algorithm~\ref{alg:CoA}. Formally, InfoNCE losses~\citep{oord2018representation} for explicit vision-to-language and language-to-vision alignment are defined as follows:
\begin{flalign}
    \mathcal{L}_{v2l}(\mathbf{F}^{v}_{i},\mathbf{F}^{l}_{i})=-\sum_{i=1}^{{N}}\log \frac{\exp(sim(\mathbf{F}^{v}_{i},\mathbf{F}^{l}_{i})/\tau)}{\sum_{j=1}^{{N}} \mathbbm{1}_{[j\neq i]}exp(sim(\mathbf{F}^{v}_{i},\mathbf{F}^{l}_{j})/\tau)},&
    \label{eq:v2l_loss}
\end{flalign}
\begin{flalign}
    \mathcal{L}_{l2v}(\mathbf{F}^{l}_{i},\mathbf{F}^{v}_{i})= -\sum_{i=1}^{{N}}\log \frac{\exp(sim(\mathbf{F}^{l}_{i},\mathbf{F}^{v}_{i})/\tau)}{\sum_{j=1}^{{N}} \mathbbm{1}_{[j\neq i]}exp(sim(\mathbf{F}^{l}_{i},\mathbf{F}^{v}_{j})/\tau)},&
    \label{eq:l2v_loss}
\end{flalign}
where $\mathbf{F}^{v}_{i}$ and $\mathbf{F}^{l}_{i}$ are visual and language features from the same video, respectively. $N$ is the batch size, and $sim(\cdot)$ denotes the pairwise cosine similarity~\citep{chen2020simple}. $\mathbbm{1}_{\neq i}\in\{0,1\}$ is an indicator function evaluating to 1 iff $j\neq i$, and $\tau$ is a temperature parameter. Finally, the total CoA loss is:
\begin{equation}
    \mathcal{L}_{CoA}=\frac{1}{2} \mathbbm{E}_{(\mathbf{F}^{v}_{i}, \mathbf{F}^{l}_{i})\sim(\mathbf{F}^{v}, \mathbf{F}^{l})}[\mathcal{L}_{v2l}(\cdot)+\mathcal{L}_{l2v}(\cdot)].
    \label{eq:CL_loss}
\end{equation}

\noindent\textbf{\emph{Remark 1:}} By optimizing $\mathcal{L}_{CoA}$, visual and language features can be well aligned in the embedding space, thereby facilitating subsequent multi-modal fusion and reasoning. In this way, the contrastive alignment can be seen as an effective preprocessing strategy for the one-stage multi-modal fusion.

\begin{algorithm}[t]
\small
\caption{ \small Contrastive Alignment Learning Algorithm.}
\label{alg:CoA}
\begin{algorithmic}[1]
\REQUIRE batch size $N$, temperature $\tau$, visual and language embeddings $\{\mathbf{F}^{v}_{0,i}, \mathbf{F}^{l}_{0,i}\}_{i=1}^{N}$, two linear projections $g_{v}, g_{l}$, visual branch $\mathscr{V}$, linguistic branch $\mathscr{L}$.

\FOR{each sampled minibatch $\{\mathbf{F}^{v}_{0,i}, \mathbf{F}^{l}_{0,i}\}_{i=1}^{N}$}
    \FOR{each $i \in \{1, \ldots, N\}$}
        \STATE $\mathbf{F}_{i}^{v} = g_{v}(\mathbf{F}_{0,i}^{v})$ \hfill {\color{gray}\#~Linear projection}
        
        \STATE $\mathbf{F}_{i}^{l} = g_{l}(\mathbf{F}_{0,i}^{l})$ \hfill {\color{gray}\#~Linear projection}
    \ENDFOR

    \FOR{each $i \in \{1, \ldots, N\}$ and $j \in \{1, \ldots, N\}$}
        \STATE $sim(\mathbf{F}^{v}_{i}, \mathbf{F}^{l}_{j})=\mathbf{F}^{v}_{i}\cdot \mathbf{F}^{l}_{j}/(||\mathbf{F}^{v}_{i}||||\mathbf{F}^{l}_{j}||)$~{\color{gray} \hfill \#~Pairwise cosine similarity}
        
        \STATE $sim(\mathbf{F}^{l}_{i}, \mathbf{F}^{v}_{j})=\mathbf{F}^{l}_{i}\cdot \mathbf{F}^{v}_{j}/(||\mathbf{F}^{l}_{i}||||\mathbf{F}^{v}_{j}||)$~{\color{gray} \hfill \#~Pairwise cosine similarity}
    \ENDFOR

    \STATE define $\mathcal{L}_{v2l}(\cdot)$ as Eq.~(\ref{eq:v2l_loss})
    
    \STATE define $\mathcal{L}_{l2v}(\cdot)$ as Eq.~(\ref{eq:l2v_loss})
    
    \STATE $\mathcal{L}_{CoA}=\frac{1}{2} \mathbbm{E}_{(\mathbf{F}^{v}_{i}, \mathbf{F}^{l}_{i})\sim(\mathbf{F}^{v}, \mathbf{F}^{l})}[\mathcal{L}_{v2l}(\cdot)+\mathcal{L}_{l2v}(\cdot)]$
    
    \STATE update networks $g_{v}$, $g_{l}$, $\mathscr{V}$, and $\mathscr{L}$ to minimize $\mathcal{L}_{CoA}$
\ENDFOR

\STATE \textbf{return} Aligned visual branch $\mathscr{V}$ and linguistic branch $\mathscr{L}$
\end{algorithmic}
\end{algorithm}

\myPara{Visual-Linguistic Transformer.} As shown in Fig.~\ref{fig:Transformer_Encoders}(c), the visual-linguistic transformer follows the basic architecture of transformer~\citep{vaswani2017attention}, which consists of two linear projections (one for each modality), a learnable $[{\rm OBJ}]$ token, and a stack of transformer encoder layers. Following~\citep{vaswani2017attention,devlin2018bert,chen2021transformer}, learnable position embeddings are added to the input tokens of each transformer encoder layer to retain position information.

Specifically, the visual-linguistic transformer takes the aligned visual features $\mathbf{F}^{v}_{0}\in\mathbb{R}^{C_{v}'\times N_{v}'}$ and language features $\mathbf{F}^{l}_{0}\in\mathbb{R}^{C_{l}\times N_{l}}$ as inputs. Then, these multi-modal features are projected as visual embedding $\widetilde{\mathbf{F}}^{v}\in\mathbb{R}^{C_{p}\times N_{v}'}$ and linguistic embedding $\widetilde{\mathbf{F}}^{l}\in\mathbb{R}^{C_{p}\times N_{l}}$ with the same channel dimension using two linear projections, where $C_{p}=256$. To facilitate the tracking model to learn multi-modal associations, we add a learnable $[{\rm OBJ}]$ token $\mathbf{O}\in\mathbb{R}^{C_{p}\times 1}$ to $\widetilde{\mathbf{F}}^{v}$ and $\widetilde{\mathbf{F}}^{l}$. It is randomly initialized at the beginning, and optimized to learn object-aware multi-modal corresponding from both visual tokens and language tokens during the whole model training. Formally, we define the joint input tokens of the visual-linguistic transformer calculated by a concatenate operation, \ie, $\rm Concat(\cdot)$, as follows:
\begin{equation}
    \widetilde{\mathbf{X}}={\rm Concat}(\widetilde{\mathbf{F}}^{v}_{1},\widetilde{\mathbf{F}}^{v}_{2},...,\widetilde{\mathbf{F}}^{v}_{C_{v}'},\widetilde{\mathbf{F}}^{l}_{1},\widetilde{\mathbf{F}}^{l}_{2},...,\widetilde{\mathbf{F}}^{l}_{N_{l}},\mathbf{O}),
    \label{eq:fusion_module}
\end{equation}
where $\widetilde{\mathbf{X}}\in \mathbb{R}^{C_{p}\times(N_{v}'+N_{l}+1)}$ denotes the joint input tokens, $\widetilde{\mathbf{F}}^{v}_{1},\widetilde{\mathbf{F}}^{v}_{2},...,\widetilde{\mathbf{F}}^{v}_{C_{v}'}$ and $\widetilde{\mathbf{F}}^{l}_{1},\widetilde{\mathbf{F}}^{l}_{2},...,\widetilde{\mathbf{F}}^{l}_{N_{l}}$ are visual tokens and language tokens. Then, we feed the joint input tokens into the visual-linguistic transformer to learn unified representations by encoding $\widetilde{\mathbf{X}}$ into a shared semantic space.

\noindent\textbf{\emph{Remark 2:}} Although the architecture of the visual-linguistic transformer is simple, we will empirically verify that it can achieve efficient one-stage multi-modal fusion and significantly improve tracking performance. In Section~\ref{sec:ablation_study}, we will demonstrate that the $[{\rm OBJ}]$ token in Eq.~(\ref{eq:fusion_module}) is beneficial to learn consolidated and unified VL representations as it is enriched by both visual and linguistic tokens.

\begin{figure}[t]
  \centering
  \includegraphics[width=1.0\linewidth]{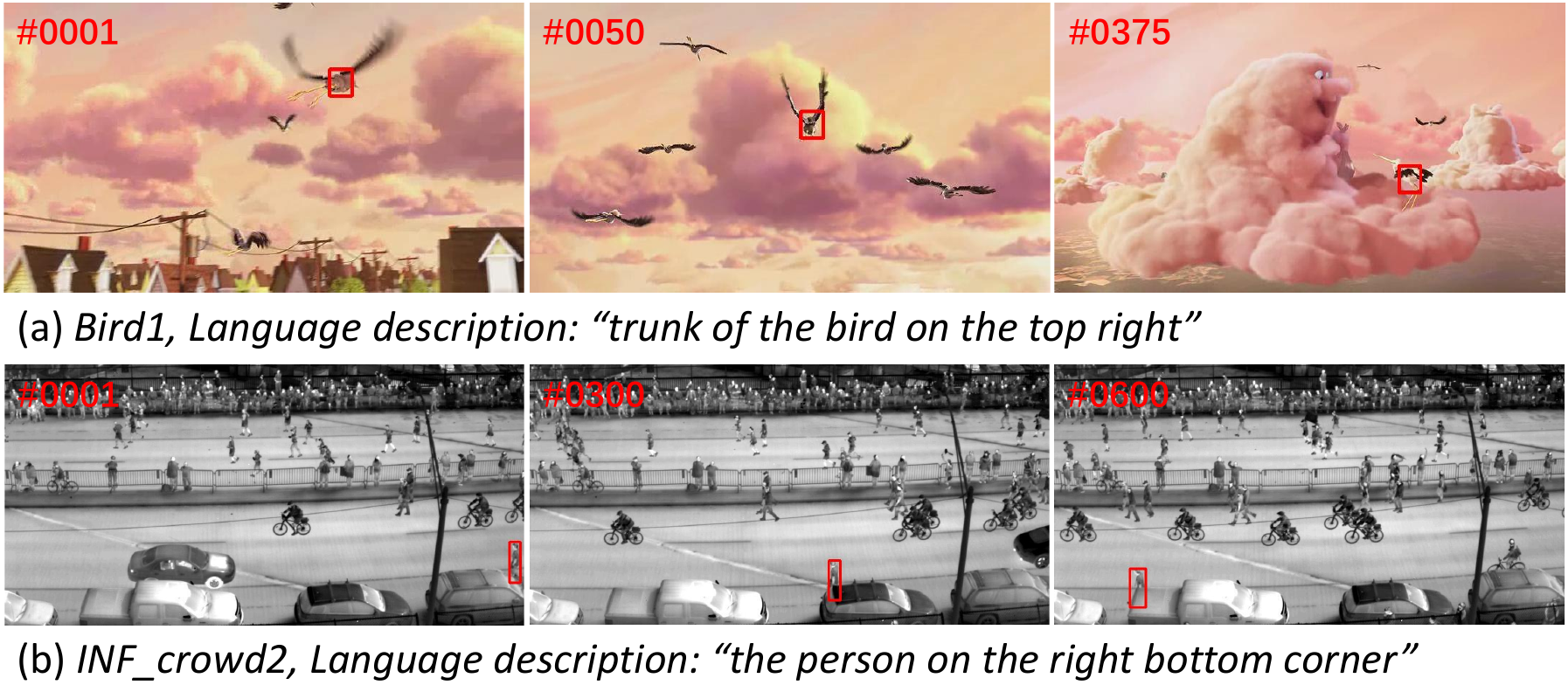} 
  \caption{Ambiguous language annotations on existing VL tracking datasets. \emph{Bird1} and \emph{INF\_crowd2} are from~\citep{li2017tracking} and~\citep{wang2021towards}.}
   \label{fig:Ambiguous_language_annotations}
\end{figure}

\subsection{Tracking Head and Loss}

\myPara{Tracking Head.} Following~\citep{danelljan2019atom,chen2021transformer}, the tracking is decoupled as a problem of binary classification and bounding-box regression. Correspondingly, the tracking head in this work consists of a classification head and a regression head, which are two layers of MLP and one layer of MLP~\citep{pinkus1999approximation}, respectively. For classification, pixels within the ground-truth box are positive samples, otherwise negative samples. The classification head is applied to improve the foreground-background discrimination ability of the model. The regression head can enhance the localization ability of the model by predicting the center coordinates of the target, as well as the width and height of the target.

\begin{figure}[t]
  \centering
  \includegraphics[width=1.0\linewidth]{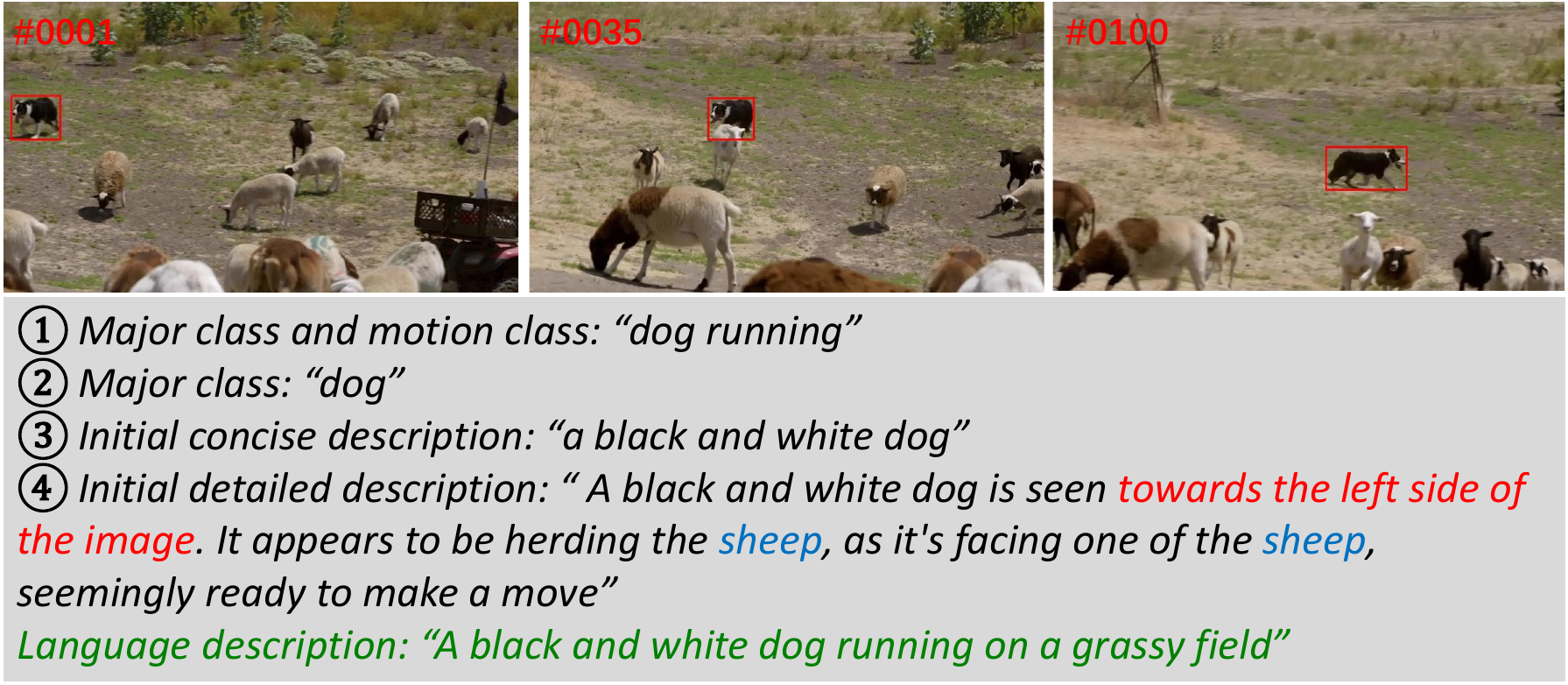} 
  \caption{ An example of four different types of language annotations. The initial detailed description may contain some redundant (\eg, {\color{ifblue}\emph{sheep}}) or even erroneous (\eg, {\color{red}\emph{towards the left side of the image}}) information for tracking. Following the language annotation rule in Section~\ref{sec:data_collection_and_annotation}, we can obtain a more accurate and concise {\color{darkgreen}\emph{language description}}.}
   \label{fig:GOT10k_language_annotations}
\end{figure}

\begin{table}[t]
\footnotesize
  \centering
    \caption{Quality of pseudo-language descriptions on the GOT-10K and TrackingNet training datasets. We use the CLIP score to measure the reliability of different types of language annotations, \ie, \Circled{\scriptsize 1} major class and motion class, \Circled{\scriptsize 2} major class, \Circled{\scriptsize 3} initial concise, and \Circled{\scriptsize 4} initial detailed descriptions.}
  \setlength{\tabcolsep}{4.3mm}{
  \begin{tabular}{lcccc}
    \toprule 
    Dataset &   Language Annotation Type &  CLIP Score   \\
    \midrule

   \multirow{4}*{GOT-10K} & Major Class and Motion Class  &  0.635293 \\
    
    & Major Class  & 0.625388  \\
    
    & Initial Concise~\citep{li2024dtvlt} & 0.635145  \\
    
    & Initial Detailed~\citep{li2024dtvlt} & 0.639789  \\
    \hline
   TrackingNet & Major Class  &  0.626654 \\
    \bottomrule
  \end{tabular}
  }
  \label{tab:CLIP_Score}
\end{table}

\myPara{Training Loss.} For classification, we use the binary cross-entropy loss, $\mathcal{L}_{ce}=\sum_{i}y_{i}\log(p_{i})+(1-y_{i})\log(1-p_{i})$, where $y_{i}$ denotes the ground-truth label, and $p_{i}$ denotes the predicted confidence. For regression, our model is trained in an end-to-end manner with the combination of $\ell_{1}$-norm loss $\mathcal{L}_{1}(\cdot)$~\citep{carion2020end} and the generalized IoU loss $\mathcal{L}_{GIoU}(\cdot)$~\citep{rezatofighi2019generalized}. Like~\citep{chen2021transformer}, only positive samples (\ie, predicted bounding boxes) are considered when calculating the regression loss:
\begin{equation}
\mathcal{L}_{reg}=\sum_{i}\mathbbm{1}_{y_{i}=1}[\lambda_{1}\mathcal{L}_{1}(B_{i},\hat{B_{i}})+\lambda_{G}\mathcal{L}_{GIoU}(B_{i},\hat{B_{i}})],
    \label{eq:regression_loss}
\end{equation}
where $y_{i}=1$ represents the positive sample, $B_{i}$ and $\hat{B_{i}}$ denote the ground-truth bounding box and predicted bounding box, respectively, and $\lambda_{1}$, $\lambda_{G}$ are two hyper-parameters. 

The overall loss for COST is $\mathcal{L}=\mathcal{L}_{CoA}+\mathcal{L}_{reg}+\alpha\mathcal{L}_{ce}$, where $\alpha$ is a balance factor.

\begin{table*}[t]
\footnotesize
\centering
\caption{Overall results of 35 representative trackers (including CNN, CNN-Transformer, and Transformer based methods) on the proposed VL-SOT230 dataset. TransT is the baseline of the proposed COST. ``Trans" denotes Transformer in the feature column. The best results are marked in \textbf{bold}.}
\label{tab:Overall_results}
\begin{center}
\setlength{\tabcolsep}{2.4mm}{
\begin{tabular}{lcccccccc}

	\toprule
	\multirow{2}*{Method} & \multirow{2}*{Publication} & 
      \multicolumn{5}{c}{Performance}  & \multirow{2}*{Feature} & \multirow{2}*{VL-based}\\
        
	\cmidrule(r){3-7}	& & AUC (\%) & $P$ (\%)  & $P_{norm}$ (\%) & cAUC (\%) & mACC (\%) & \\
				
		\midrule	
               SiamFC~\citep{bertinetto2016fully} & ECCVW-2016 & 23.4 & 40.2 & 26.9 & 22.9 & 23.6 & CNN & \ding{55} \\
               
               ECO~\citep{danelljan2017eco} & CVPR-2017 & 23.6 & 46.8 & 27.3 & 22.9 & 23.6 & CNN & \ding{55} \\

               VITAL~\citep{song2018vital} & CVPR-2018 & 15.3 & 272 & 17.6 & 14.9 & 15.3 & CNN & \ding{55} \\

               ATOM~\citep{DanelljanBKF19} & CVPR-2019 & 24.3 & 49.0 & 27.4 & 23.5 & 24.3 & CNN & \ding{55} \\
               
               SiamPRN++~\citep{LiWWZXY19} & CVPR-2019 & 24.8 & 44.7 & 28.7 & 24.2 & 24.9 & CNN & \ding{55} \\
               
               DiMP~\citep{BhatDGT19iccv} & ICCV-2019 & 29.2 & 50.7 & 34.5 & 28.6 & 29.4 & CNN & \ding{55} \\
                
               Ocean~\citep{zhang2020ocean} & ECCV-2020 & 18.8 & 34.5 & 20.4 & 17.8 & 18.4 & CNN & \ding{55} \\
               
               KYS~\citep{bhat2020know} & ECCV-2020 & 30.4 & 51.9 & 35.3 & 29.7 & 30.6 & CNN & \ding{55} \\

               SiamFC++~\citep{XuWLYY20} & AAAI-2020 & 21.0 & 38.4 & 24.4 & 20.3 & 20.9 & CNN & \ding{55} \\
               
               PrDiMP~\citep{danelljan2020probabilistic} & CVPR-2020 & 26.9 & 45.7 & 30.9 & 26.4 & 27.1 & CNN & \ding{55} \\

               SiamBAN~\citep{ChenZLZJ20} & CVPR-2020 & 24.6 & 42.7 & 27.8 & 24.1 & 24.8 & CNN & \ding{55} \\

               SiamCAR~\citep{guo2020siamcar} & CVPR-2020 & 26.1 & 46.9 & 28.3 & 25.4 & 26.3 & CNN & \ding{55} \\

                LightTrack~\citep{yan2021lighttrack}  & CVPR-2021 & 21.8 & 36.5 & 23.5 & 20.2 & 20.9 & CNN & \ding{55} \\
                               
                SiamGAT~\citep{guo2021graph} & CVPR-2021 & 20.6 & 39.1 & 20.9 & 19.8 & 20.6 & CNN & \ding{55} \\
                  
               STMTrack~\citep{fu2021stmtrack} & CVPR-2021 & 25.0 & 44.3 & 29.0 & 24.3 & 25.0 & CNN & \ding{55} \\

                AutoMatch~\citep{zhang2021learn} & ICCV-2021 & 21.6 & 38.9 & 22.3 & 20.5 & 21.2 & CNN & \ding{55} \\

               HiFT~\citep{cao2021hift} & ICCV-2021 & 21.9 & 39.3 & 24.6 & 20.5 & 21.2 & CNN & \ding{55} \\

              STARK-ST50~\citep{yan2021learning}  & ICCV-2021 & 30.0 & 50.3 & 31.4 & 29.2 & 30.1 & CNN+Trans & \ding{55} \\

               TCTrack~\citep{cao2022tctrack} & CVPR-2022 & 22.4 & 42.4 & 24.8 & 21.7 & 22.4 & CNN & \ding{55} \\
               
               UDAT~\citep{ye2022unsupervised} & CVPR-2022 & 28.2 & 45.4 & 31.8 & 27.6 & 28.3 & CNN & \ding{55} \\

               TransInMo~\citep{guo2022learning} & IJCAI-2022 & 29.3 & 49.7 & 33.2 & 28.8 & 29.5 & Trans & \ding{55} \\
                              
               OSTrack~\citep{ye2022joint} & ECCV-2022 & 28.5 & 47.5 & 31.0 & 27.7 & 28.4 & Trans & \ding{55} \\
            
               Aba-ViTrack~\citep{li2023adaptive} & ICCV-2023 & 27.4 & 46.4 & 29.6 & 26.5 & 27.2 & Trans & \ding{55} \\
               
               GRM~\citep{gao2023generalized} & CVPR-2023 & 28.9 & 48.0 & 31.3 & 28.0 & 28.8 & Trans & \ding{55} \\
               
               ARTrack~\citep{wei2023autoregressive} & CVPR-2023  & 30.5 & 52.4 & 32.9 & 29.5 & 30.4 & Trans & \ding{55} \\

               SeqTrack-B256~\citep{chen2023seqtrack} & CVPR-2023 & 29.9 & 50.0 & 31.8 & 29.1 & 30.0 & Trans & \ding{55} \\
               
               ZoomTrack~\citep{kou2023zoomtrack} & NeurIPS-2023 & 30.6 & 51.0 & 33.0 & 29.8 & 30.6 & Trans & \ding{55} \\

                \midrule
                
               VLT\_SCAR~\citep{guo2022divert} & NeurIPS-2022 & 21.3 & 39.9 & 22.6 & 20.6 & 21.3 & CNN & \ding{51} \\

               VLT\_TT~\citep{guo2022divert} & NeurIPS-2022 & 25.2 & 44.2 & 28.3 & 24.7 & 25.3 & CNN+Trans & \ding{51} \\

                JointNLT~\citep{zhou2023joint} & CVPR-2023  & 20.3 & 35.7 & 21.5 & 19.5 & 20.2 & Trans & \ding{51} \\
                 
               MMTrack~\citep{zheng2023towards} & TCSVT-2023 & 29.2 & 47.9 & 30.9 & 28.4 & 29.3 & Trans & \ding{51} \\

              CiteTracker-256~\citep{li2023citetracker} & CVPR-2023  & 27.0 & 45.1 & 29.1 & 25.9 & 26.7 & Trans & \ding{51} \\
              
             UVLTrack~\citep{ma2024unifying} & AAAI-2024 & 30.7 & 52.3 & 32.9 & 30.0 & 30.9 & Trans & \ding{51} \\

             \midrule
            
             TransT~\citep{chen2021transformer} & CVPR-2021 & 30.2 & 52.1 & 35.0 & 29.7 & 30.5 & CNN+Trans & \ding{55} \\

             \textbf{COST} & \textbf{Ours} & \textbf{33.3{\scriptsize~\textcolor{darkgreen}{(+3.1\%)}}} & \textbf{56.2{\scriptsize~\textcolor{darkgreen}{(+4.1\%)}}} & \textbf{37.9{\scriptsize~\textcolor{darkgreen}{(+2.9\%)}}} & \textbf{32.6{\scriptsize~\textcolor{darkgreen}{(+2.9\%)}}} & \textbf{33.6{\scriptsize~\textcolor{darkgreen}{(+3.1\%)}}} & CNN+Trans & \ding{51} \\

	       \bottomrule
			\end{tabular}
		}
	\end{center}
\end{table*}

\section{Experimental Evaluation}

\subsection{Experimental Setup}
We implement our COST using Python 3.6 and Pytorch 1.10.2. The speed of COST is 36 frames per second (FPS) with a single NVIDIA RTX 3090 GPU. 

\myPara{Offline Training.} The tracker is optimized using AdamW optimizer with learning rate { 0.0001} and decay rate { 0.0001}. We train the tracker for 1,000 epochs in total, sampling 4,200 video-language pairs per epoch. Following~\citep{chen2021transformer}, the hyper-parameters $\lambda_{1}$ and $\lambda_{G}$ are set to 5 and 2, respectively. We set $\tau=0.5$ and $\alpha=1$. The batch size $N$ is set to 14. During training, we normalize the images using mean and standard deviation statistics from ImageNet~\citep{deng2009imagenet}. On the search frame and template frame, we crop 4 times and 2 times of the target box to obtain the visual search region and visual template region. Then, the visual search and template regions are resized to 256×256 and 128×128, respectively. 

The training data includes the training splits of four VL tracking datasets (\ie, OTB99-L~\citep{li2017tracking}, LaSOT~\citep{fan2019lasot}, TNL2K~\citep{wang2021towards}, WebUAV-3M~\citep{zhang2022webuav}), two visual tracking datasets (\ie, GOT-10k~\citep{huang2019got}, and TrackingNet~\citep{muller2018trackingnet}), and COCO~\citep{lin2014microsoft}.  We adopt two common data augmentation techniques (\ie, random translation and brightness jitter)~\citep{chen2021transformer} to enlarge the training set. For two visual tracking datasets (\ie, GOT-10k, TrackingNet) without language annotations, we follow~\citep{guo2022divert,guo2024divert} to provide a pseudo-language description for each video. To reduce ambiguity, we only concatenate words of the major class and motion class as the pseudo-language description for GOT-10k (see Fig.~\ref{fig:GOT10k_language_annotations}). Since TrackingNet only provides major class annotations, we use them as pseudo-language descriptions (see Tab.~\ref{tab:CLIP_Score}). In Fig.~\ref{fig:GOT10k_language_annotations}, we present a manually annotated language description (green font) following the language annotation rule in Section~\ref{sec:data_collection_and_annotation}. The language description is more precise and contains no redundant information. However, manually annotating language descriptions is highly costly, so we still utilize pseudo-language descriptions on GOT-10k and TrackingNet. To verify the reliability of the pseudo-language descriptions, we use the CLIP score~\citep{radford2021learning} to measure the consistency between video frames and pseudo-language descriptions (see Tab.~\ref{tab:CLIP_Score}). Note that we also compare two fine-grained descriptions (\ie, initial concise and initial detailed descriptions) from~\citep{li2024dtvlt}. The results show that the two types of pseudo-language descriptions (\ie, major class and motion class, major class) we used achieve comparable CLIP scores compared to the fine-grained initial concise and initial detailed descriptions. From Tab.~\ref{tab:CLIP_Score} and Fig.~\ref{fig:GOT10k_language_annotations}, we can observe that although the initial detailed description has the highest CLIP score, it may contain errors (\eg, ``towards the left side of the image'') or distracting information (\eg, ``sheep''). More discussions about the reliability of pseudo-language descriptions are provided in Section~\ref{sec:Further_Discussions}.

\begin{figure*}[t]
\vspace{-1.5cm}
\centering
\subfloat{\includegraphics[width =0.25\linewidth]{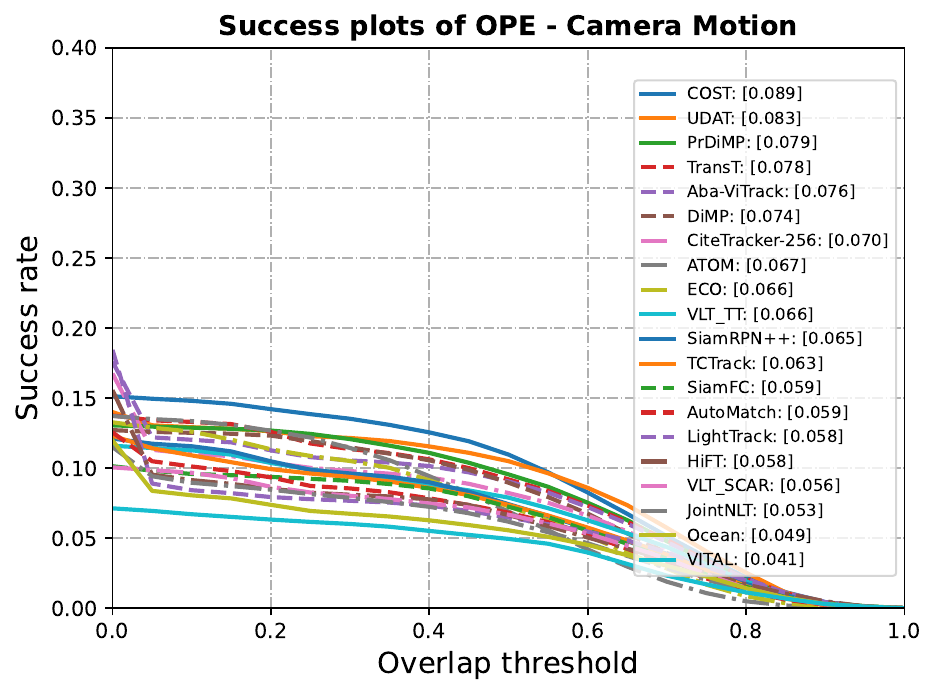}}
~\subfloat{\includegraphics[width =0.25\linewidth]{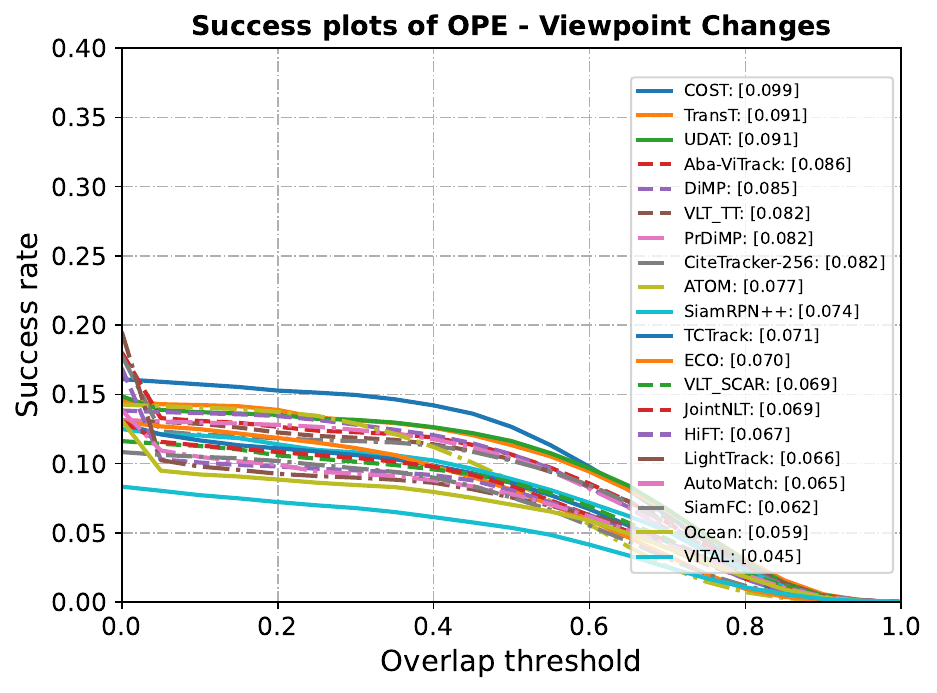}}
~\subfloat{\includegraphics[width =0.25\linewidth]{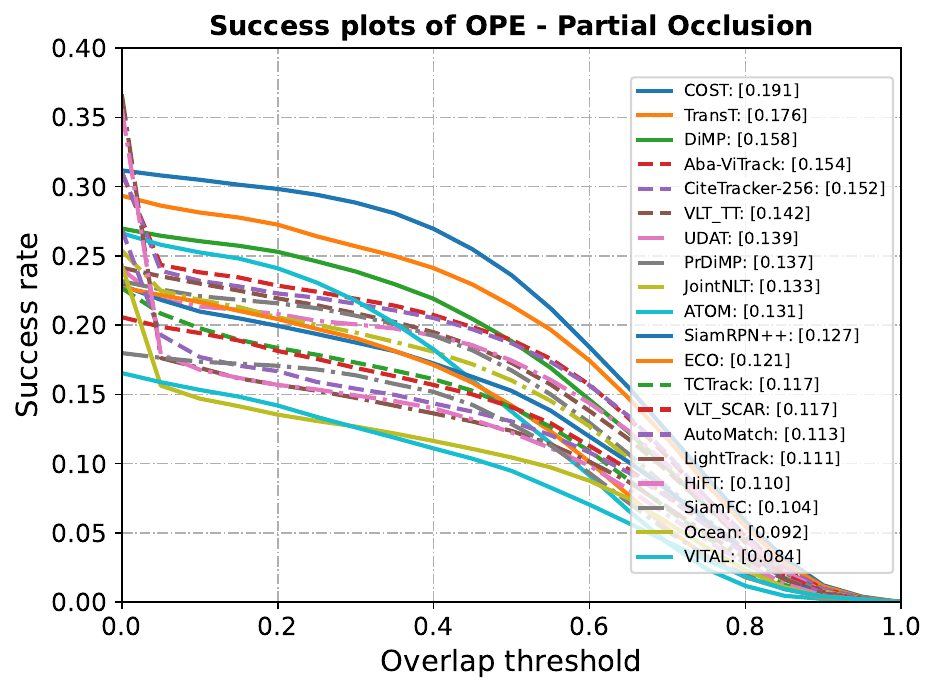}}\\
\vspace{-0.3cm}
\subfloat{\includegraphics[width =0.25\linewidth]{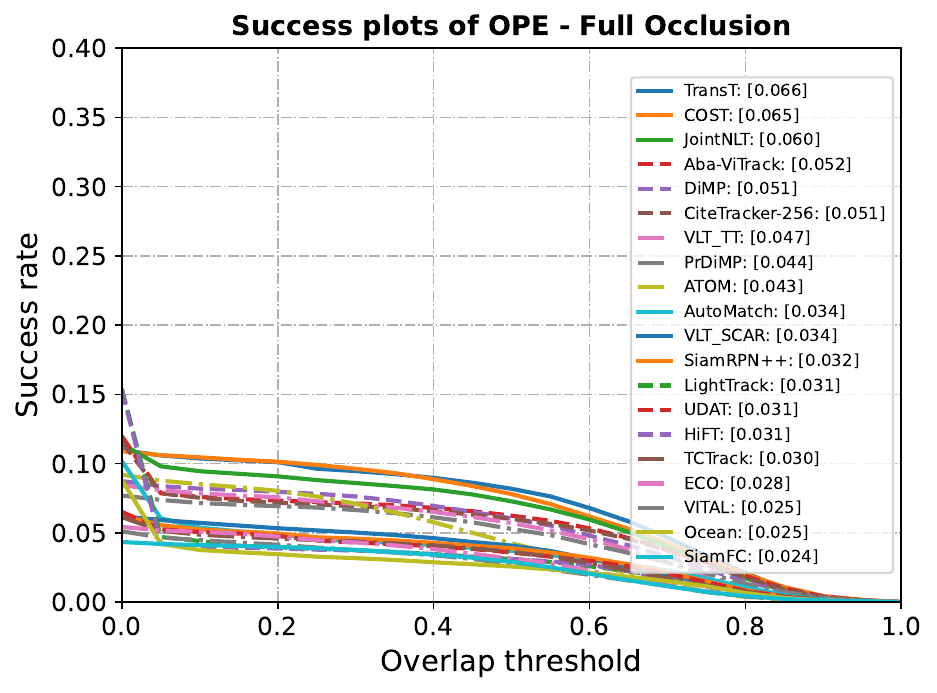}}
~\subfloat{\includegraphics[width =0.25\linewidth]{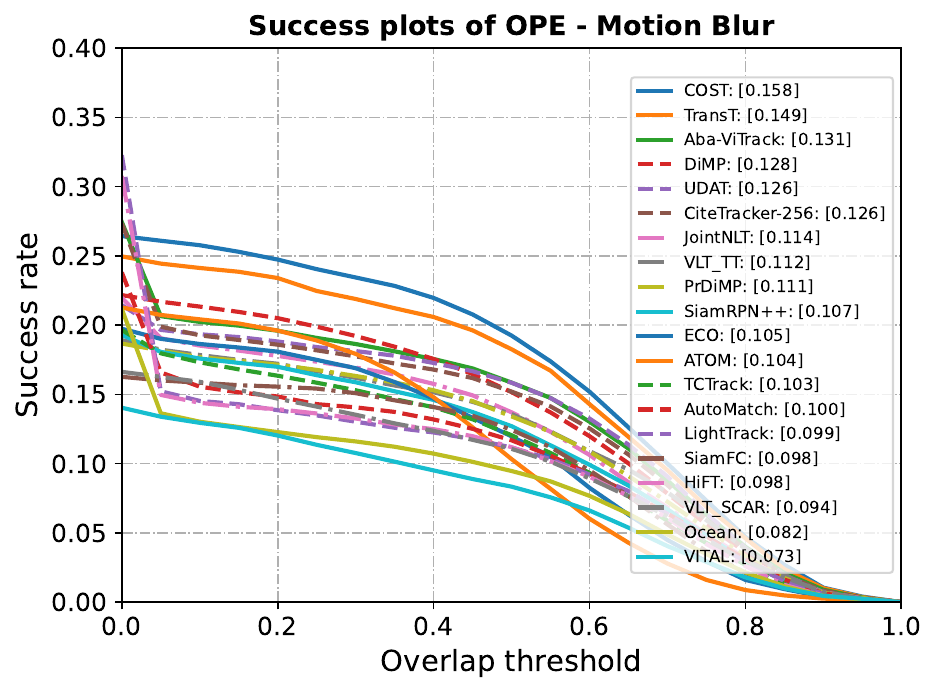}}
~\subfloat{\includegraphics[width =0.25\linewidth]
{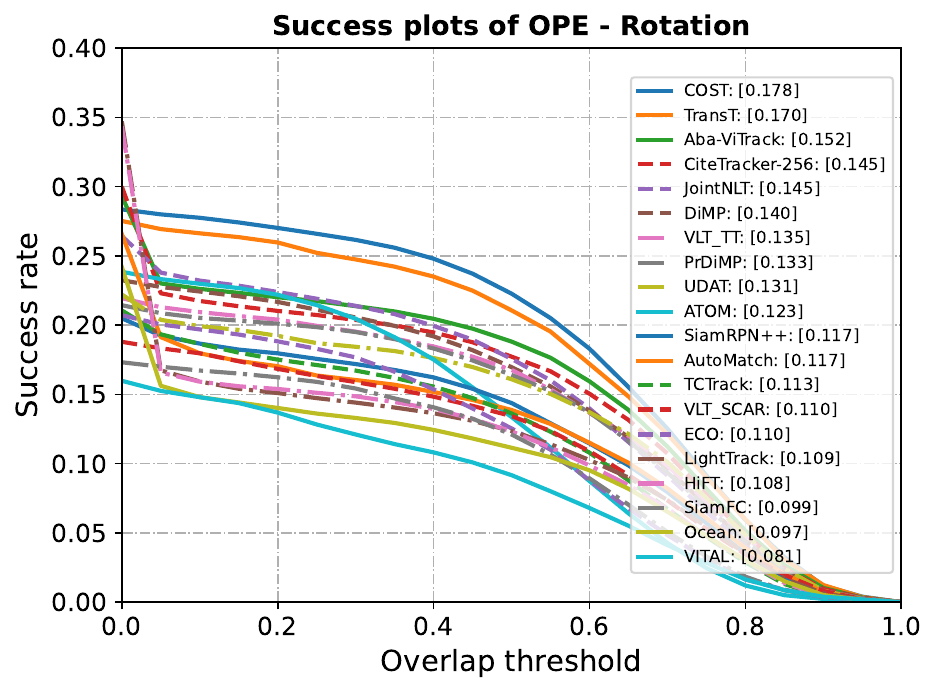}}\\
\vspace{-0.3cm}
\subfloat{\includegraphics[width =0.25\linewidth]
{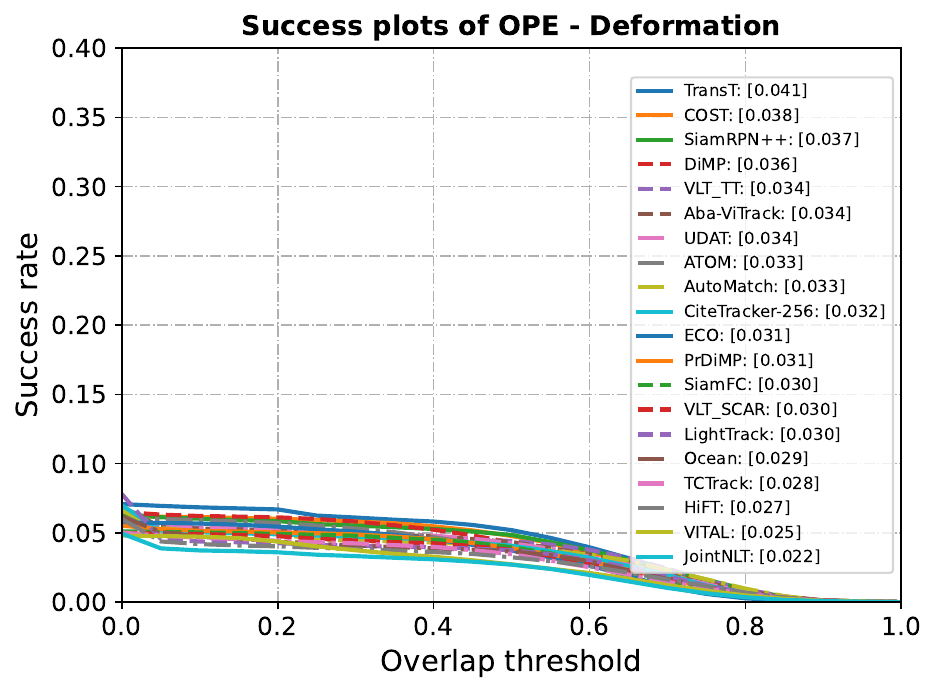}}
~\subfloat{\includegraphics[width =0.25\linewidth]{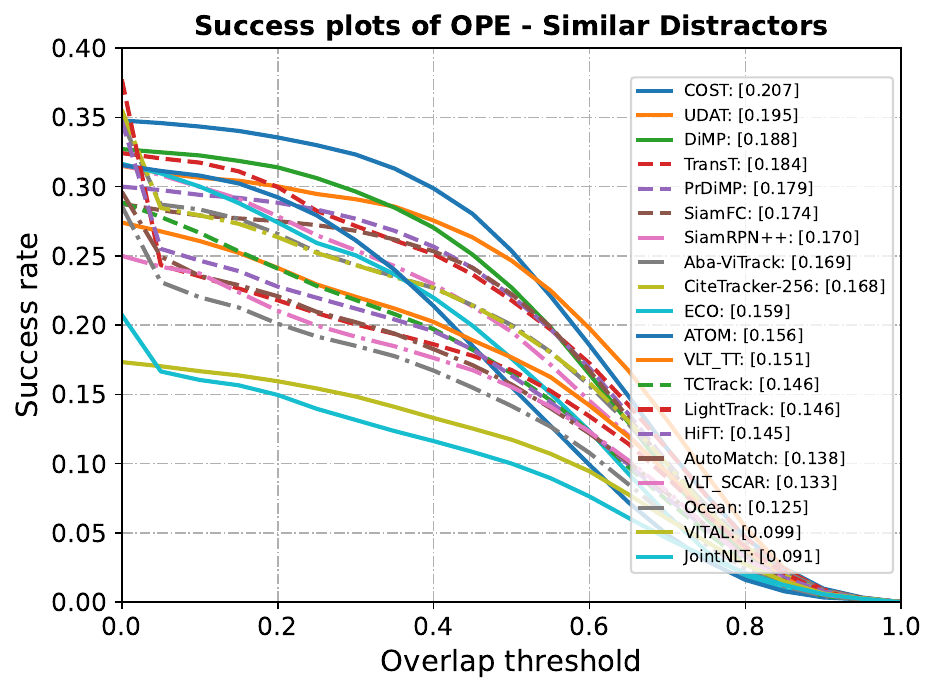}}
\subfloat{\includegraphics[width =0.25\linewidth]{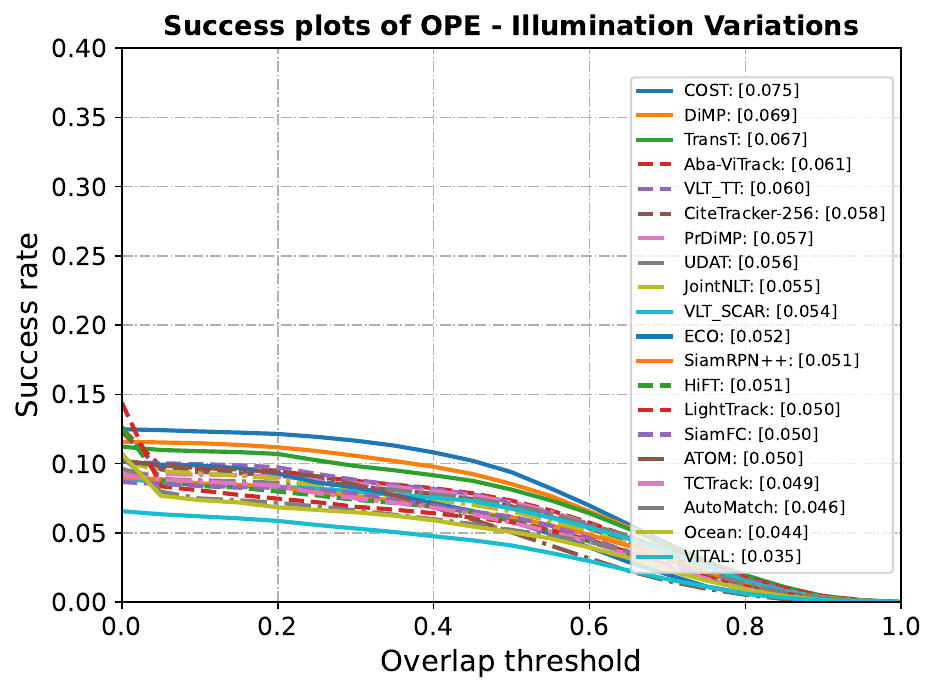}}\\
\vspace{-0.3cm}
\subfloat{\includegraphics[width =0.25\linewidth]{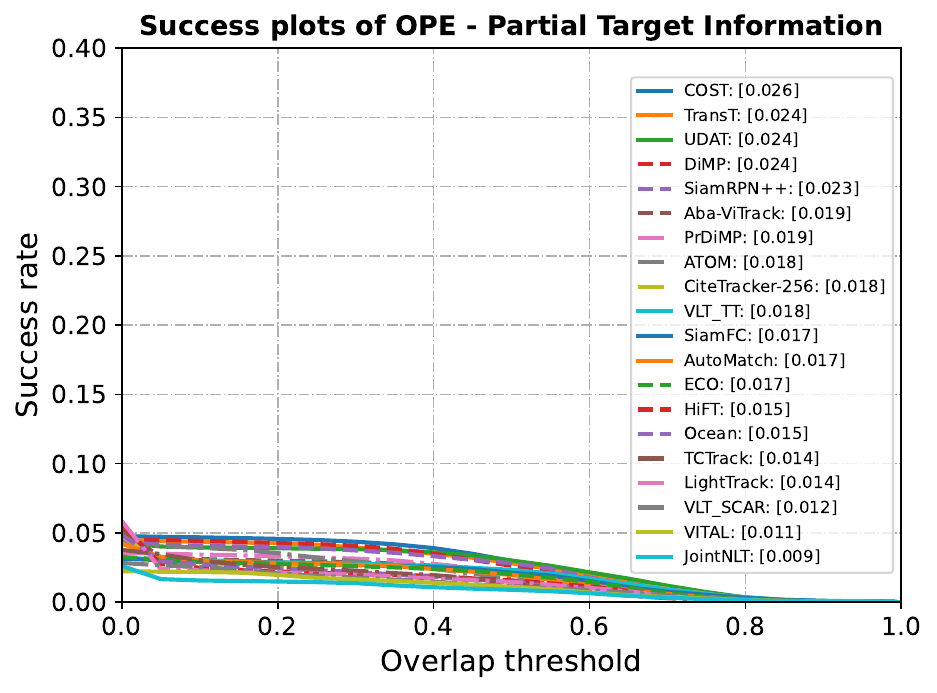}}
~\subfloat{\includegraphics[width =0.25\linewidth]{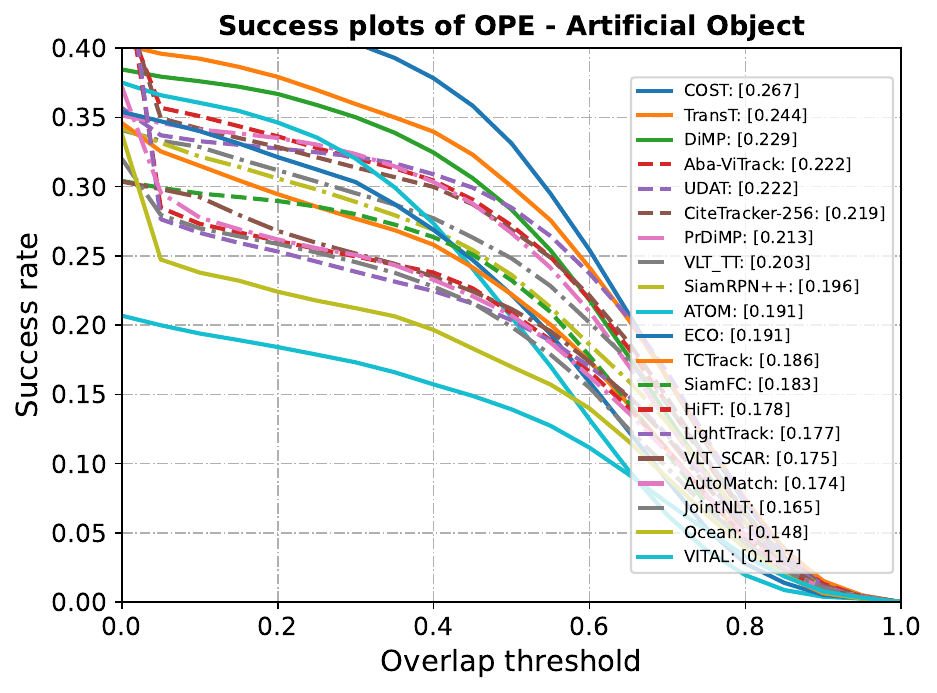}}
~\subfloat{\includegraphics[width =0.25\linewidth]{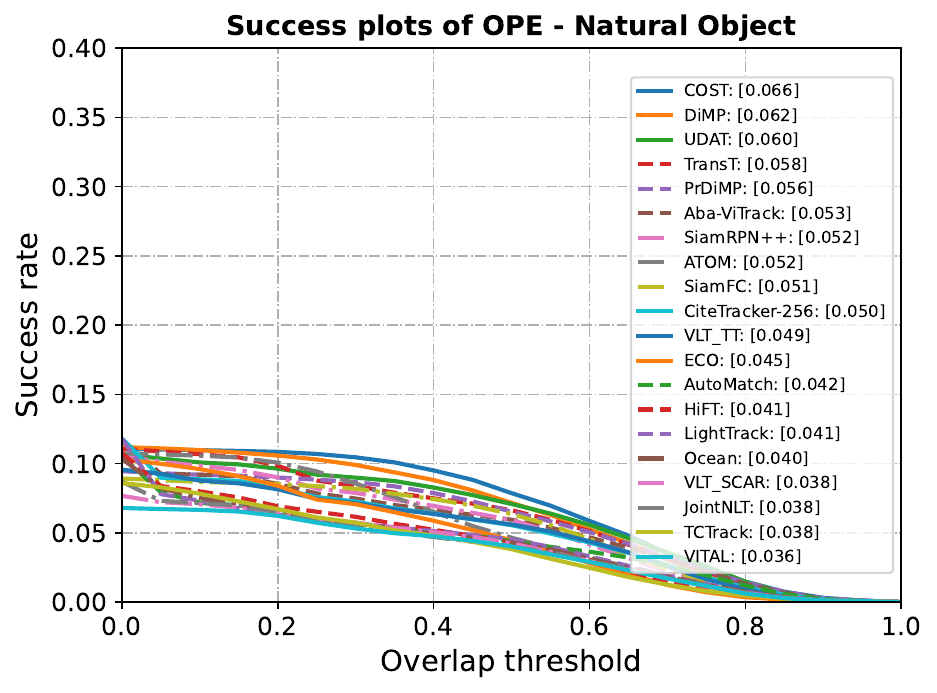}}\\
\vspace{-0.3cm}
\subfloat{\includegraphics[width =0.25\linewidth]
{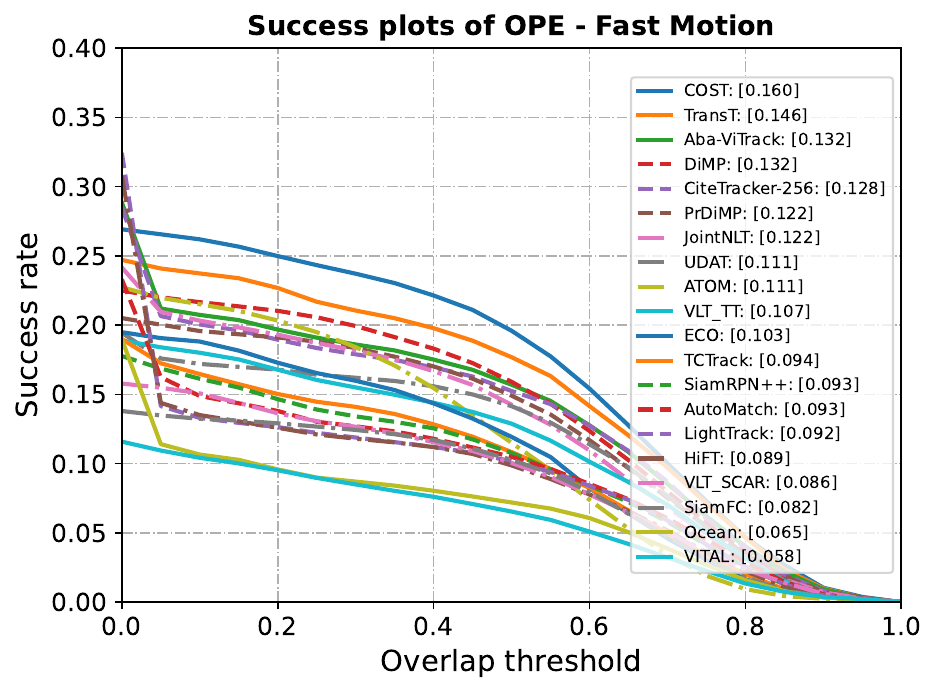}}
~\subfloat{\includegraphics[width =0.25\linewidth]{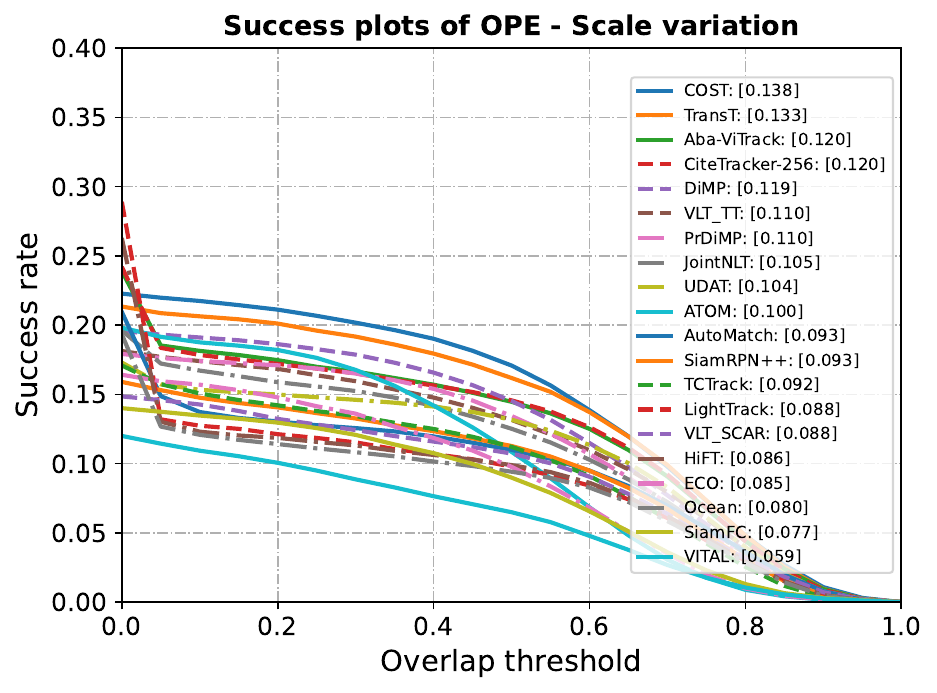}}
~\subfloat{\includegraphics[width =0.25\linewidth]{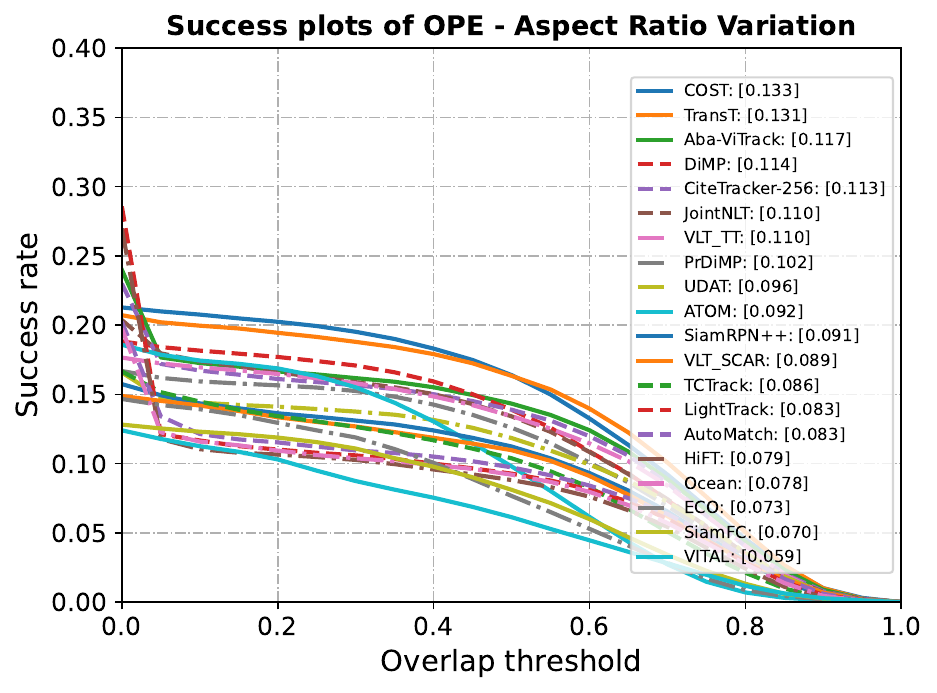}}

\caption{Performance of SOTA trackers on different tracking attributes on VL-SOT230. In each sub-figure, trackers are ranked by the success rate. Best viewed in color with zooming in.}
\label{fig:VLSOT230_attributes_1}
\vspace{-0.5cm}
\end{figure*}

\myPara{Online Tracking.} In online tracking, the tracking head produces confidence scores for 441 candidate boxes. Following~\citep{chen2021transformer}, we use a $21\times21$ Hanning window to penalize the confidence scores and select the box with the highest confidence score as the prediction for the current frame. We evaluation trackers on five existing datasets (\ie, OTB99-L, TNL2K, LaSOT, LaSOT\_Ext~\citep{fan2021lasot}, and WebUAV-3M), and our VL-SOT500. Five metrics, \ie, precision ($P$), normalized precision ($P_{norm}$), success rate (AUC), complete success rate (cAUC)~\citep{zhang2022webuav}, and mean
accuracy (mACC)~\citep{nam2023antiuav} are used to measure the performance of different trackers.

\begin{figure*}[ht]
\centering
\subfloat{\includegraphics[width =0.3\linewidth]{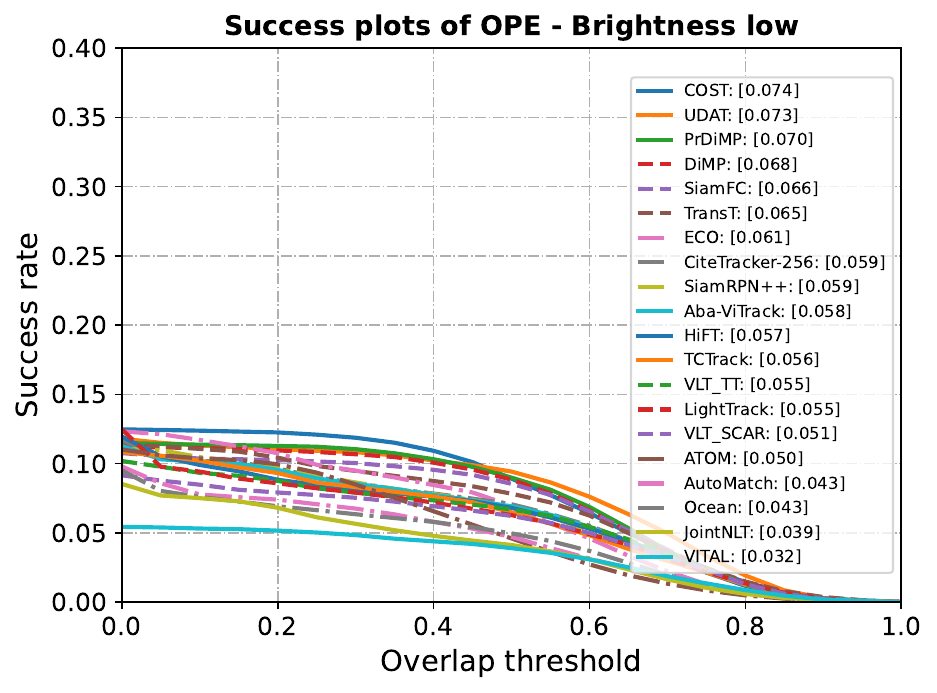}}
~\subfloat{\includegraphics[width =0.3\linewidth]{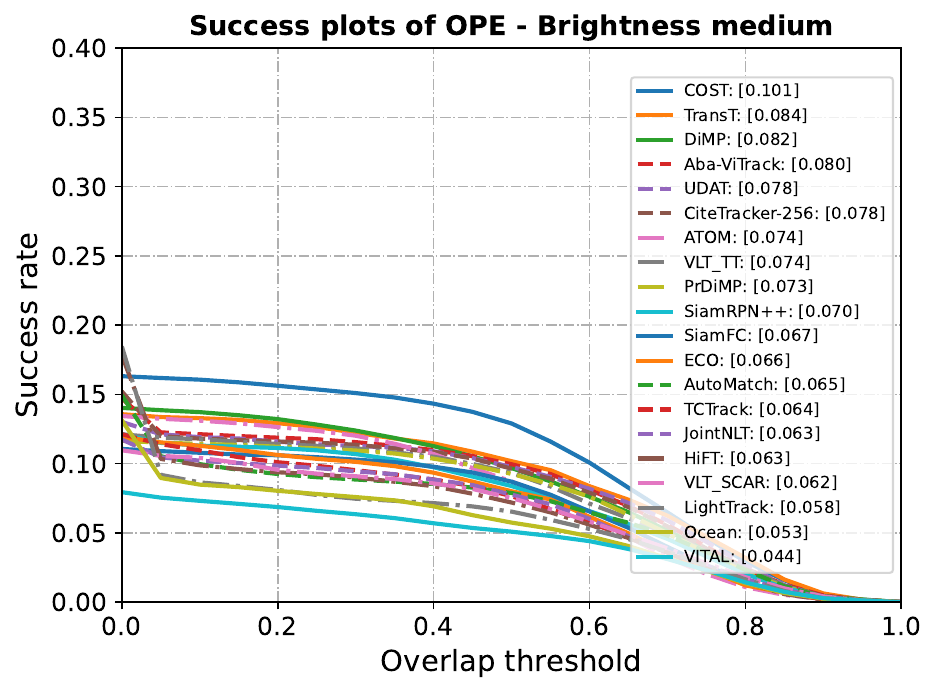}}
~\subfloat{\includegraphics[width =0.3\linewidth]{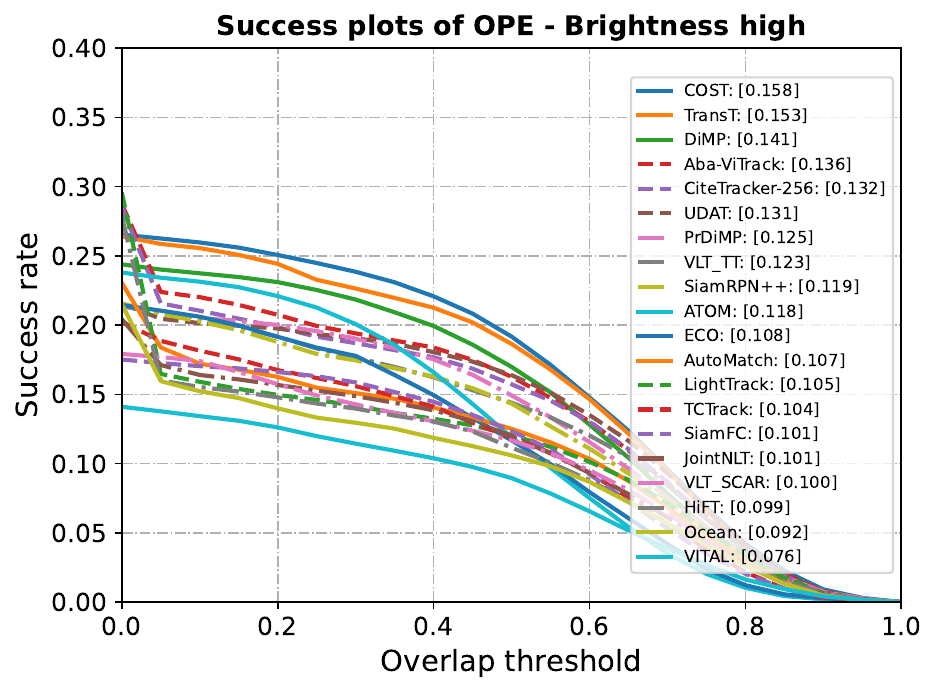}}\\
\vspace{-0.3cm}
\subfloat{\includegraphics[width =0.3\linewidth]{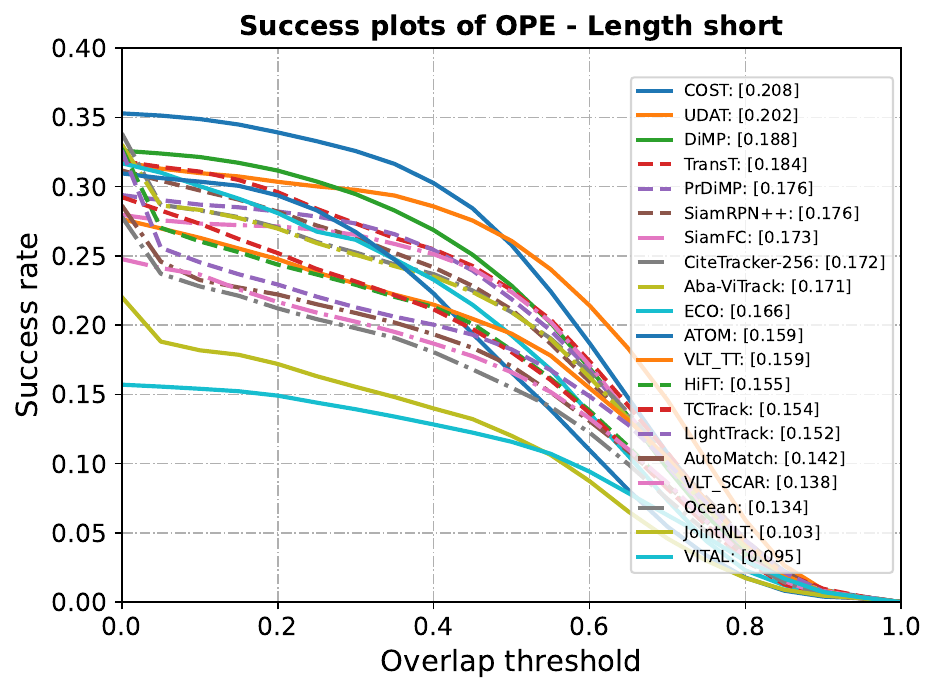}}
~\subfloat{\includegraphics[width =0.3\linewidth]{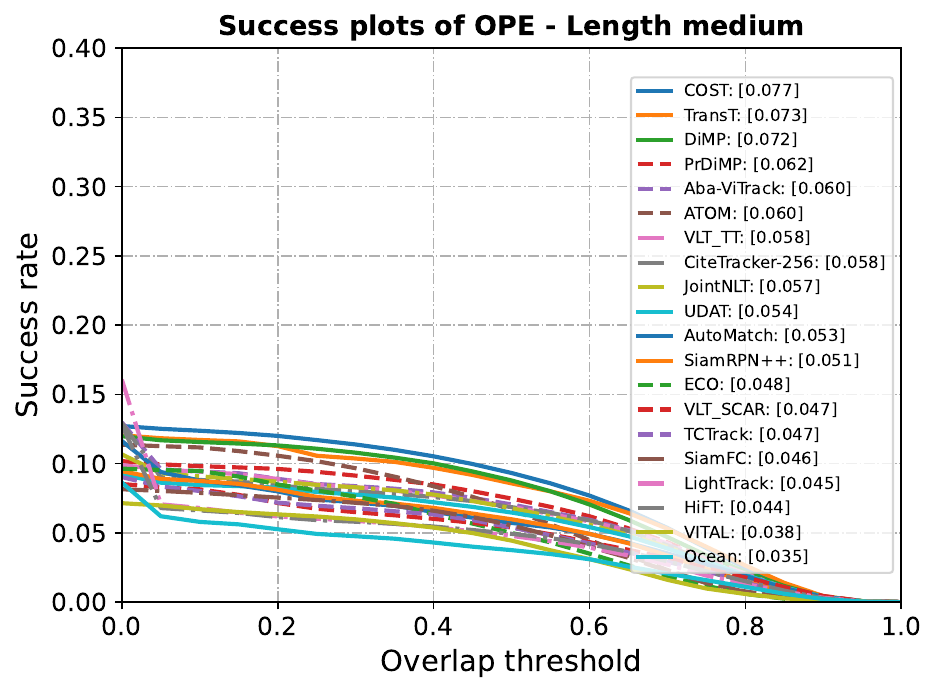}}
~\subfloat{\includegraphics[width =0.3\linewidth]{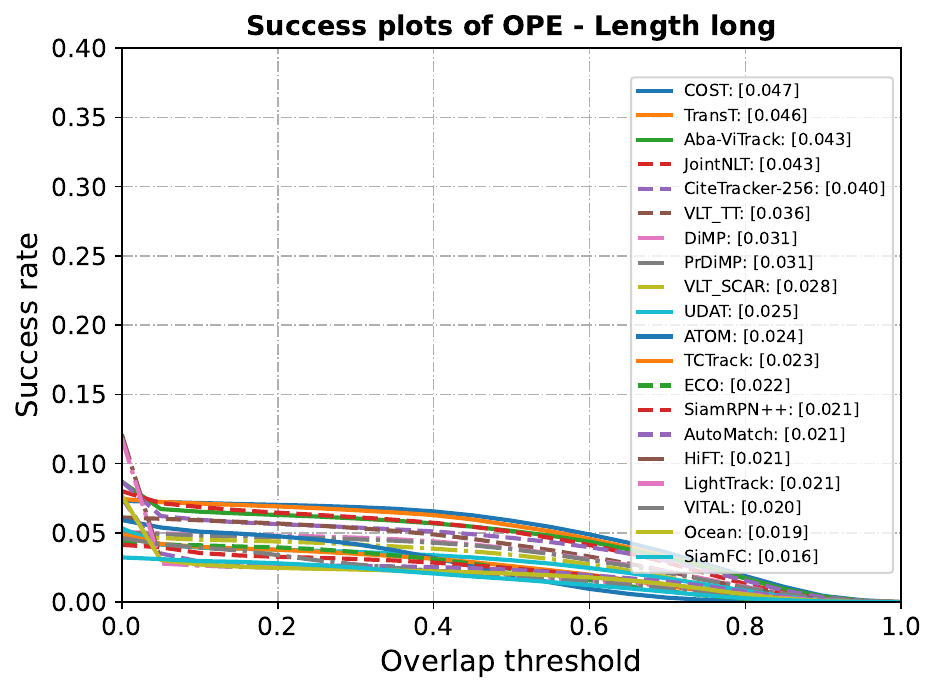}}
\caption{Evaluation with different brightness and video length on VL-SOT230. The details regarding the definitions of brightness and video length can be found in Tab.~\ref{tab:Attribute_defination}. Best viewed in color with zooming in.}
\label{fig:VLSOT230_attributes_2}
\end{figure*}

For a fair experimental comparison, we evaluate our tracker with two settings: 1) On the LaSOT test set, LaSOT\_Ext, VL-SOT230 and VL-SOT270, we adopt \emph{aligned training data.} Following the recent SOTA VL/visual trackers~\citep{guo2022divert,wang2021transformer,chen2021transformer,yan2021learning,danelljan2020probabilistic,feng2021siamese}, we use four training sets (\ie,~LaSOT, GOT-10k, COCO, and TrackingNet) with bounding boxes and language annotations to train our tracker in this setting. 2) On OTB99-L~\citep{li2017tracking}, TNL2K~\citep{wang2021towards}, and WebUAV-3M test sets, we adopt \emph{complete training data.} Specifically, we further train the tracker based on the previously pretrained weights using the training sets of OTB99-L~\citep{li2017tracking}, TNL2K~\citep{wang2021towards}, and WebUAV-3M~\citep{zhang2022webuav}. The reason is that most of the language descriptions provided by these datasets describe the target based on the first frame. Existing research~\citep{zhang2023all} demonstrates that ambiguous language descriptions (see Fig.~\ref{fig:Ambiguous_language_annotations}) cannot accurately describe the state of the target throughout the video sequence, and therefore using these ambiguous language descriptions may mislead the tracker. Thus, we discard the linguistic branch before testing on these datasets~\citep{li2017tracking,wang2021towards,zhang2022webuav}, and fine-tune the visual branch on corresponding training splits. According to contrastive learning theory~\citep{radford2021learning}, after the aligned training, the visual branch and the linguistic branch are aligned in the semantic space. Therefore, the features extracted from the visual branch are aligned with language features in the semantic space, where the implicit linguistic information in the visual branch helps to enhance the robustness of tracking. The impact of the above two training settings will be discussed in Section~\ref{sec:Further_Discussions}.

\subsection{Evaluation on VL-SOT230 Dataset}

\myPara{Overall Performance.} VL-SOT230 is our newly proposed multi-modal generic small object tracking dataset, consisting of 230 video sequences with high-quality bounding box annotations and language descriptions. We comprehensively evaluate 35 advanced visual and VL trackers on VL-SOT230. The overall performance is summarized in Tab.~\ref{tab:Overall_results}. The top three trackers are COST, UVLTrack, and ZoomTrack, which all adopt either CNN+Transformer or Transformer for feature extraction. These advanced trackers highlight the powerful modalities modeling capabilities of Transformer~\citep{vaswani2017attention}. Specifically, our COST outperforms the baseline algorithm TransT by 3.1\%, 4.1\%, 2.9\%, 2.9\%, and 3.1\% in terms of AUC, $P$, $P_{norm}$, cAUC, and mACC scores, respectively. Moreover, compared to two of the latest SOTA VL trackers (\ie, MMTrack~\citep{zheng2023towards} and UVLTrack~\citep{ma2024unifying}), which employ more advanced techniques (\ie, sequence-to-sequence model and unified architecture), our COST also demonstrates significant advantages.

\myPara{Attrubute-based Performance.} To further examine the performance of trackers on various challenging tracking attributes, we report evaluation results of different attributes on VL-SOT230. As shown in Fig.~\ref{fig:VLSOT230_attributes_1}, our COST achieves the best evaluation results in 13 tracking attributes compared to other SOTA visual and VL trackers. These results demonstrate the ability of our method to efficiently achieve multi-modal fusion and reasoning for VL tracking. In the tracking attributes of full occlusion and deformation, COST slightly underperforms the visual-based baseline algorithm, TransT. After careful analysis, this can be attributed to COST tracking incorrect targets that are semantically similar to the real target after full occlusion or severe deformation. In contrast, TransT may randomly locate the true target region after reappearance or deformation. Furthermore, in scenarios with different brightness and video length (see Fig.~\ref{fig:VLSOT230_attributes_2}), COST consistently achieves the best evaluation results. These outstanding results highlight the effectiveness of using language information to enhance small object tracking, offering valuable insights for future research in multi-modal small object tracking.

\subsection{ Evaluation on VL-SOT270 Dataset}

  VL-SOT270 is a new multi-modal high-speed small object tracking dataset, comprising 270 highly challenging video sequences. Based on the proposed VL-SOT270 dataset, we evaluated 35 deep tracking models, including deep discriminative correlation filter (DCF)-based trackers (\eg, ATOM~\citep{DanelljanBKF19}, DiMP~\citep{BhatDGT19iccv}, KYS~\citep{bhat2020know}, PrDiMP~\citep{danelljan2020probabilistic}), Siamese network-based trackers (\eg, SiamPRN++~\citep{LiWWZXY19}, SiamBAN~\citep{ChenZLZJ20}, SiamCAR~\citep{guo2020siamcar}, SiamGAT~\citep{guo2021graph}), and recent transformer-based trackers (\eg, TransT~\citep{chen2021transformer}, OSTrack~\citep{ye2022joint}, Aba-ViTrack~\citep{li2023adaptive}, GRM~\citep{gao2023generalized}, ARTrack~\citep{wei2023autoregressive}, SeqTrack-B256~\citep{chen2023seqtrack}, ZoomTrack~\citep{kou2023zoomtrack}). To unveil the capability of the language modality in high-speed small object tracking, we compare six recent VL trackers (\ie, VLT\_SCAR~\citep{guo2022divert}, VLT\_TT~\citep{guo2022divert}, JointNLT~\citep{zhou2023joint}, MMTrack~\citep{zheng2023towards}, CiteTracker-256~\citep{li2023citetracker}, UVLTrack~\citep{ma2024unifying}). 

\begin{figure*}[t]
  \centering
  \includegraphics[width=1.0\linewidth]{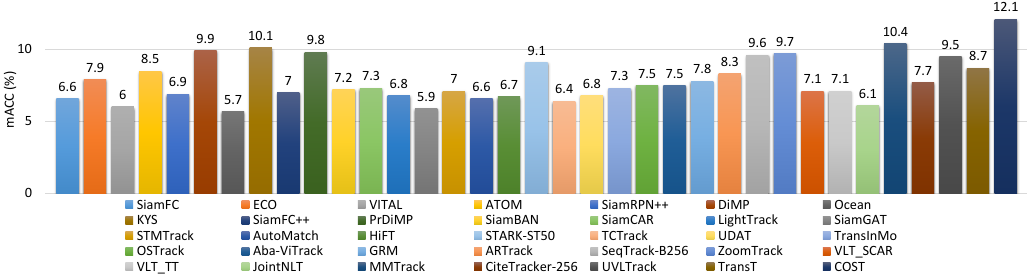} 
  \caption{ Evaluation of 35 deep trackers on VL-SOT270 using mACC score. High-speed small objects result in poor tracking performance for existing methods. Best viewed in color.}
   \label{fig:mACC_VL_SOT270}
\end{figure*}

\begin{figure*}[ht]
\centering
\subfloat{\includegraphics[width =0.5\linewidth]{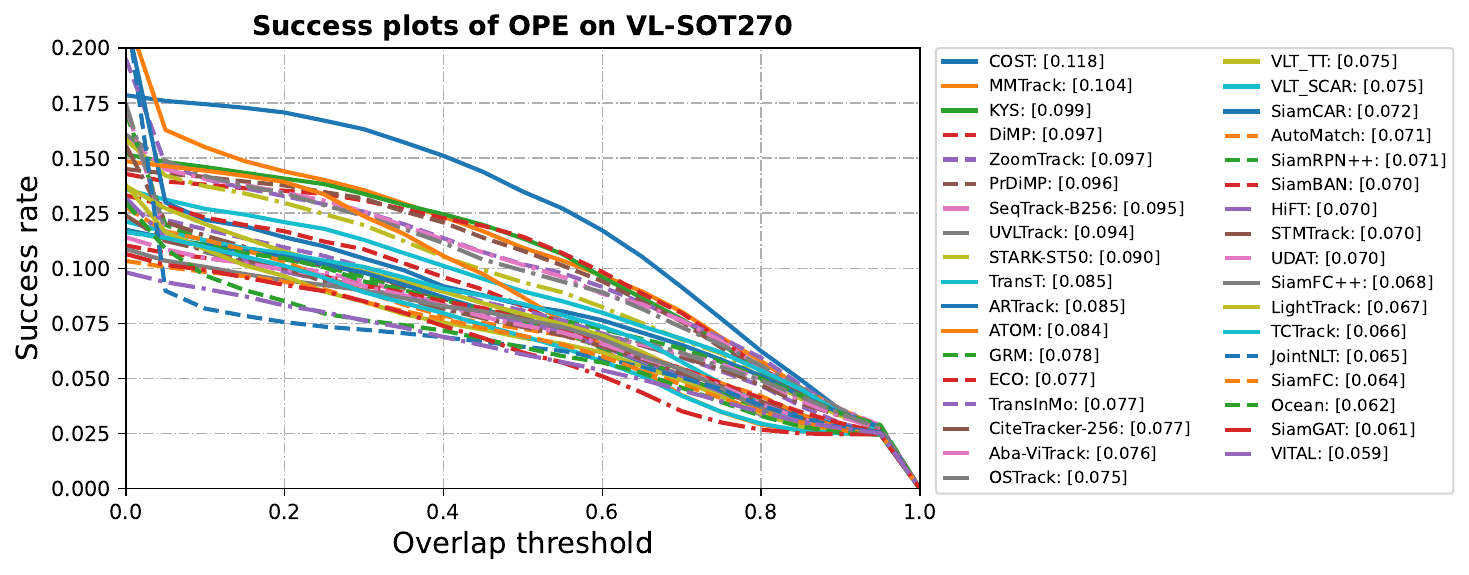}}
~\subfloat{\includegraphics[width =0.5\linewidth]{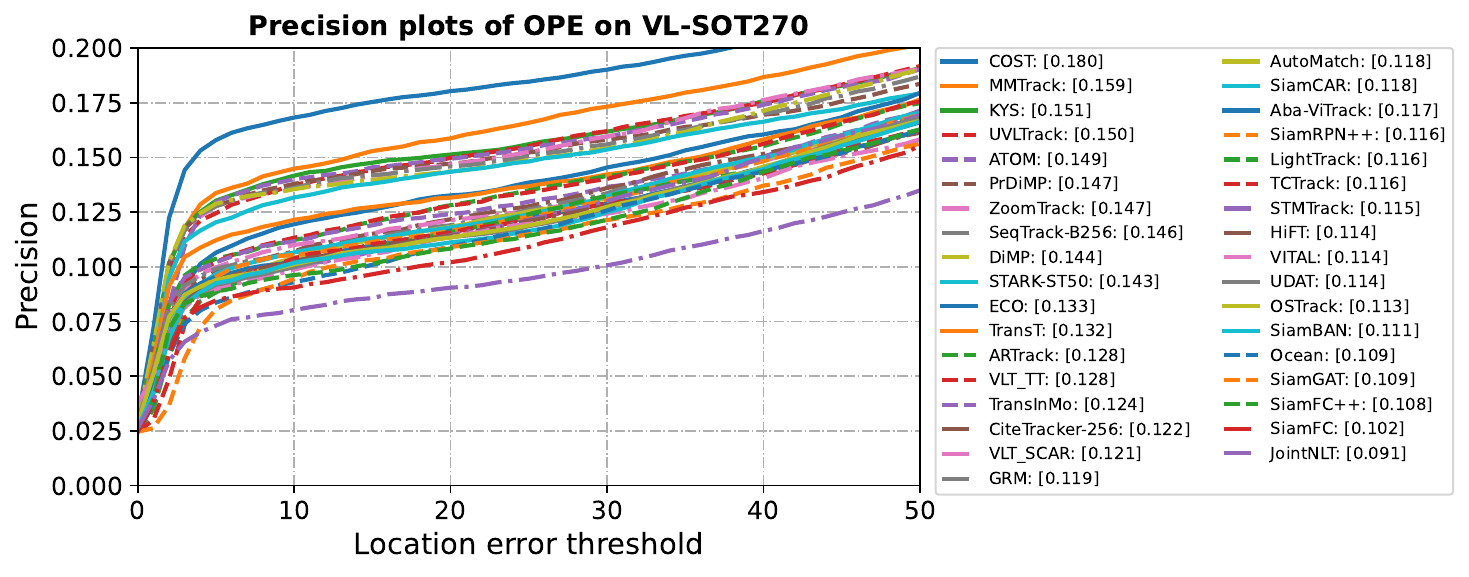}}\\
\subfloat{\includegraphics[width =0.5\linewidth]{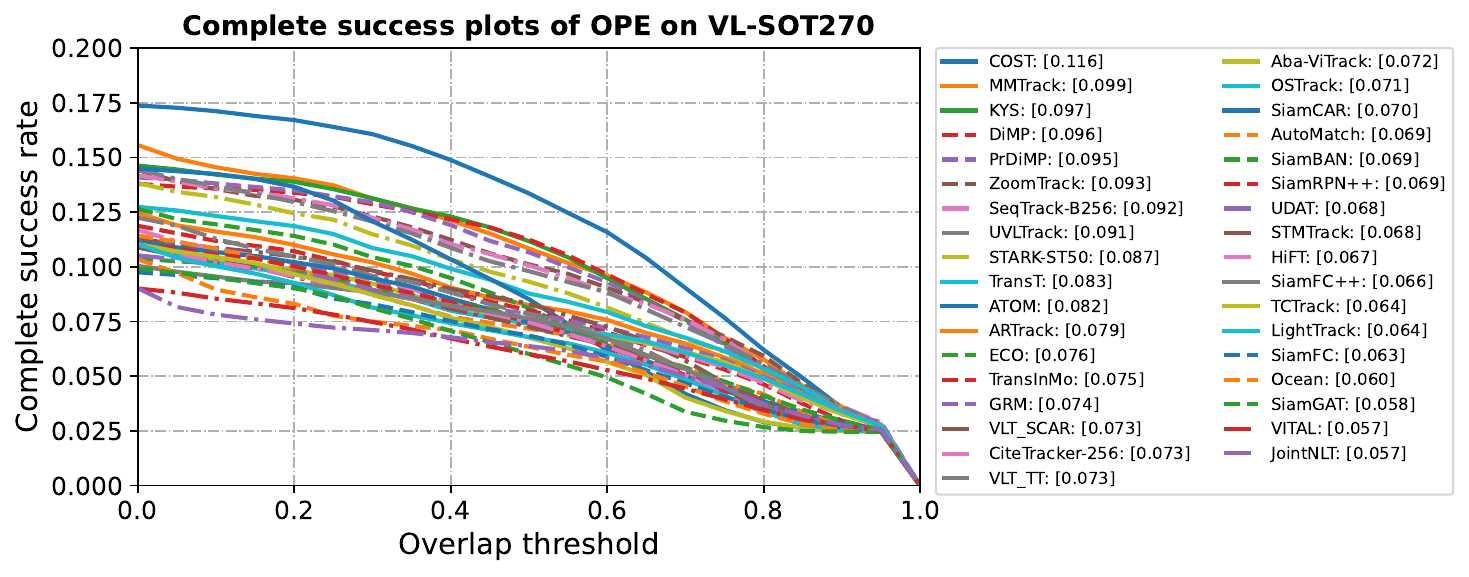}}
~\subfloat{\includegraphics[width =0.5\linewidth]{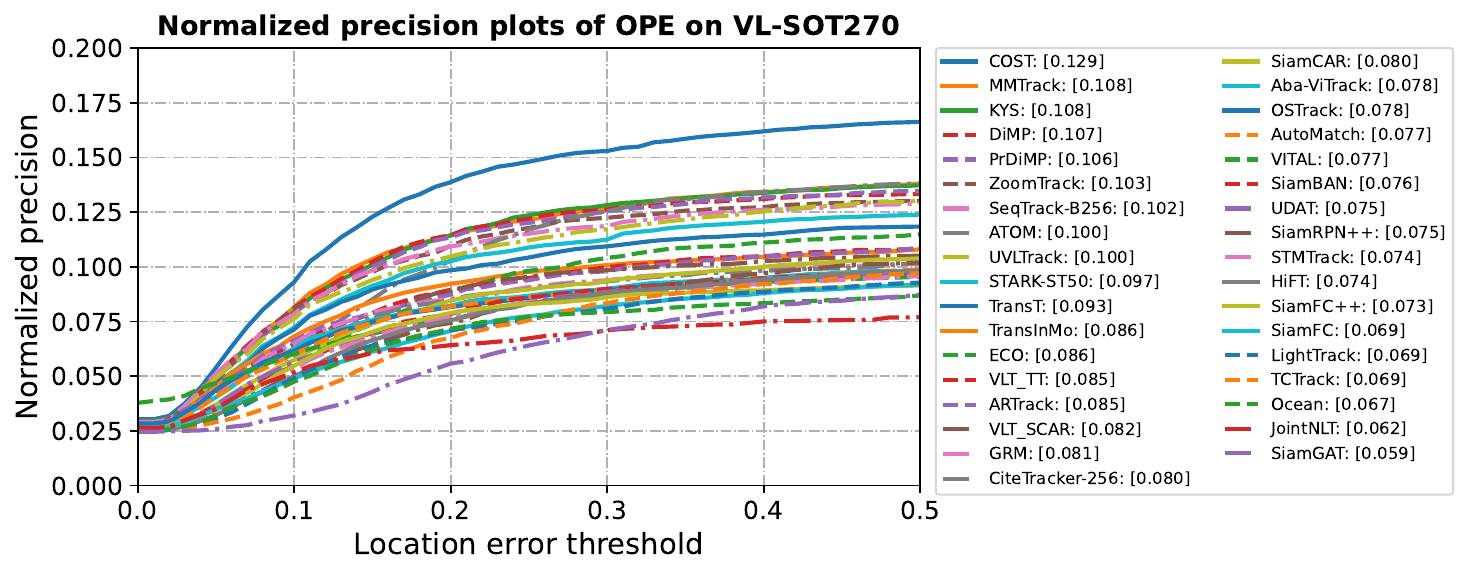}}
\caption{ Evaluation of 35 deep trackers on VL-SOT270 using AUC, $P$, cAUC, and $P_{\rm norm}$ scores. Best viewed in color with zooming in.}
\label{fig:overall_results_on_VLSOT270}
\end{figure*}

\begin{table*}[t]
\footnotesize
\centering
\caption{
Performance comparison on four existing VL tracking benchmarks (\ie, LaSOT, LaSOT\_Ext, OTB99-L, and TNL2K) with visual-based and VL-based trackers. TransT is the baseline of the proposed COST. The top two results are marked in \textbf{bold} and \underline{underline}. ``-'' indicates that the corresponding item was not reported in the original paper.}
\label{tab:four_datasets}
\setlength{\tabcolsep}{2.7mm}{
\begin{tabular}{llccccccccc}
\toprule
    \multicolumn{1}{l}{\multirow{2}[1]{*}{Type}} & \multicolumn{1}{l}{\multirow{2}[1]{*}{Method}} &
	\multicolumn{3}{c}{LaSOT} & \multicolumn{2}{c}{LaSOT\_Ext} &  \multicolumn{2}{c}{{OTB99-L}} & \multicolumn{2}{c}{{TNL2K}} \\

    \cmidrule(r){3-5} \cmidrule(r){6-7} \cmidrule(r){8-9} \cmidrule(r){10-11}
	\multicolumn{1}{c}{} & & AUC (\%)  &  $P$ (\%) & $P_{norm}$ (\%) &  AUC (\%) & $P$ (\%) &  AUC (\%)  & $P$ (\%)  & AUC (\%) & $P$ (\%) \\
 
\midrule

\multirow{14}{*}{Visual-based} 

 & SiamFC~\citep{bertinetto2016fully}   & 33.6 & 33.9 & 42.0 & 23.0 & 26.9 & 58.7 & 79.2 & 29.5 & 28.6\\

 & ECO~\citep{danelljan2017eco}   & 32.4 & 30.1 & 33.8 & 22.0 & 24.0 & - & - & 32.6 & 31.7 \\

 & PrDiMP~\citep{danelljan2020probabilistic}  & 59.8  & 60.8 & 68.4 & - & -  & 69.5 & 89.5 & 47.0 & 45.9 \\

 & AutoMatch~\citep{zhang2021learn}   & 58.3  & 59.9 & - & 37.6 & 43.0  & {71.6}  & {93.2} & 47.2 & 43.5 \\

 & Ocean~\citep{zhang2020ocean}     & 56.0 & 56.6 & 65.1  & - &  - & 68.0 & 92.1 &  38.4 & 37.7 \\

 & KYS~\citep{bhat2020know}   & 55.4 & - & 63.3 & - & -  & - & - & 44.9 & 43.5 \\

 & ATOM~\citep{danelljan2019atom}  & 51.5 & 50.5 & 57.6 & 37.6 & 43.0 & 67.6 & 82.4 & 40.1 & 39.2 \\

 & SiamRPN++~\citep{li2019siamrpn++} & 49.6 & 49.1 & 56.9 & 34.0 & 39.6 & 63.8 & 82.6 & 41.3 & 41.2 \\

 & GlobalTrack~\citep{huang2020globaltrack}  & 51.7 & 52.8 & 59.7 &  35.6 & 41.1 & - & - & 40.5 & 38.6  \\

 & SiamCAR~\citep{guo2020siamcar}   & 50.7 & 51.0 & 60.0 & 33.9 &  41.0  & 68.8 & 89.1 & 35.3 & 38.4 \\

 & TrDiMP~\citep{wang2021transformer}    & 63.9 & 66.3 & - & - & - & 70.5 & {92.5} & - & -\\

 & SiamRCNN~\citep{voigtlaender2020siam}   & 64.8  & 68.4 & 72.2 & - & - & 70.0 & 89.4 & {52.3} & {52.8} \\

 & SimTrack-B/32~\citep{chen2022backbone}   & {66.2} & - & {76.1} & - & - & - & - & 51.1 & 48.1 \\
 
 & STARK-ST50~\citep{yan2021learning}   & {66.4} & {71.2} & {76.3} & {47.8} & {55.1} & 69.6 & 91.4 & - & - \\

 & OSTrack~\citep{ye2022joint}   & {69.1} & \textbf{75.2} & {78.7} & {47.4} & {53.3} & 70.6 & {92.1} & {54.3}  & {56.3}   \\
 
\midrule

\multirow{6}{*}{VL-based} & SNLT~\citep{feng2021siamese}  & 54.0 & 57.6 & 63.6 & 26.2 & 30.0 & 66.6 & 80.4 & 27.6 & 41.9  \\

 & AdaSwitcher~\citep{wang2021towards}   & 51.2 & 55.2 & - & - & -  & 68.2 &  88.1 & 41.7 & 42.0 \\

 & CapsuleTNL~\citep{ma2021capsule}  & 61.5 & 63.3 & - & - & - & 71.1 &  92.4 & - & -  \\

 & VLT\_SCAR~\citep{guo2022divert}  & 63.9 & 67.9 & 73.3 & 44.7 & 51.6  & 73.9 &  89.8 & 49.8 & 51.0 \\
 
 & VLT\_TT~\citep{guo2022divert}  & \underline{67.3}  & {72.1} & \underline{77.6} & \underline{48.4} & \underline{55.9} & \underline{76.4} & \underline{93.1} & 53.1 & 53.3  \\

& JointNLT~\citep{zhou2023joint} & 60.4 & 63.6 & - & - & -  & 65.3 &  85.6 & \underline{56.9} & \underline{58.1}  \\

\hline 
Visual-based  & TransT~\citep{chen2021transformer}   & 64.9 & 69.0 & 73.8 & 44.8 & 52.5  & {70.8} & 91.2 & 50.7 & 51.7  \\
VL-based & \textbf{COST (Ours) } & \textbf{69.2}  & \underline{74.6} & \textbf{79.3} & \textbf{52.0} & \textbf{59.3} & \textbf{77.3} & \textbf{94.5} & \textbf{57.5} & \textbf{58.6}  \\

 
  &  Gain & \textcolor{darkgreen}{+4.3\%} &  \textcolor{darkgreen}{+5.6\%} & \textcolor{darkgreen}{+5.5\% } & \textcolor{darkgreen}{+7.2\%} & \textcolor{darkgreen}{+6.8\%} & \textcolor{darkgreen}{+6.5\% } & \textcolor{darkgreen}{+3.3\%} & \textcolor{darkgreen}{+6.8\%} & \textcolor{darkgreen}{+6.9\%}\\
\bottomrule
\end{tabular}
}
\end{table*}

The benchmark results are presented in Fig.~\ref{fig:mACC_VL_SOT270} and Fig.~\ref{fig:overall_results_on_VLSOT270}. Our observations are as follows: \textbf{1)} The top 3 trackers are COST, MMTrack, and KYS. Both our COST and MMTrack are VL tracking models, demonstrating that the use of language information indeed helps enhance the performance of high-speed small object tracking. KYS is a DCF-based method, but by leveraging contextual information to model the appearance model, it achieves improved robustness and accuracy, highlighting the importance of context for small object tracking. \textbf{2)} High-speed small object tracking poses significant challenges for existing methods. For instance, comparing the mACC scores on VL-SOT230 and VL-SOT, the latter shows a drop of approximately 20\% as shown in Fig.~\ref{fig:mACC_VL_SOT270}. We are surprised to find that only three tracking algorithms (COST, MMTrack, and KYS) achieve mACC scores exceeding 10\%, specifically 12.1\%, 10.4\%, and 10.1\%. Other SOTA tracking methods, such as OSTrack, GRM, ARTrack, SeqTrack-B256, ZoomTrack, VLT\_TT, JointNLT, CiteTracker-256, and UVLTrack, all perform poorly on VL-SOT270. We believe these evaluation results fully demonstrate the significant value of our dataset for small object tracking and the entire tracking community. Based on the dataset we proposed, researchers have large room to develop advanced trackers. \textbf{3)} Compared to the baseline tracker TransT, the proposed COST shows improvements in mACC, AUC, $P$, cAUC, and $P_{\rm norm}$ scores by 3.4\%, 3.3\%, 4.8\%, 3.3\%, and 3.6\%, respectively. This highlights the superiority of our transformer-based one-stage fusion framework for small object tracking. We will further validate the generalization capability of our method on five generic VL tracking datasets in Section~\ref{sec:existing_datasets}.

\begin{figure}[t]
  \centering
\includegraphics[width=1.0\linewidth]{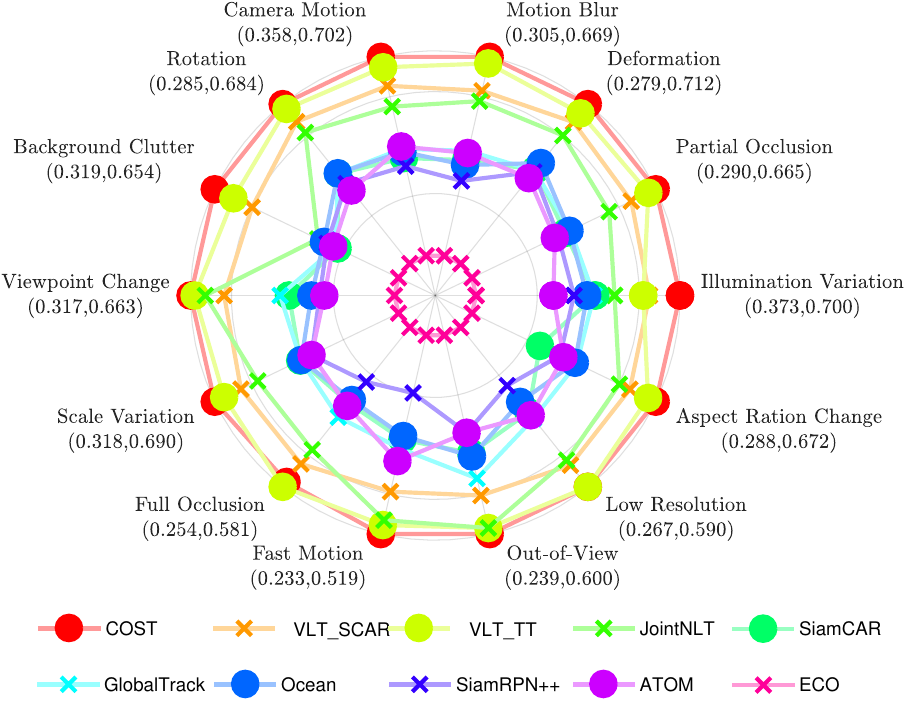}
  \caption{Comparison of normalized AUC scores for different object attributes evaluated on the LaSOT test set.}
   \label{fig:LaSOT_Attribute}
\end{figure}

\begin{figure}[t]
\centering
\subfloat{\includegraphics[width =0.5\columnwidth]{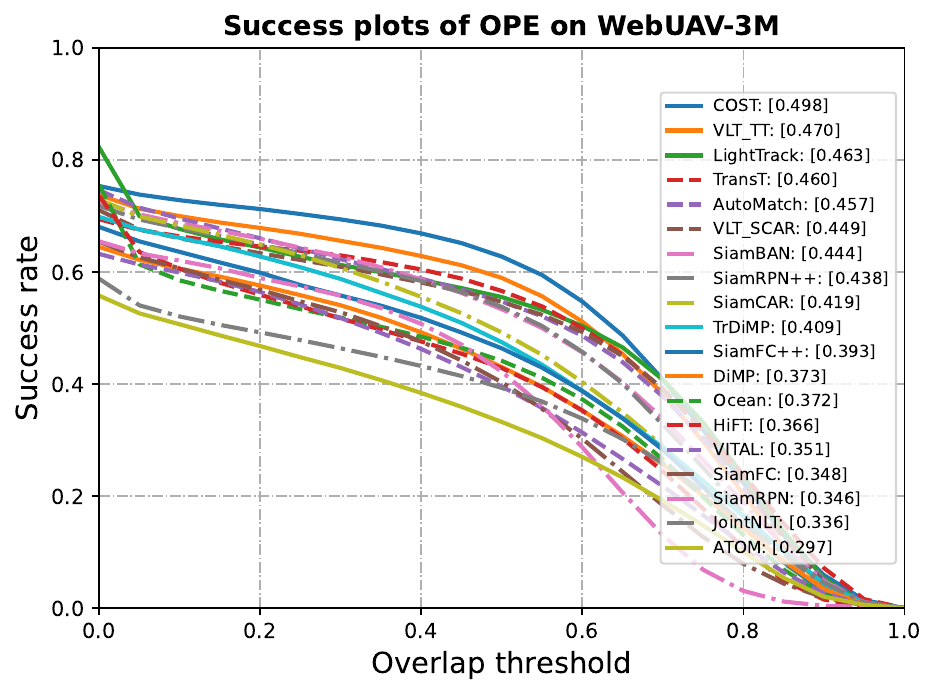}}
~\subfloat{\includegraphics[width =0.5\columnwidth]{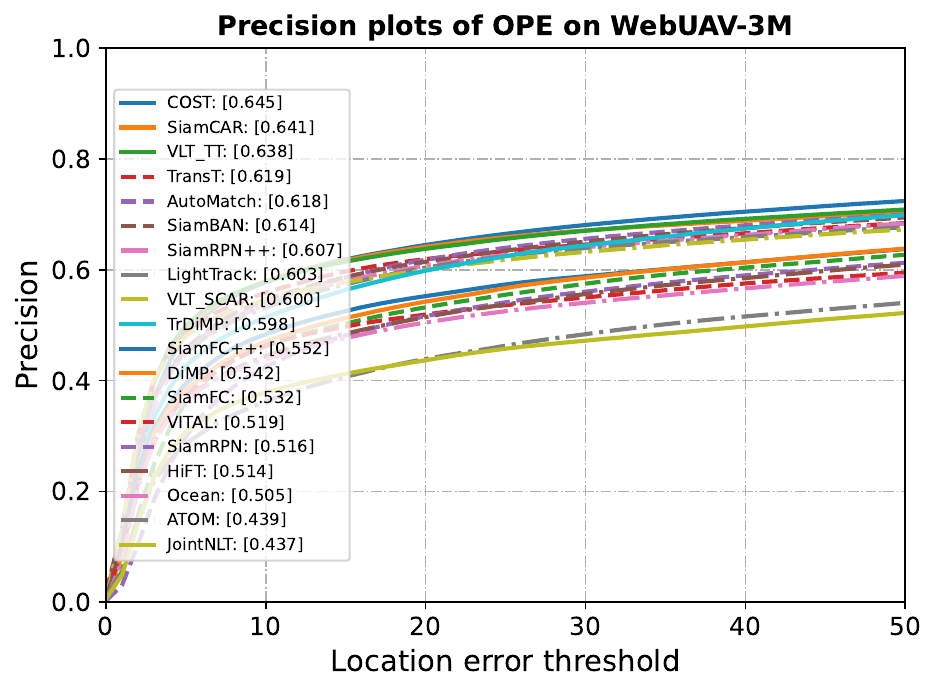}}\\
\vspace{-0.3cm}
\subfloat{\includegraphics[width =0.5\columnwidth]{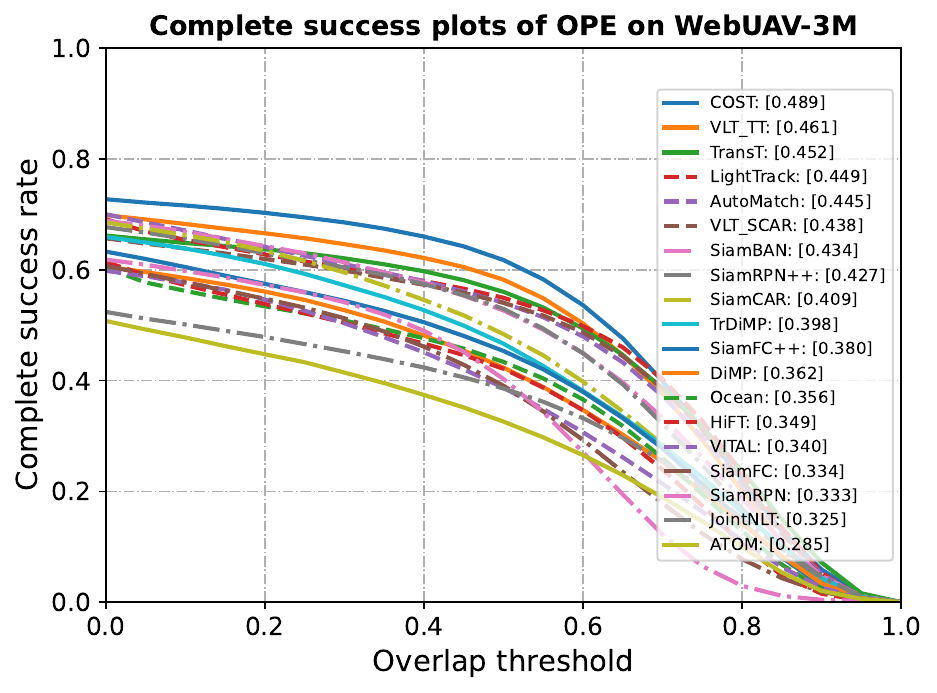}}
~\subfloat{\includegraphics[width =0.5\columnwidth]{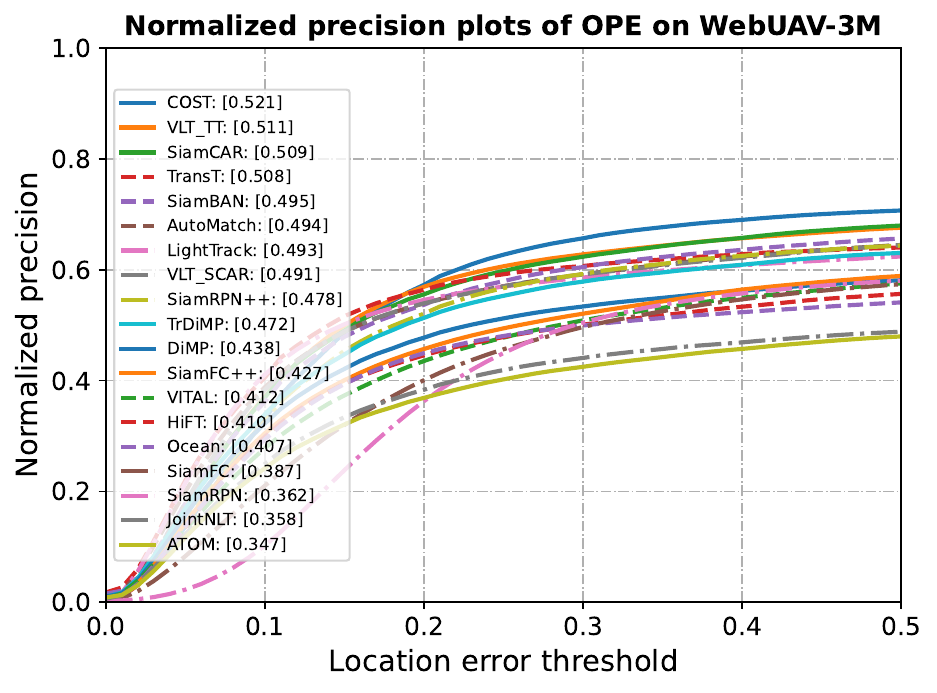}}
\caption{Precision, success, normalized precision, and complete success plots on the WebUAV-3M test set. Best viewed by zooming in.}
\label{fig:WebUAV-3M_test_set}
\end{figure}

\subsection{Generalization Evaluation on Existing VL Tracking Datasets}
\label{sec:existing_datasets}
To validate the generalization capability of the proposed method across different tracking scenarios, from small object tracking to generic object tracking, we conduct comprehensive comparisons between COST and numerous advanced visual trackers (\eg, SiamFC~\citep{bertinetto2016fully}, SiamRPN++~\citep{li2019siamrpn++}, ECO~\citep{danelljan2017eco},  PrDiMP~\citep{danelljan2020probabilistic},
TransT~\citep{chen2021transformer}, TrDiMP~\citep{wang2021transformer}, 
STARK-ST50~\citep{yan2021learning},
SimTrack-B/32~\citep{chen2022backbone},
and OSTrack~\citep{ye2022joint}) and VL trackers (\eg, SNLT~\citep{feng2021siamese}, VLT\_SCAR~\citep{guo2022divert}, VLT\_TT~\citep{guo2022divert}, JointNLT~\citep{zhou2023joint}) on five popular VL tracking benchmarks (\ie, LaSOT~\citep{fan2019lasot}, LaSOT\_Ext~\citep{fan2021lasot}, OTB99-L~\citep{li2017tracking}, TNL2K~\citep{wang2021towards}, WebUAV-3M~\citep{zhang2022webuav}).

\myPara{LaSOT.} LaSOT~\citep{fan2019lasot} is a large-scale VL tracking dataset, which provides high-quality bounding box annotations and language descriptions. It contains 1,120 videos for the training set and 280 videos for the test set. Tab.~\ref{tab:four_datasets} presents the overall performance of COST and other SOTA visual and VL trackers on the LaSOT test set. COST obtains the best AUC (69.2\%), $P$ (74.6\%), and $P_{norm}$ (79.3\%) scores among VL trackers. Specifically, compared with the recent SOTA VL tracker VLT\_TT~\citep{guo2022divert}. COST is comparable to the SOTA visual tracker OSTrack~\citep{ye2022joint}. The main difference between our COST and OSTrack is that the latter is a one-stream framework, which adopts a ViT model pre-trained with MAE~\citep{he2022masked} for joint feature extraction and relation modeling. However, our method (\ie, CVLF module) trained from scratch achieves substantial performance gains on multiple VL tracking datasets, implying its excellent generalization ability. In Fig.~\ref{fig:LaSOT_Attribute}, we report an attribute-based comparison of ten representative trackers on the LaSOT test set, indicating that the proposed COST outperforms other trackers on 13 attributes. In the full occlusion scene, VLT\_TT performs slightly better than our method. This is mainly because VLT\_TT uses additional attribute word annotations~\citep{guo2022divert}, which can provide more accurate information than ambiguous language descriptions.

\myPara{LaSOT\_Ext.} LaSOT\_Ext~\citep{fan2021lasot} dataset is an extended version of LaSOT, containing 15 object classes with 150 challenging long videos. As reported in Tab.~\ref{tab:four_datasets}, COST outperforms the transformer-based visual trackers STARK-ST50~\citep{yan2021learning} and TransT~\citep{chen2021transformer} by 4.2\% and 7.2\% in terms of AUC score, respectively. Furthermore, our COST based on one-stage multi-modal fusion achieves a new SOTA $P$ score of 59.3\% among VL trackers on the LaSOT\_Ext dataset, surpassing VLT\_TT~\citep{guo2022divert} and SNLT~\citep{feng2021siamese} by 3.4\% and 29.3\%, respectively (see Tab.~\ref{tab:four_datasets}).

\myPara{OTB99-L.} OTB99-L is an early VL tracking dataset annotated by Li~\etal~\citep{li2017tracking}, which contains 99 videos: 51 videos for training and 48 videos for testing. As shown in Tab.~\ref{tab:four_datasets}, COST obtains 77.3\% and 94.5\% in terms of AUC and $P$ scores, respectively, surpassing all compared trackers. Compared with the latest VL tracker JointNLT~\citep{zhou2023joint}, the performance gains are 12.0\% and 8.9\% in terms of AUC and $P$, respectively.

\begin{table}[t]
\footnotesize
  \centering
  \caption{ Ablation study of various components in our method, including the linguistic branch (LB), learnable [OBJ] token, CoA, and visual-linguistic transformer (VLT). Note that removing the linguistic branch degrades our method into the visual-based baseline tracker TransT. ``w/o'' represents ``without''.}
  \setlength{\tabcolsep}{1.8mm}{
  \begin{tabular}{lccccc}
   \toprule
    
    \multicolumn{1}{l}{\multirow{2}[1]{*}{Method}} &         
	\multicolumn{3}{c}{LaSOT} &         
	\multicolumn{2}{c}{LaSOT\_Ext}\\
    \cmidrule(r){2-4} \cmidrule(r){5-6}  \multicolumn{1}{c}{} &   AUC (\%) & $P_{norm}$ (\%) & $P$ (\%) &   AUC (\%)  &  $P$(\%)  \\
   
    \midrule
    w/o LB (baseline) &  55.2 &  64.8 & 55.1 & 36.6 & 38.8 \\

    w/o [OBJ] token &  57.6	& 66.9	& 58.8	& 40.7	& 44.1 \\
    
    w/o CoA &  56.8	&  66.2	&  57.5	&  39.6	&  42.7\\  
    
    w/o VLT & 57.0& 	66.5& 	58.4	& 40.1	& 43.6\\

    COST &  \textbf{58.6} &  \textbf{67.1} & \textbf{60.1}  & \textbf{41.8} & \textbf{45.2} \\ 

   \bottomrule
  \end{tabular}
  }
  \label{tab:ablation_study}
\end{table}

\begin{table}[t]
\footnotesize
  \centering
    \caption{ Evaluation results of our method with different language models on the LaSOT test set. The trainable parameters of language models are listed.}
  \setlength{\tabcolsep}{1.5mm}{
  \begin{tabular}{lcccc}
    \toprule 
    Method &   Parameters (M) & AUC (\%) & $P_{norm}$ (\%) & $P$ (\%)  \\
    \midrule

    Baseline &  0  & 55.2 &  64.8 & 55.1  \\
    
    COST-GloVe & 0  &  56.1 &  65.7 & 55.7  \\
    
    COST-BERT$_{\rm BASE}$ & 110 
 & \textbf{56.8} &  \textbf{66.2} & 57.5 \\
    
    COST-BERT$_{\rm LARGE}$ & 340 & 56.7 &  66.1 & \textbf{58.0}  \\
    \bottomrule
  \end{tabular}
  }
  \label{tab:language_embedding}
\end{table}

\begin{table}[t]
\footnotesize
  \centering
    \caption{Evaluation results of three visual tracking models with different language models on the LaSOT test set. }
  \setlength{\tabcolsep}{3.0mm}{
  \begin{tabular}{llcc}
       \toprule 
    Tracking Model & Language Model & AUC (\%) & $P_{norm}$ (\%)   \\
    \midrule
    \multirow{3}{*}{SiamRPN~\citep{li2018high}} & None  & 42.2 &  50.9 \\
     & GloVe & 43.6 &  52.8 \\
     & BERT$_{\rm BASE}$ & \textbf{46.0} &  \textbf{54.6} \\
    \midrule
    \multirow{3}{*}{SiamRPN++~\citep{li2019siamrpn++}} & None & 48.9 &  58.0 \\
     & GloVe & 49.9 &  59.4 \\
     & BERT$_{\rm BASE}$ & \textbf{54.0} &  \textbf{63.6} \\
    \midrule

    \multirow{3}{*}{TransT~\citep{chen2021transformer}}   & None & 55.2  &  64.8   \\
       & GloVe & 56.1 &  65.7 \\
       & BERT$_{\rm BASE}$ & \textbf{56.8} &  \textbf{66.2} \\
    \bottomrule
  \end{tabular}
  }
  \label{tab:tracking_models}
\end{table}

\begin{table}[t]
\footnotesize
  \centering
  \caption{ Comparison with different transformer encoder layers of visual-linguistic transformer on the LaSOT test set.}
  \setlength{\tabcolsep}{1.55mm}{
  \begin{tabular}{lccccc}
    \toprule 
    Method & Layers (L) &  Parameters (M) & AUC (\%)& $P_{norm}$ (\%) & $P$ (\%)\\
    \midrule

    COST & 2 & 141.3 & 54.3 &  63.6 & 52.3 \\
    COST & 4 & 143.9 &  54.6 &  64.5 & 52.4 \\
    COST & 6 & 146.5 & \textbf{56.8} & \textbf{66.2} & \textbf{57.5} \\
    \bottomrule
  \end{tabular}
  }
\label{tab:transformer_encoder_decoder}
\end{table}

\myPara{TNL2K.} TNL2K~\citep{wang2021towards} is a large-scale dataset for the task of language-initialized tracking, covering a wide range of common challenges in tracking, \eg, adversarial sample and thermal crossover. It consists of 1,300 training videos and 700 test videos. Tab.~\ref{tab:four_datasets} demonstrates that our method achieves SOTA results compared to existing visual tracking algorithms. Furthermore, COST obtains an AUC score of 57.5\%, which is 4.4\% higher than the previous SOTA VL tracker VLT\_TT.

\begin{figure}[t]
  \centering
  \includegraphics[width=1.0\linewidth]{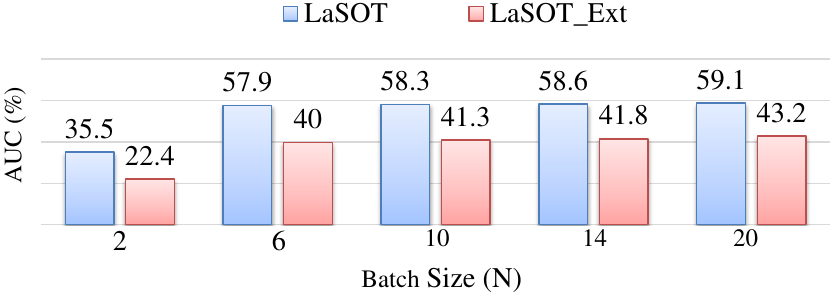} 
  \caption{ Impact of batch size on the LaSOT test set and LaSOT\_Ext dataset.}
   \label{fig:Batch_size}
\end{figure} 

\myPara{WebUAV-3M.} WebUAV-3M~\citep{zhang2022webuav} is the latest million-scale tracking dataset with vision, language, and audio annotations, which contains 4,500 challenging videos: 3,520 for training, 200 for validation, and 780 for testing. We report the precision, success rate, normalized precision, and complete success rate in Fig.~\ref{fig:WebUAV-3M_test_set}. Results demonstrate that COST obtains the best performance compared to other SOTA visual and VL trackers. For instance, compared with the recent SOTA VL tracker VLT\_TT~\citep{guo2022divert}, the performance gains are 2.8\%, 2.8\%, 7.0\%, and 1.0\% in terms of AUC, cAUC, $P$, and $P_{norm}$, respectively, demonstrating the effectiveness of the proposed one-stage multi-modal fusion framework.

\subsection{Ablation Study} 
\label{sec:ablation_study}
To validate the impact of different components, we conduct ablation experiments on two large-scale VL tracking datasets, including LaSOT~\citep{fan2019lasot} and LaSOT\_Ext~\citep{fan2021lasot}. Following~\citep{fan2019lasot,fan2021lasot,zhang2023all}, ablation experiments are trained on the LaSOT training set and evaluated on the LaSOT test set and LaSOT\_Ext dataset.

\myPara{Component-wise Analysis.} We study the impact of each component in our method, including the linguistic branch (LB), CoA, learnable [OBJ] token, and visual-linguistic transformer (VLT) with five variants of the COST. \textbf{1) w/o LB (baseline)}, which solely employs the visual branch to extract visual features from both search and template images, then uses the tracking head to predict target locations. In this configuration, neither the linguistic branch nor VLT is utilized, reducing our method to the baseline tracker. \textbf{2) w/o [OBJ] token}, which replaces the learnable [OBJ] token with zero tensors of identical dimensions for multi-modal learning, then utilizes the learned features for tracking targets. \textbf{3) w/o CoA}, which directly feeds the two modalities with significantly divergent feature distributions into VLT for multi-modal fusion, then performs target prediction using the learned features. \textbf{4) w/o VLT}, which removes the VLT, concatenates the aligned visual and language features with a learnable [OBJ] token before feeding them into the tracking head for target localization. \textbf{5) COST}, our complete model, employs separate visual and linguistic branches to extract modality-specific features, then utilizes contrastive loss as an explicit constraint to align them in the semantic space. The aligned multi-modal features, along with a learnable [OBJ] token, are fed into the VLT to learn unified representations for target state estimation.

 Overall, the ablation study results (see Tab.~\ref{tab:ablation_study}) demonstrate that each component contributes to our method. Our main observations are as follows: \textbf{1)} The performance gaps (55.2\% vs. 58.6\% on LaSOT and 36.6\% vs. 41.8\% on LaSOT\_Ext in terms of AUC score) between the base model (w/o LB) and COST clearly demonstrate the advantage of incorporating linguistic information for tracking. \textbf{2)} With the learnable [OBJ] token, COST achieves performance gains of 1.0\% (from 57.6\% to 58.6\%) and 1.1\% (from 40.7\% to 41.8\%) in terms of AUC score on LaSOT and LaSOT\_Ext, respectively. These improvements validate that the learnable [OBJ] token is beneficial for learning consolidated VL representations as it can enhance multi-modal associations by both visual and linguistic context during training. \textbf{3)} Without using the CoA, COST decreases by 1.8\% (from 58.6\% to 56.8\%) and 2.2\% (from 41.8\% to 39.6\%) in terms of AUC score on LaSOT and LaSOT\_Ext, respectively. This validates the superiority of CoA in significantly facilitating multi-modal fusion and reasoning. \textbf{4)} By comparing our COST with w/o VLT, we can observe that the proposed VLT improves tracking performance by 1.6\% (from 57.0\% to 58.6\%) and 1.7\% (from 40.1\% to 41.8\%) in AUC on LaSOT and LaSOT\_Ext, respectively, demonstrating its effectiveness in learning unified VL representations.

\myPara{Impact of Language Models.} We test COST with different language models, including a word embedding model GloVe~\citep{pennington2014glove} and two transformer-based language models (\ie, BERT$_{\rm BASE}$~\citep{devlin2018bert}, BERT$_{\rm LARGE}$~\citep{devlin2018bert}). As shown in Tab.~\ref{tab:language_embedding}, transformer-based language models are better than the word embedding model. To balance cost and performance, we adopt BERT$_{\rm BASE}$ as our default setting.

\myPara{Impact of Visual Tracking Models.} To verify the proposed method can be generalized to different tracking frameworks. We compare our retrained transformer-based tracking model~\citep{chen2021transformer} with two popular CNN-based tracking models (\ie, SiamRPN~\citep{li2018high}, SiamRPN++~\citep{li2019siamrpn++}) using different language models. Comparisons of these trackers are shown in Tab.~\ref{tab:tracking_models}. Results demonstrate that the transformer-based tracking model is superior to CNN-based tracking models. Due to the excellent performance of the transformer-based tracking model, we use it as the default setting for the visual branch.

\myPara{Impact of Transformer Fusion Layers.} The impact of different encoder layers ($L$) of the visual-linguistic transformer is shown in  
Tab.~\ref{tab:transformer_encoder_decoder}. COST achieves a stable performance gain as the number of encoder layers increases. Increasing the number of transformer encoder layers may further improve performance, but result in more parameters. This departs from our motivation of designing a simple yet efficient one-stage transformer-based framework. In this work, we set $L=6$ as the default setting due to it achieves a nice balance of high performance and a reasonable computational load.

 \myPara{{ Impact of Batch Size.}}  We conduct experiments to explore the impact of different batch sizes, with results presented in Fig.~\ref{fig:Batch_size}. Our observations are as follows: \textbf{1)} When the batch size is set to default value 14, our method achieves favorable AUC scores, \ie, 58.6\% on LaSOT and 41.8\% on LaSOT\_Ext. Due to the 24GB memory limitation of the RTX 3090 GPU, we can only use a relatively small batch size (\ie, $N\!\leq\!14$). However, we believe that a relatively small batch size may help mitigate the impact of false negative samples on the tracking model~\citep{chen2021uscl,zhang2022hico}. \textbf{2)} Consistent with popular CL methods~\citep{he2020momentum,oord2018representation}, increasing the batch size can improve performance, but also increase memory usage. As shown in Fig.~\ref{fig:Batch_size}, on a more powerful RTX A6000 GPU with 48GB of memory, increasing the batch size to 20 results in certain gains, \ie, 0.5\% on LaSOT and 1.4\% on LaSOT\_Ext. Thus, adopting a larger batch size to enhance tracking performance may be worthwhile when resources permit.

\begin{figure*}[t]
\centering
\subfloat{\includegraphics[width =0.25\linewidth]{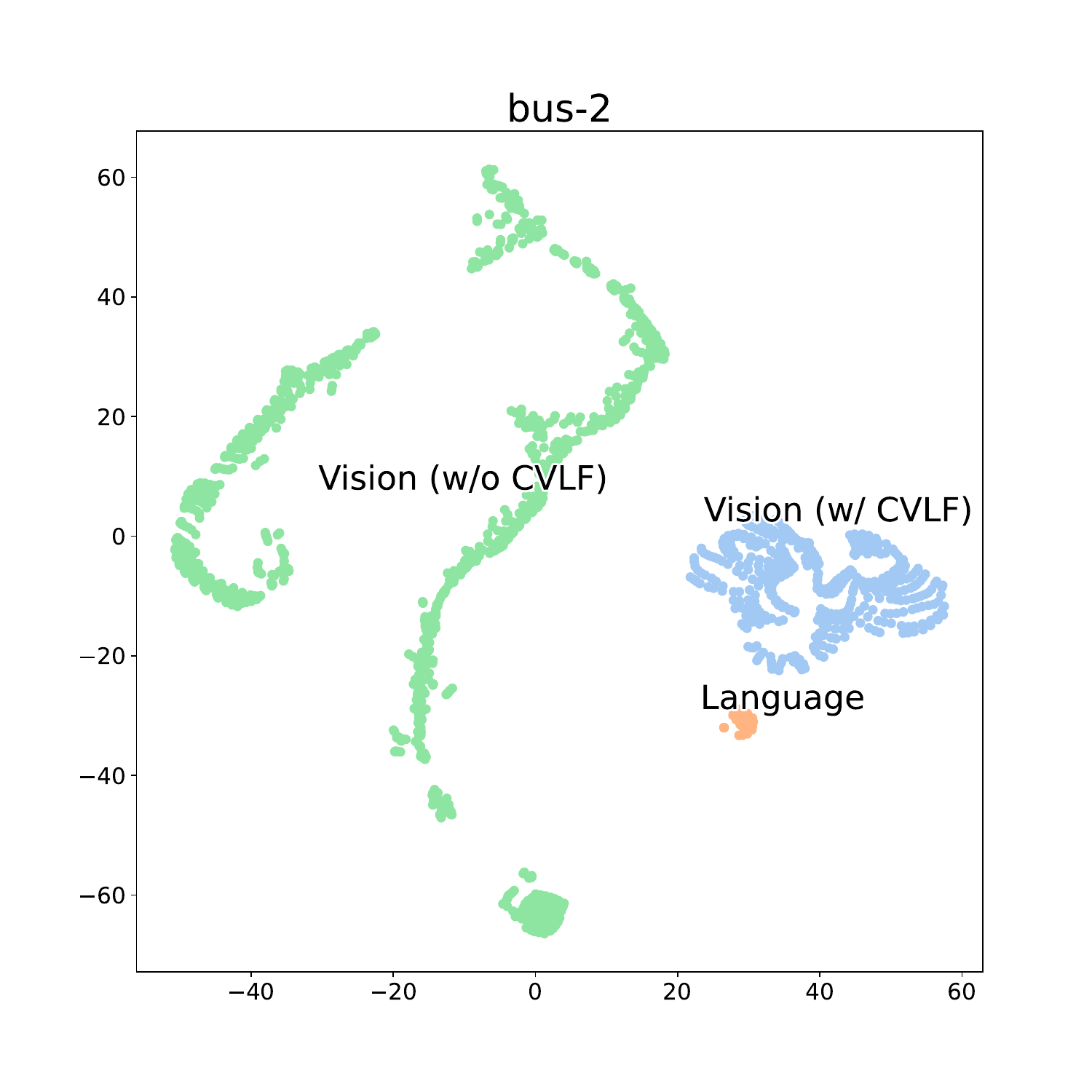}}
\subfloat{\includegraphics[width =0.25\linewidth]{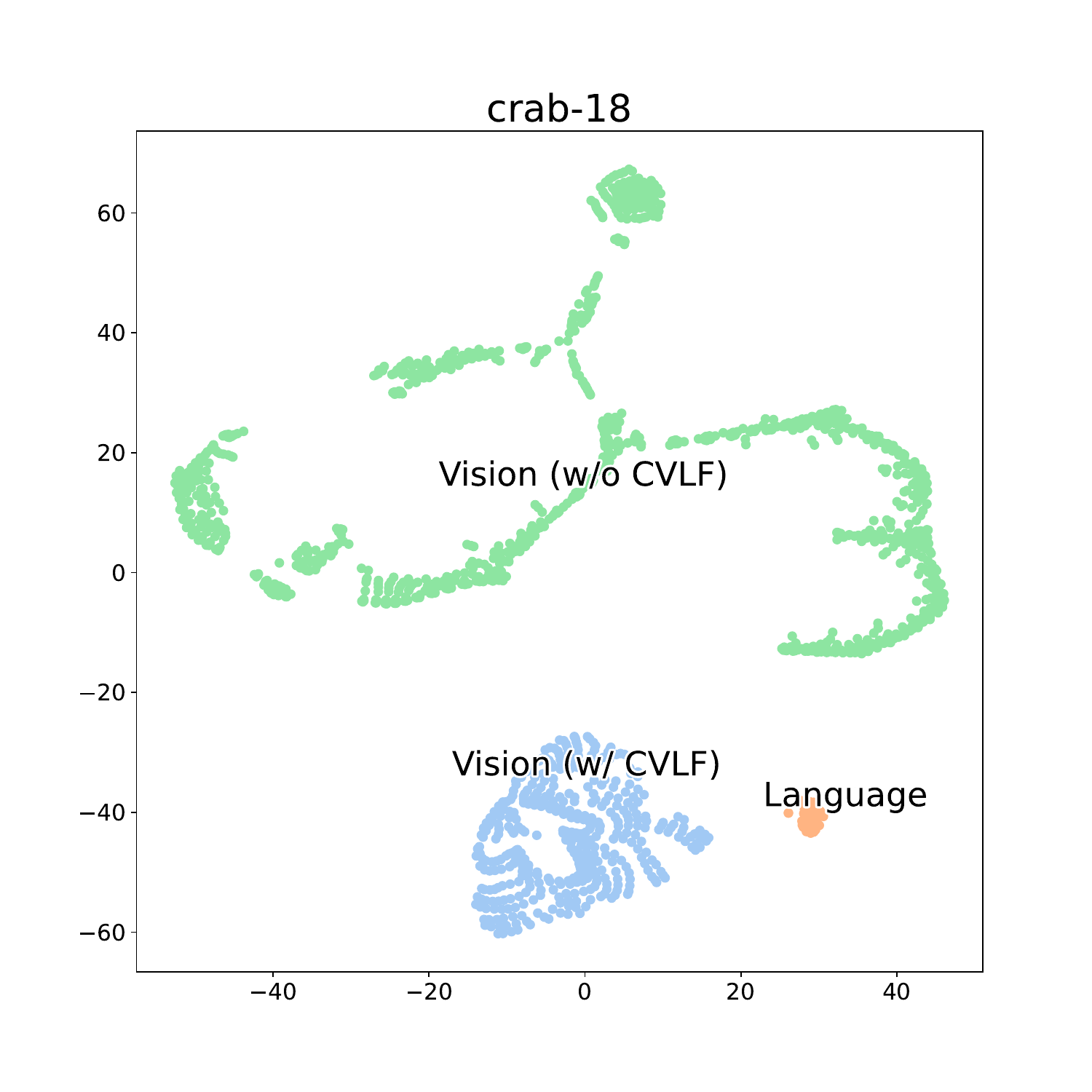}}
\subfloat{\includegraphics[width =0.25\linewidth]{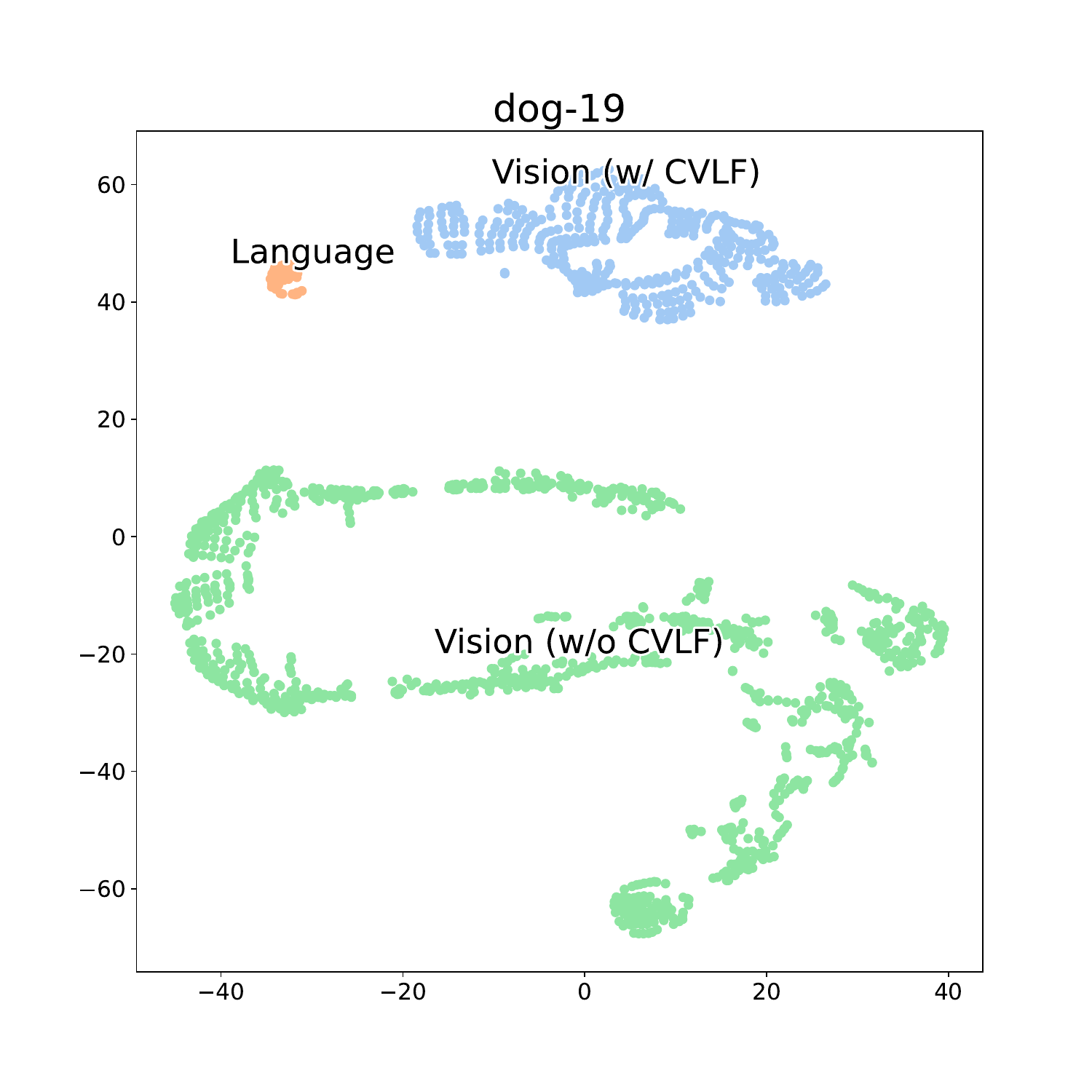}}
\subfloat{\includegraphics[width =0.25\linewidth]{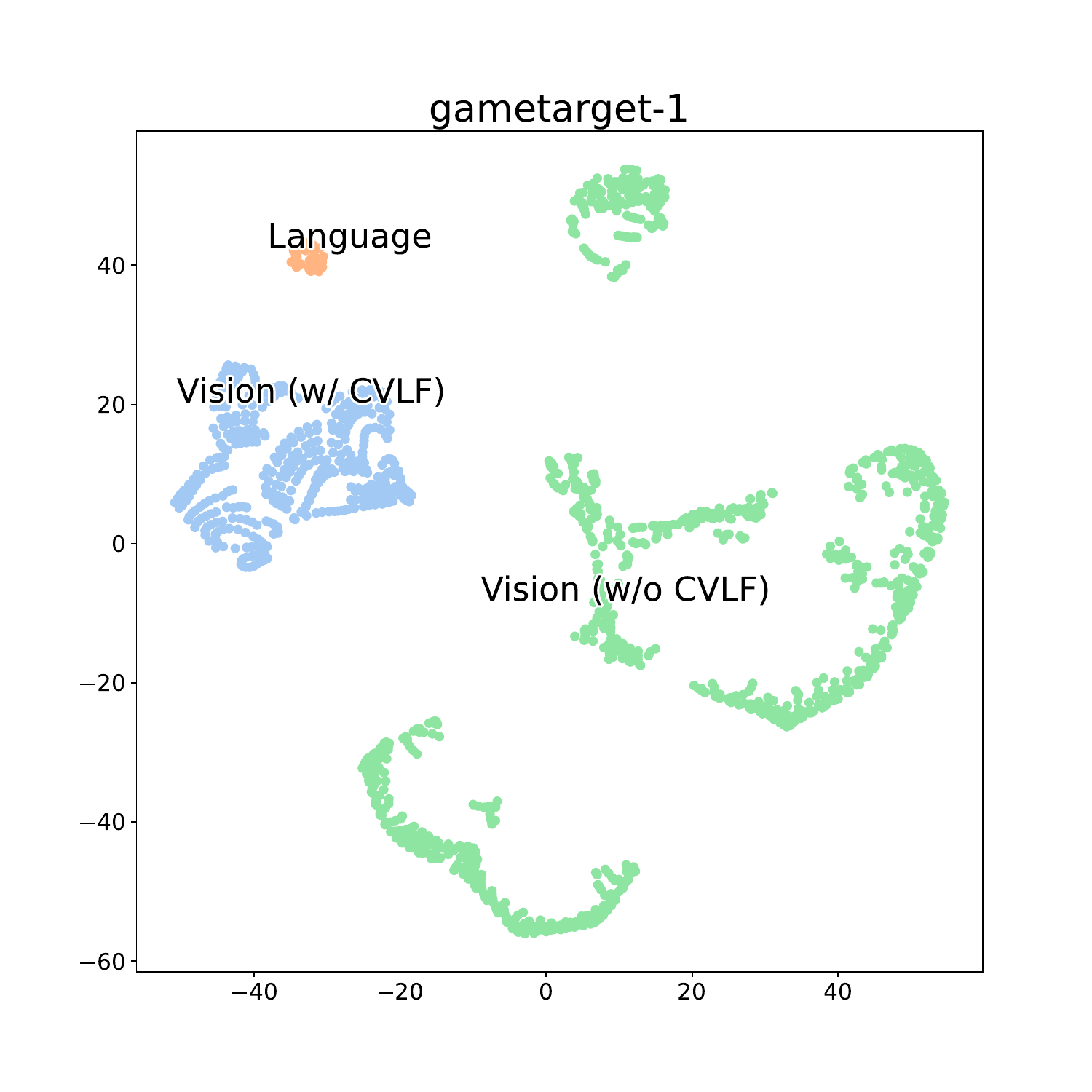}}
\caption{ Visual and linguistic feature distributions visualized by t-SNE~\citep{van2008visualizing} on four challenging video sequences (\ie, \emph{bus-2}, \emph{crab-18}, \emph{dog-19}, and \emph{gametarget-1}). The two trackers are trained without (w/o) and with (w/) the CVLF module, respectively. Our CVLF can effectively align the visual and language features for both \emph{normal-sized and small objects} in the feature space.}
\label{fig:tSNE}
\end{figure*}

\begin{figure*}[t]
  \centering
  \includegraphics[width=1.0\linewidth]{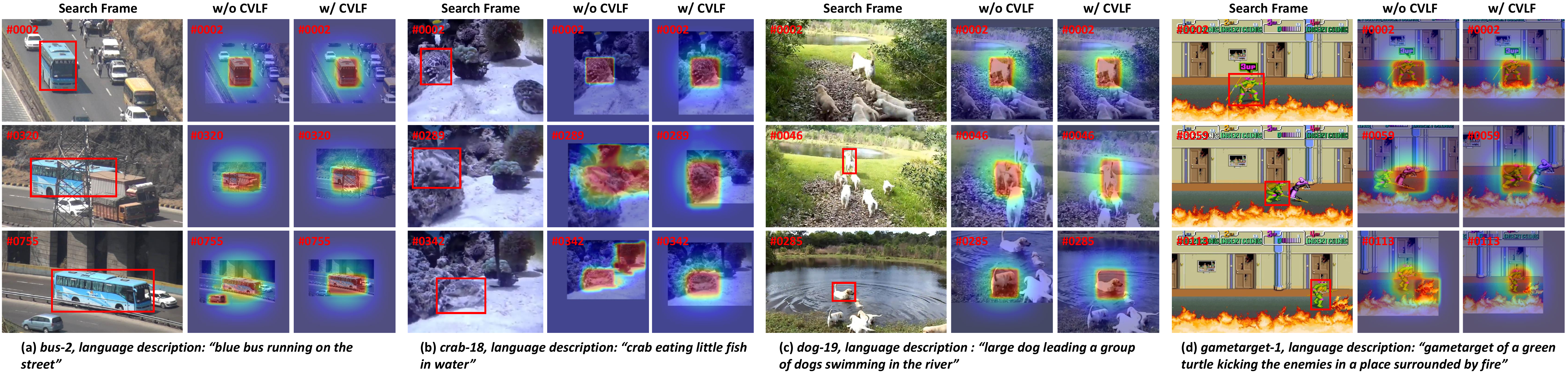}
  \caption{ Visualization of confidence scores of without and with the CVLF module on four challenging video sequences (\ie, \emph{bus-2}, \emph{crab-18}, \emph{dog-19}, and \emph{gametarget-1}). Best viewed by zooming in.}
   \label{fig:transformer_fusion_attention}
\end{figure*}

\subsection{Visualization of Tracking}
 In Fig.~\ref{fig:tSNE}, we provide four visualization examples with objects of different sizes using t-SNE~\citep{van2008visualizing} to verify the ability of our one-stage multi-modal fusion framework to achieve efficient multi-modal representation learning. To this end, we train two trackers on the LaSOT training set without and with the proposed CVLF module, respectively. Without the CVLF module, the tracker only uses the visual branch for tracking. Fig.~\ref{fig:tSNE} shows the distributions of visual features and corresponding language features on four video sequences (\ie, \emph{bus-2}, \emph{crab-18}, \emph{dog-19}, and \emph{gametarget-1}). We can observe that the tracker with the CVLF module significantly reduces the distribution discrepancies between vision and language modalities, \ie, sparse visual features are more concentrated, and features of matched video-language pairs are closer in feature space. We argue this attributes to the proposed CoA enabling great alignment of visual features and language features, as well as the visual-linguistic transformer encouraging learning unified VL representations.

To further show the superiority of our framework in multi-modal representation learning, we visualize the confidence scores of tracking results on search regions. As shown in Fig.~\ref{fig:transformer_fusion_attention}, the target can be consistently tracked even when there are similar background distractors, appearance changes, occlusion, \etc. For dog-19, the language description of the target is \emph{``large dog leading a group of dogs swimming in the river''}. Without the CVLF module to inject and integrate semantic information, the tracker is easily fooled by similar objects (small dogs around). The tracker with the CVLF module is more reliable and has a more concentrated prediction of the target (large dog) than the compared tracker. Similar results can be found from video sequences crab-18, bus-2, and gametarget-1. These results demonstrate that our framework achieves efficient multi-modal fusion and effectively utilizes discriminative semantic information for accurate target localization.

\begin{figure*}[t]
  \centering
\includegraphics[width=1.0\linewidth]{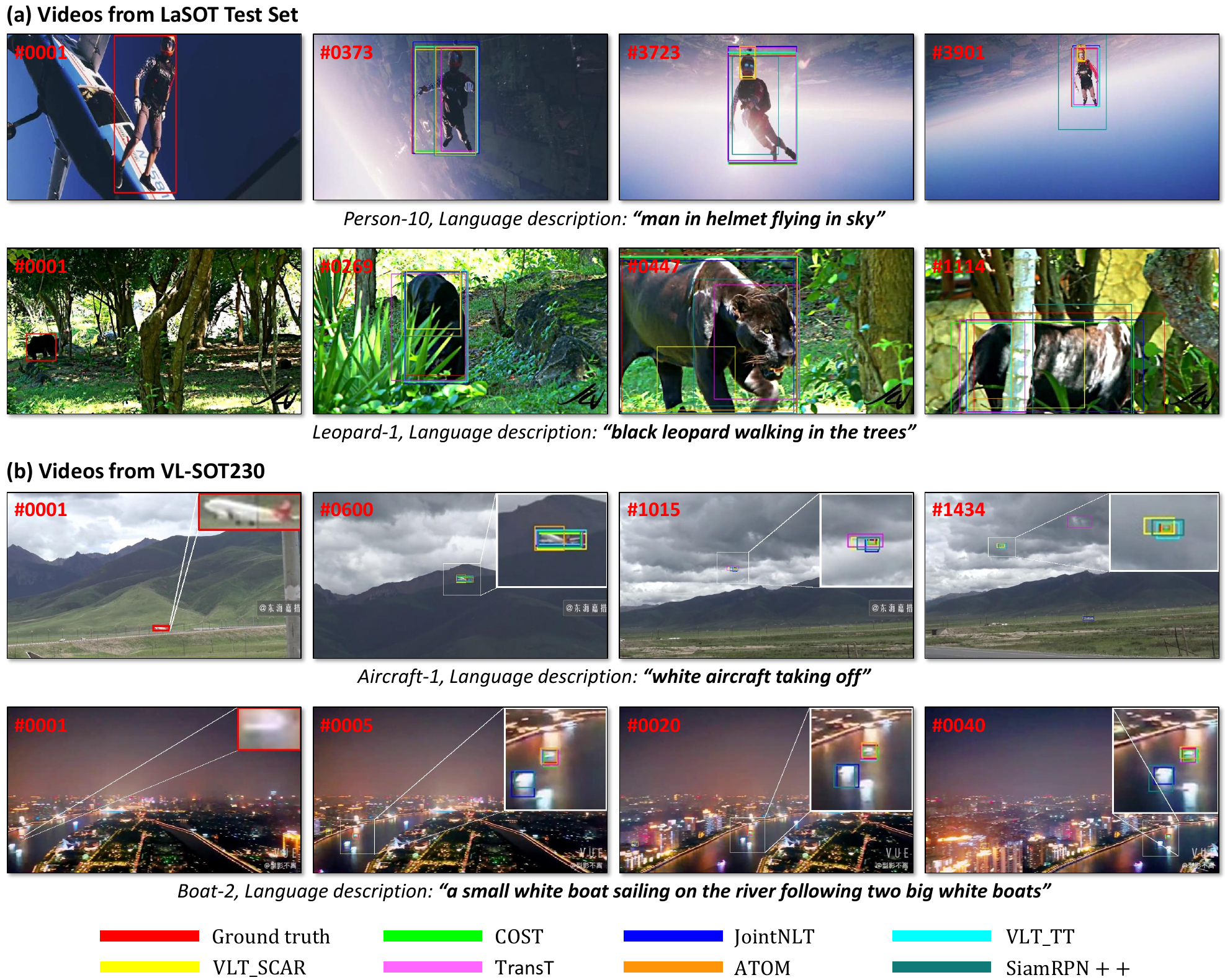}
  \caption{Qualitative comparison of SOTA trackers along with our COST. We selected challenging video sequences from (a) the LaSOT test set and (b) the proposed VL-SOT230. Best viewed in color with zooming in.}
   \label{fig:qualitative_performance}
\end{figure*}

\subsection{Qualitative Performance}
To qualitatively demonstrate the effectiveness of the proposed method, we first visualize the tracking results of our COST and SOTA VL trackers (\ie, JointNLT, VLT\_SCAR, and VLT\_TT) and visual trackers (\ie, ATOM, SiamRPN++, and TransT) on the popular VL tracking dataset LaSOT with two challenging video sequences (see Fig.~\ref{fig:qualitative_performance}(a)). These video sequences usually have normal-sized objects but have complex challenges, such as serious appearance changes, deformation, illumination variations, motion blur, background clutter, and occlusion. Fig.~\ref{fig:qualitative_performance}(a) shows that COST achieves the best performance compared to other SOTAs, demonstrating the effectiveness of the one-stage multi-modal fusion framework in complex environments. Moreover, we found that COST and VLT\_TT outperform the baseline method TransT in localization accuracy on these videos, due to the high-level semantics provided to enhance the unified VL representation. 

Fig.~\ref{fig:qualitative_performance}(b) shows the visualization results of our COST and other SOTA methods on the proposed multi-modal small object tracking dataset VL-SOT230. We make the following observations: \textbf{1)} Small objects typically have weaker appearance and features compared to normal-sized objects. Therefore, algorithms (\eg, COST, VLT\_TT, and JointNLT) that utilize language information can often enhance the robustness of tracking systems. \textbf{2)} From Fig.~\ref{fig:qualitative_performance} and Tabs.~\ref{tab:Overall_results} and~\ref{tab:four_datasets}, we observe that the performance of current SOTA methods significantly drops on small object tracking datasets (\eg, VL-SOT230) compared to datasets focused on normal-sized objects (e.g., LaSOT, OTB99-L, and TNL2K). This indicates there is vast potential for improvement in the small object tracking field. In this work, we propose a multi-modal solution for small object tracking and pioneeringly suggest using language descriptions to enhance the performance of small object tracking.

\noindent\textbf{\emph{Remark 3:}} To intuitively verify the effectiveness of our method for small object tracking, we first highlight the challenges commonly associated with small objects, \eg, weak visual information and fast motion (see Figs.~\ref{fig:Example_Sequences} and~\ref{fig:qualitative_performance}). Then, we use feature distribution maps to show that our method facilitates the alignment of visual and language features in the feature space for small objects (see Fig.~\ref{fig:tSNE}). Finally, through visualized results (see Fig.~\ref{fig:qualitative_performance}), we empirically showcase the excellent performance of the proposed method in complex small-object scenarios (\eg, rapid motion, less effective visual
information, and brightness variations).

\begin{table*}[t]
\footnotesize
  \centering
    \caption{ Comparison of different multi-modal fusion manners on the LaSOT test set. The visual features are extracted from CNNs (\ie, AlexNet~\citep{alexnet2012nips}, and ResNet50~\citep{he2016deep}) and transformer (\ie,~\citep{vaswani2017attention}), respectively. For language features, GloVe~\citep{pennington2014glove} provides word embeddings as features, while BERT$_{\rm BASE}$~\citep{devlin2018bert} provides transformer features.}
  \setlength{\tabcolsep}{6mm}{
  \begin{tabular}{lllcc}
    \toprule 
    Tracking Model & Visual Features  & Language Features & AUC (\%) & $P_{norm}$ (\%)   \\
    \midrule

    \multirow{2}{*}{SiamRPN~\citep{li2018high}} & \multirow{2}{*}{AlexNet}  & GloVe & 43.6 &  52.8 \\
    &   & BERT$_{\rm BASE}$ & \textbf{46.0} & \textbf{54.6} \\
    \midrule
    
    \multirow{2}{*}{SiamRPN++~\citep{li2019siamrpn++}} & \multirow{2}{*}{ResNet50}  &  GloVe & 49.9 &  59.4 \\
     &  & BERT$_{\rm BASE}$ & \textbf{54.0} &  \textbf{63.6} \\
    \midrule

    \multirow{2}{*}{TransT~\citep{chen2021transformer}}  & \multirow{2}{*}{Transformer} & GloVe & 56.1 &  65.7 \\
      & & BERT$_{\rm BASE}$ & \textbf{56.8} &  \textbf{66.2} \\
    \bottomrule
  \end{tabular}
  }
  \label{tab:different_fusion_manners}
\end{table*}

\begin{table*}[t]
{
\footnotesize
  \centering
  \caption{Impact of training data using aligned training data setting (\ie, LaSOT, GOT-10k, COCO, and TrackingNet). The symbol $^{*}$ indicates using language descriptions or pseudo-language descriptions of the corresponding dataset.}
  
  \setlength{\tabcolsep}{4mm}{
  \begin{tabular}{lccccc}
   \toprule
    \multicolumn{1}{c}{\multirow{2}[1]{*}{Training Data}} &         
	\multicolumn{3}{c}{LaSOT} &         
	\multicolumn{2}{c}{LaSOT\_Ext}\\
    \cmidrule(r){2-4} \cmidrule(r){5-6}  \multicolumn{1}{c}{} &   AUC (\%) & $P$ (\%) & $P_{norm}$ (\%)  &   AUC (\%)  &  $P$ (\%)  \\
   
    \midrule
    LaSOT, GOT-10k, COCO, TrackingNet &  64.2 & 69.0  & 73.7  & 44.6 & 50.9  \\

    LaSOT$^{*}$, GOT-10k$^{*}$, COCO$^{*}$, TrackingNet$^{*}$&  \textbf{69.2} &  \textbf{74.6} & \textbf{79.3} & \textbf{52.0} & \textbf{59.3}  \\

   \bottomrule
  \end{tabular}
  }

  \label{tab:training_data_1}
}
\end{table*}

\begin{table*}[t]
{
\footnotesize
  \centering
  \caption{Impact of training data using complete training data setting (\ie, OTB99-L, TNL2K, WebUAV-3M). Here, the ``pretrained'' indicates the usage of the language-injected pretrained weights from the aligned training data setting.}
\begin{center}
  \setlength{\tabcolsep}{4.5mm}{
  \begin{tabular}{cccccccc}
   \toprule
 
    \multicolumn{1}{c}{\multirow{2}[1]{*}{Training Data}} &  \multicolumn{1}{c}{\multirow{2}[1]{*}{Pretrained}}&        
	\multicolumn{2}{c}{OTB99-L} &         
	\multicolumn{2}{c}{TNL2K} &         
	\multicolumn{2}{c}{WebUAV-3M} \\
 
    \cmidrule(r){3-4} \cmidrule(r){5-6} \cmidrule(r){7-8}  \multicolumn{1}{c}{}  &  &  AUC (\%) & $P$ (\%) &   AUC (\%)  &  $P$ (\%)  &   AUC (\%)  &  $P$ (\%) \\
   
    \midrule
    OTB99-L, TNL2K, WebUAV-3M &  \ding{55} &  72.1 &  90.5 & 50.5  & 51.9  & 45.0  & 62.1 \\

     OTB99-L, TNL2K, WebUAV-3M & \ding{51} & \textbf{77.3} & \textbf{94.5} & \textbf{57.5} & \textbf{58.6} & \textbf{49.8}  & \textbf{64.5}  \\

   \bottomrule
  \end{tabular}
  }
  \end{center}
  \label{tab:training_data_2}
  }
\end{table*}

\subsection{Further Discussions}
\label{sec:Further_Discussions}
\myPara{Homogeneous and Heterogeneous Multi-modal Fusion Manners.} We conduct experiments on the LaSOT dataset to show the impact of different multi-modal fusion manners. From Tab.~\ref{tab:different_fusion_manners}, we have the following observations. First, the transformer-based tracking model (\ie, TransT~\citep{chen2021transformer}) is superior to CNN-based tracking models (\ie, SiamRPN, SiamRPN++). Second, the transformer-based language model (\ie, BERT$_{\rm BASE}$~\citep{devlin2018bert}) provides better language features than the word embedding model (\ie, GloVe~\citep{pennington2014glove}). Third, the homogeneous multi-modal fusion manner (\ie, Transformer-Transformer) is superior to the heterogeneous multi-modal fusion manners (\ie, Transformer-Word Embedding, CNN-Transformer, and CNN-Word Embedding).

\begin{table*}[t]
	{
    \footnotesize
	\caption{Reliability of the two pseudo-language descriptions (\ie, major class and motion class, and major class). We compare them with two fine-grained language descriptions (\ie, initial concise and initial detailed descriptions) from~\citep{li2024dtvlt} to verify the quality of two pseudo-language descriptions on GOT-10k using AUC (\%)/$P$ (\%)/$P_{norm}$ (\%) scores.}
	\label{tab:relability_of_pseudo_language_descriptions}
	\begin{center}
		\setlength{\tabcolsep}{4.0mm}{
		\begin{tabular}{lccccccc}
			\toprule
				\diagbox [width=12em,trim=l] {{Test}}{~~Training} & {Major Class and Motion Class} & {Major Class} & {Initial Concise }  & {Initial Detailed} \\
			\midrule
			    
			   {Major Class and Motion Class} &  \textbf{79.0/68.0/90.1}  & 77.7/67.3/88.6 & 78.1/67.8/89.3   & 78.1/67.1/88.9 \\
				
		       {Major Class} &  \textbf{79.1/68.0/90.3}  & 78.0/66.9/89.0 & 77.7/66.8/88.7   & 77.9/66.9/88.7  \\
				
               {Initial Concise } &  78.5/67.8/89.5  & 77.6/66.9/88.5 & \textbf{78.6/68.0/89.9}   & 78.0/67.0/88.9 \\
				
			   {Initial Detailed} &  \textbf{79.1/68.1/90.2}  & 77.6/66.7/88.7 & 78.3/67.8/89.5   & 78.1/67.0/88.8  \\
               \midrule
               {Average} & \textbf{78.9}/\textbf{68.0}/\textbf{90.0}   & 77.7/67.0/88.7  & 78.2/67.6/89.4  & 78.0/67.0/88.9   \\
			 \bottomrule
		\end{tabular}
		}
	\end{center}
    }
\end{table*}

\begin{table}[t]
\footnotesize
  \centering
  \caption{ Efficiency comparison between COST and baseline tracker. }
  \setlength{\tabcolsep}{2.8mm}{
  \begin{tabular}{lccccc}
    \toprule 
    Inference Efficiency & TransT &  COST & $\Delta$  \\
    \midrule
    
    Visual Feature Extraction (s) & 0.0231 & 0.0231 & 0.0000 \\
    Language Feature Extraction (s) & 0.0000 & 0.0204 & -0.0204 \\
    Multi-modal Fusion (s) & 0.0000  & 0.0071 & -0.0071 \\
    Prediction (s) & 0.0005  & 0.0005 & 0.0000 \\
    \midrule
    Speed (FPS) & 42   & 36    &   -6 \\
    \bottomrule
  \end{tabular}
  }
\label{tab:time_analysis}
\end{table}

\myPara{Impact of Training Data.} We explore the impact of training data using aligned training data and complete training data. Following the recent SOTA VL/visual trackers~\citep{guo2022divert,wang2021transformer,chen2021transformer,yan2021learning,danelljan2020probabilistic,feng2021siamese}, we first train two trackers (without and with the proposed CVLF module) using four training sets (\ie,~LaSOT, GOT-10k, COCO, and TrackingNet) with bounding boxes and language annotations. Since GOT-10k and TrackingNet are without language annotations, we follow~\citep{guo2022divert} to provide a pseudo-language description for each video. Results are reported in Tab.~\ref{tab:training_data_1}. Our tracker with the proposed CVLF module (the fourth row) delivers significant performance improvements on the LaSOT test set and LaSOT\_Ext.

Furthermore, we fine-tune these two trackers using complete training data (\ie, OTB99-L, TNL2K, and WebUAV-3M). As shown in Tab.~\ref{tab:training_data_2}, the tracker with the language-injected pretrained weights (the fourth row) outperforms the tracker without pretraining (the third row). This is due to that the injected language information helps to enhance the tracking robustness under complex environments.

 \myPara{Reliability of Pseudo-language Descriptions.}  To further verify the reliability of the two pseudo-language descriptions (\ie, major class and motion class, and major class), we conduct comprehensive experiments on the GOT-10k dataset~\citep{huang2019got}, as it contains four types of language annotations (major class and motion class, major class, initial concise, and initial detailed descriptions). The two fine-grained language descriptions (initial concise and initial detailed descriptions) are from~\citep{li2024dtvlt}. We train the proposed COST on the GOT-10k training set (with 9,335 videos) and test it on the GOT-10k validation set (with 180 videos). 

Based on the results in Tab.~\ref{tab:relability_of_pseudo_language_descriptions}, we analyze the reliability and generalization capability of pseudo-language descriptions for VL tracking from the following perspectives: 

\textbf{1)} The descriptions of "major class and motion class" demonstrate significant superiority. Training with ``major class and motion class'' achieves the highest average performance (AUC: 78.9\%, $P$: 68.0\%), surpassing other language annotation types. This indicates that incorporating motion-related semantic cues (\eg, ``dog running'') enhances the alignment between visual dynamics and language descriptions, thereby improving tracking robustness. Notably, when tested on ``initial concise'' descriptions, this setup achieves competitive results (AUC: 78.5\%, $P$: 67.8\%), suggesting that motion-aware pseudo-languages generalize well even to fine-grained language annotations.

\textbf{2)} Overly fine-grained ``initial detailed'' descriptions have limitations. Despite achieving the highest CLIP score (see Tab.~\ref{tab:CLIP_Score}), ``initial detailed'' descriptions yield suboptimal average tracking performance (78.0\% in AUC, 67.0\% in $P$). This discrepancy arises because overly detailed annotations often introduce redundant or erroneous phrases (\eg, ``towards the left side of the image'' in Fig.~\ref{fig:GOT10k_language_annotations}) that mislead the tracker. For instance, when training with ``initial detailed'' descriptions and testing on other annotation types, performance drops by 0.5–1.2\% in AUC compared to ``major class and motion class'', highlighting the risks of noise in verbose language annotations.

\textbf{3)} Models trained on coarse-grained pseudo-languages (\eg, ``major class'') exhibit stable generalization. For example, training with "major class" achieves an average AUC of 77.7\%, which is only 1.2\% lower than ``major class and motion class''. This suggests that concise class-level annotations (\eg, ``dog'') provide sufficient semantic priors for tracking, albeit with less discriminative power compared to motion-enriched descriptions. However, testing on ``initial detailed'' annotations with models trained on coarse labels leads to performance degradation (78.1\% vs. 77.6\% for ``major class''), emphasizing the need for annotation consistency.

\textbf{4)} The best performance is achieved when training and testing use the same annotation type (except ``major class and motion class''). This indicates that alignment between training and testing annotation styles is critical. We hypothesize that the model trained on ``major class and motion class'' demonstrates superior performance across different annotation types due to its enhanced ability to handle annotation variability, as well as its capacity to benefit from precise or comprehensive language prompts during testing.

\textbf{5)} Overall, the results validate that pseudo-language descriptions based on ``major class and motion class'' strike an optimal balance between simplicity and reliability. As they avoid the noise inherent in detailed annotations while retaining sufficient discriminative semantics~\citep{guo2024divert}.

\myPara{ Efficiency Analysis.}
  To gain a more in-depth understanding of the proposed VL tracker COST, we conduct an efficiency analysis in Tab.~\ref{tab:time_analysis}. Referring to the baseline tracker TransT, we analyze the inference efficiency of four main processes: visual feature extraction, language feature extraction, multi-modal fusion, and prediction. The visual feature extraction and language feature extraction include the forward of the visual branch and linguistic branch, respectively. We also compute the time taken by the tracking head for target localization. Note that the CoA is only used during training and discarded during inference, thus introducing no additional computational overhead.

Our observations are as follows: \textbf{1)} COST and TransT employ the same visual feature extractor, thus requiring the same amount of time for visual feature extraction, \ie, 0.0231s. For the prediction process, COST and TransT take the same amount of time (0.0005s), even though the input feature dimensions are 441×256 (see Fig.~\ref{fig:TransVLT}) and 1024×256, respectively. This is because the tracking head has a very simple structure, making the prediction process highly efficient. \textbf{2)} Our COST uses BERT as the language feature extractor, adding an extra 0.0204s for language feature extraction. However, since we only extract language features once in the first frame, a good balance between efficiency (\ie, a real-time speed of 36 FPS) and performance is achieved. Note that our experiments have confirmed that language information significantly enhances small object tracking while incurring only a slight computational cost. \textbf{3)} After comparing the computation time of each process, we are surprised to find that the visual branch consumes the most time, \ie, 0.0231s. This is mainly because the computational complexity of the visual transformer is quadratic with respect to the token length~\citep {chen2021transformer}. Unfortunately, the length of visual tokens is relatively long (\eg, 1024), significantly increasing the computational overhead. Employing a linear-complexity network architecture~\citep{gu2023mamba} is a promising direction for reducing computational costs.

\begin{table}[t]
\footnotesize
  \centering
    \caption{ Comparison of total parameters, GPU memory usage, and inference speed of SOTA VL trackers on a single RTX 3090 GPU.}
  \setlength{\tabcolsep}{2.5mm}{
  \begin{tabular}{lcccc}
    \toprule 
    Method &   Parameters (M) & Memory (MB) & Speed (FPS) \\
    \midrule

    VLT\_TT~\citep{guo2022divert} & \textbf{100.9}  & 2,746 &  30  \\
    
    JointNLT~\citep{zhou2023joint} &  153.0  & 3,892 &  28  \\
    
    MMTrack~\citep{zheng2023towards} & 176.9  & \textbf{1,698} &  29 \\

    CiteTracker~\citep{li2023citetracker}  & 176.3  & 2,295 &  11  \\
    
    UVLTrack~\citep{ma2024unifying}  &  168.6 & 1,844 &  32 \\
    
    COST (Ours) & 146.5  & 2,888 &  \textbf{36}  \\
    \bottomrule
  \end{tabular}
  }
  \label{tab:paramaters_flops_speeds}
\end{table}

\subsection{ Limitations \texorpdfstring{$\&$}{\&} Failure Cases} 
 \myPara{Limitations.} Although our COST presents significant superiority in the newly proposed VL small object tracking dataset and five existing tracking benchmarks, our work still has the following two limitations:

\textbf{1)} This work applies unimodal transformer encoders to extract visual and language features, thus increasing the total parameters of our method. The parameters mainly come from the language model BERT$_{\rm BASE}$ (110M) and visual transformer (29.3M). Since we only use the basic transformer encoder layers in the visual-linguistic transformer, the parameters (7.2M) of the CVLF module only account for a small fraction ($5\%$) of the total parameters (146.5M). In Tab.~\ref{tab:paramaters_flops_speeds}, we compare the total parameters, GPU memory usage, and inference speed of six recent VL trackers. VLT\_TT has the smallest parameters due to its lightweight CNN-based fusion structure~\citep{guo2022divert}. MMTrack achieves the least GPU memory usage by eliminating complex proposal mechanisms, optimizing sequence quantization, and employing memory-efficient auto-regressive decoding~\citep{zheng2023towards}. As we extract language features only in the first frame during inference and use a simple and straightforward one-stage transformer fusion framework, our method achieves a favorable tracking speed in subsequent frames, with a real-time inference speed of 36 FPS. More advanced network architectures, such as Mamba~\citep{gu2023mamba}, can further reduce model parameters and memory usage while improving tracking speed. We leave it for future work.

\begin{figure*}[t]
\centering
\includegraphics[width=1.0\linewidth]{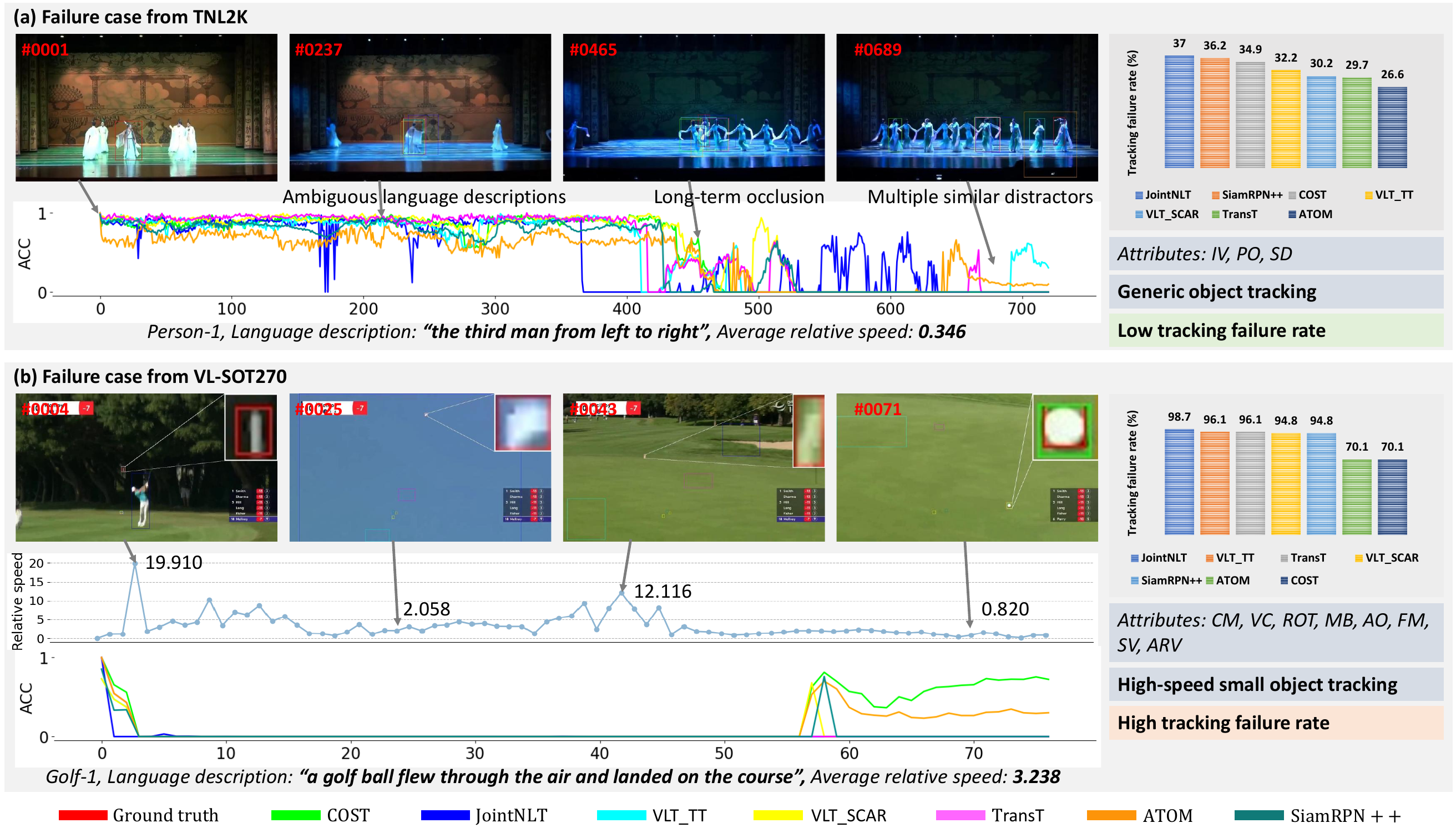}
  \caption{  Two failure cases. (a) On a challenging video sequence from the generic object tracking dataset TNL2K~\citep{wang2021towards}, COST, along with SOTA VL-based methods (JointNLT, VLT\_SCAR, VLT\_TT) and visual-based methods (TransT, ATOM, SiamRPN++), perform poorly when facing ambiguous language descriptions, long-term occlusion (over 220 frames), and multiple similar distractors. (b) On a video sequence from our proposed VL-SOT270 dataset for high-speed small object tracking, both COST and existing methods frequently lose the target object due to its rapid motion.}
   \label{fig:failure_case}
\end{figure*}

 \textbf{2)} We aim to learn VL representations using a simple and compact transformer-based framework (without extra tracking failure detection and correction modules), and therefore, our approach relies on accurate language annotations for matched video-language pairs and struggles to handle fast-moving targets. Although our method attempts to alleviate the issues of less effective visual information for small-sized objects and motion blur caused by high-speed movement from a semantic-enhanced perspective, there is still significant room for performance improvement.

\myPara{Failure Cases.} Fig.~\ref{fig:failure_case} presents two failure cases of our COST and SOTA trackers. We report the ACC score, relative speed of the target, and tracking failure rate~\citep{kristan2016novel} for an in-depth analysis. In Fig.~\ref{fig:failure_case}(a), on a challenging video sequence with ambiguous language descriptions, long-term occlusion (over 220 frames), and multiple similar distractors from the generic object tracking dataset TNL2K, our method struggles to achieve precise object localization due to the ambiguity between the language and visual modalities. Note that the above challenging factors may occur simultaneously, further increasing the possibility of tracking failure. Furthermore, we present another failure case from the proposed high-speed small object tracking dataset VL-SOT270 in Fig.~\ref{fig:failure_case}(b). In this video, the average relative speed of the target reaches 3.238, significantly surpassing the average relative speeds (0.543 and 0.700) of existing small object tracking datasets~\citep{liu2019aggregation,zhu2023tiny}. Not surprisingly, our COST and existing methods frequently lose the target in extreme high-speed small object scenarios. Comparing two cases from TNL2K and VL-SOT270, we found that the tracking failure rate of the algorithm increased significantly on the latter, indicating that high-speed small objects pose greater challenges to tracking algorithms. For instance, our COST exhibited tracking failure rates of 34.9\% and 70.1\% on two cases from TNL2K and VL-SOT270, respectively. 

Overall, from a novel language-enhanced perspective, we propose COST, a multi-modal tracker that demonstrates strong robustness when targets are visible. However, performance degradation may still occur in cases of target disappearance (\eg, full occlusion), visual-linguistic inconsistency (\eg, ambiguous language descriptions or severe deformation), or extreme high-speed motion. To address these issues, a promising solution is to incorporate a reliable memory mechanism leveraging multi-frame temporal information and motion dynamics~\citep{ravi2024sam,chunhui2023samsurvey}.

\section{Conclusion}
In this work, we propose COST, a new transformer-based one-stage multi-modal fusion framework for VL small object tracking. The core insight is to learn VL representations leveraging contrastive alignment and a simple and unified transformer architecture. To address the gap of lacking multi-modal small object tracking benchmarks, we take a step forward and propose VL-SOT500 dataset, which includes a large number of visual bounding box annotations and language descriptions. The dataset comprises two subsets, VL-SOT230 and VL-SOT270, specifically designed to advance language-enhanced generic and high-speed small object tracking. Extensive experiments showcase that our method achieves competitive or better performance compared with previous SOTAs on five VL tracking benchmarks and the newly proposed VL-SOT500. Our in-depth analysis yields numerous valuable observations and insights for VL tracking and beyond. In the future, we plan to apply our one-stage multi-modal fusion framework to more advanced tracking models and explore open vocabulary VL small object tracking.

\myPara{Acknowledgements.} This work was supported by the National Natural Science Foundation of China (No. 62471420), GuangDong Basic and Applied Basic Research Foundation (2025A1515012296), CCF-Tencent Rhino-Bird Open Research Fund, and the Major Project of Technology Innovation and Application Development of Chongqing (CSTB2023TIAD-STX0015).

\bibliographystyle{elsarticle-num} 
\bibliography{b_ref}

\clearpage
\clearpage
\bio{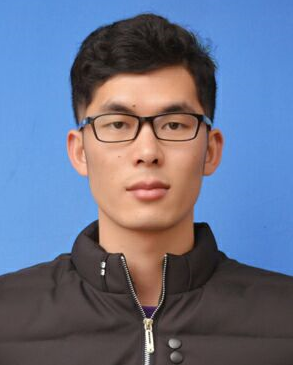}{\footnotesize \textbf{Chunhui Zhang} is currently pursuing the Ph.D. degree in Shanghai Jiao Tong University, China. He received his B.S. and M.S. degrees from Hunan University of Science and Technology, and the University of Chinese Academy of Sciences in 2016 and 2020, respectively. He also spent 2 years (2020-2022) at the Chinese University of Hong Kong, Shenzhen as a research associate. His major research interests are focused on machine learning, visual tracking, and multi-modal learning.}

\bio{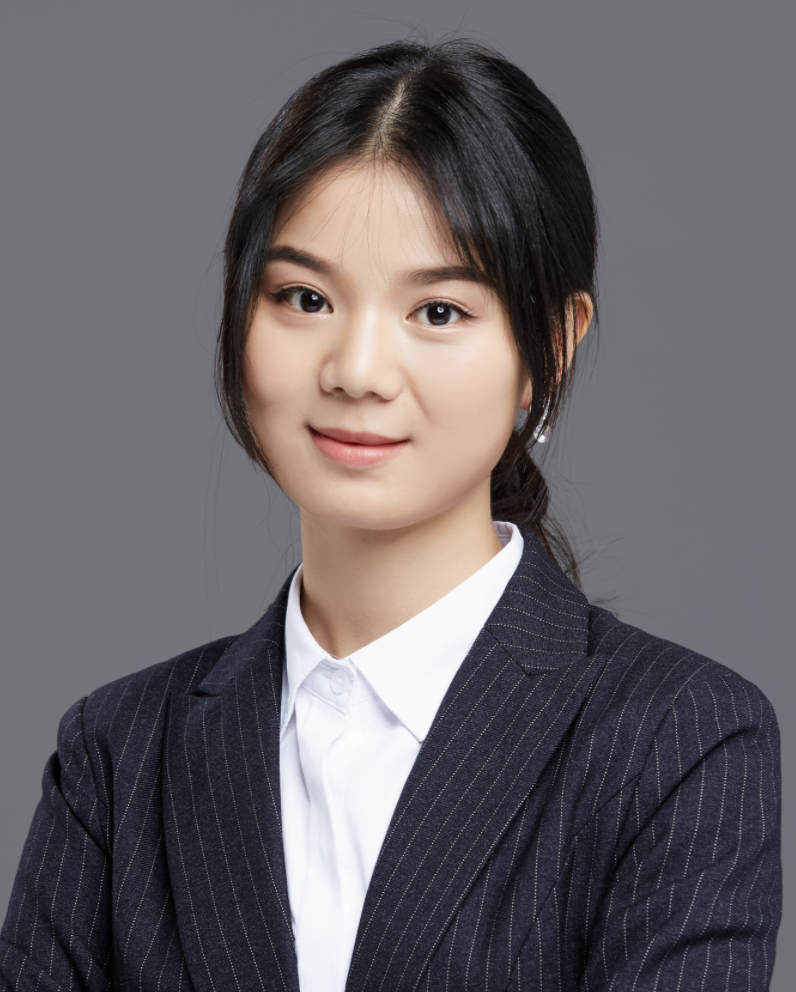}{\footnotesize \textbf{Li Liu} (Senior Member, IEEE) is an assistant professor at Hong Kong University of Science and Technology (Guangzhou), China. She received the Ph.D. degree in 2018 from Gipsa lab, University Grenoble Alpes, Grenoble, France. From September 2018 to September 2019, she was a postdoc researcher in the Department of Electrical, Computer, and Biomedical Engineering, Ryerson University, Toronto, Canada. Her current research interests include automatic audio-visual speech recognition, multi-modal fusion, Cued Speech development, lips/hand gesture recognition, and medical imaging. She has published in more than 30 top international peer-reviewed journals and conferences. She received the International Sephora Berribi Scholarship for Women Scientists and the French Phonetics Association (AFCP) Young Researcher Scholarship in 2017.}

\bio{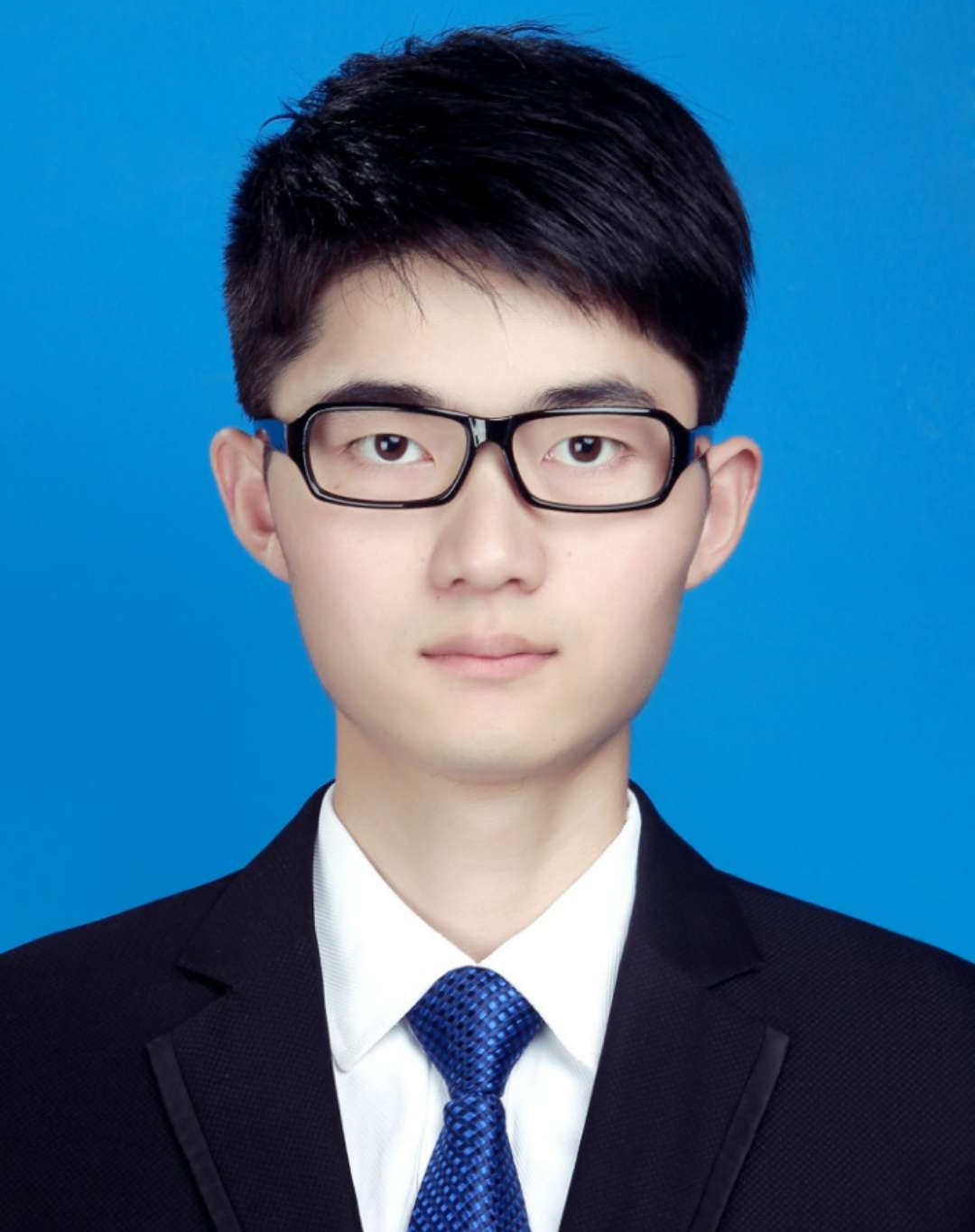}
{\footnotesize \textbf{Jialin Gao} has received the Ph.D. degree in Shanghai Jiao Tong University, China. He received the B.S. degree in electronic information engineering from the University of Electronic Science and Technology of China (UESTC), in 2016. His research interests include natural language
processing, action recognition, and temporal action localization.}

\bio{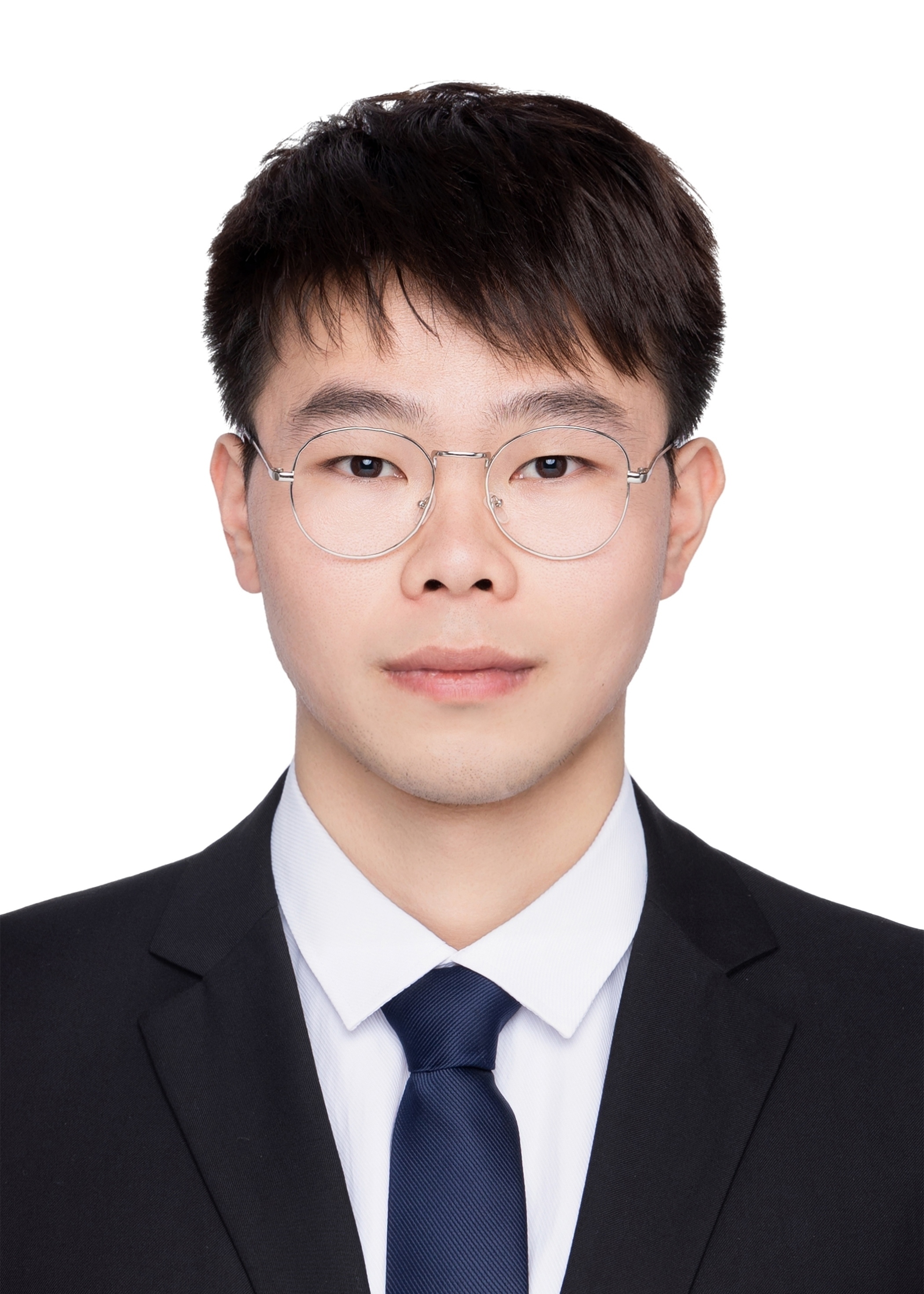}{\footnotesize \textbf{Xin Sun} has received the Ph.D. degree in Shanghai Jiao Tong University, China, in 2024. He received the B.S. degree in electronic engineering from Xi'an Jiao Tong University, China, in 2019. His research mainly focuses on video moment retrieval, referring image segmentation, and multi-modal learning.}

\bio{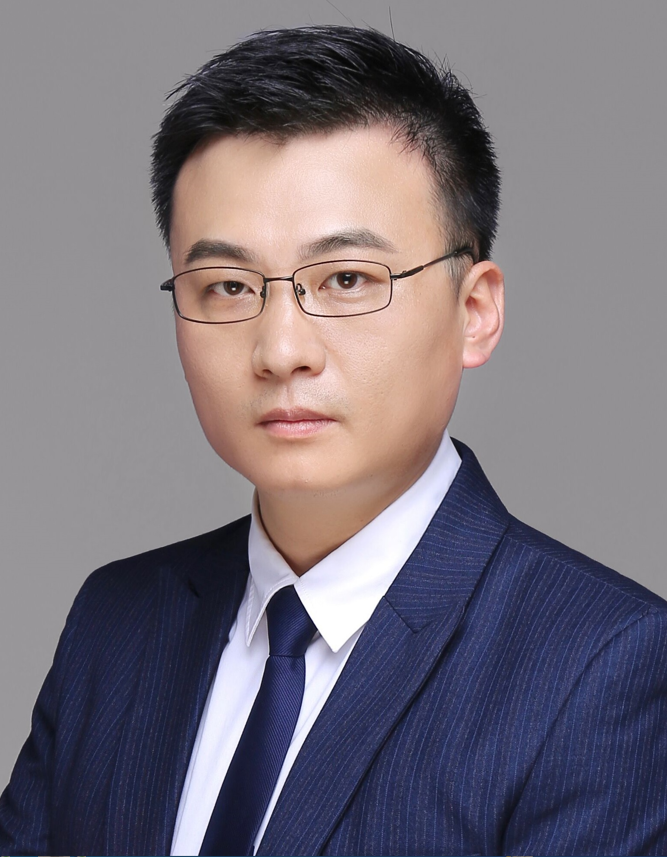}{\footnotesize \textbf{Hao Wen} received B.S. and Ph.D. degrees both in Electronic Engineering from the University of Science and Technology of China (USTC) in 2003 and 2008, respectively. His research mainly focuses on computer communication, quantum communication, computer vision, and AI others. He is currently the Officer of Strategic Technology of CloudWalk Technology, China.}

\bio{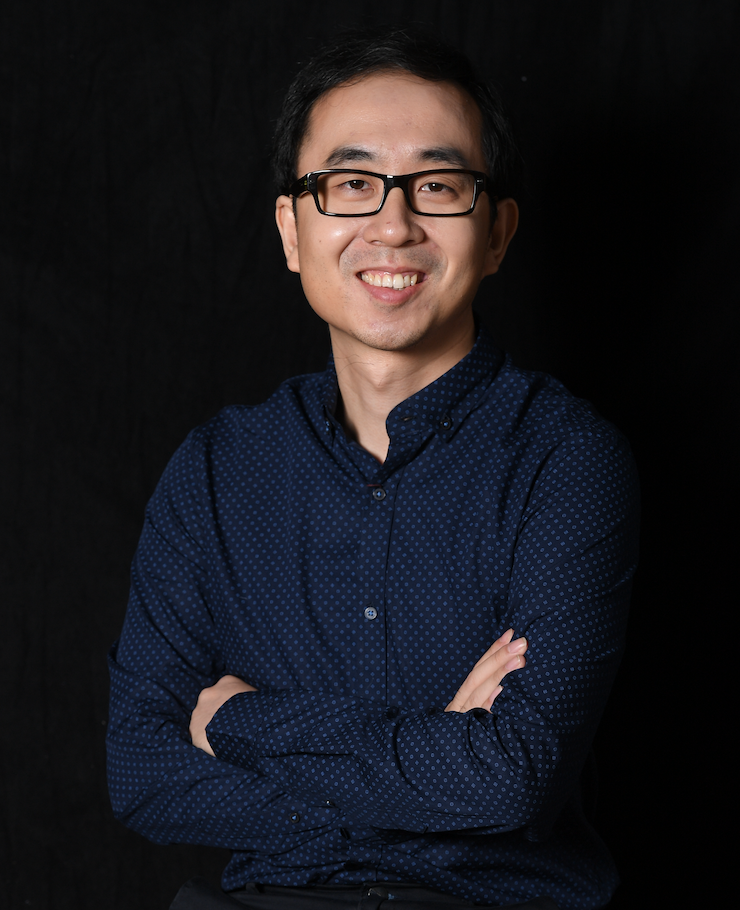}{\footnotesize \textbf{Xi Zhou} is currently the president of CloudWalk Technology, China. He is also a specially appointed Professor and a Ph.D. advisor of Shanghai Jiao Tong University, China. Before that, he was a Professor of Chongqing Institute of Green and Intelligent Technology, Chinese Academy of Sciences. He received the B.S. and M.S. degrees from the University of Science and Technology of China, in 2003 and 2006, and received the Ph.D. degree from the Department of Electrical and Computer Engineering of University of Illinois at Urbana-Champaign, in 2010. He has published more than 40 papers, and is the holder of more than 300 authorized patents. He has won 6 world-class computer vision contests, including the ImageNet Large Scale Visual Recognition Challenge (ILSVRC) in 2010.
}

\bio{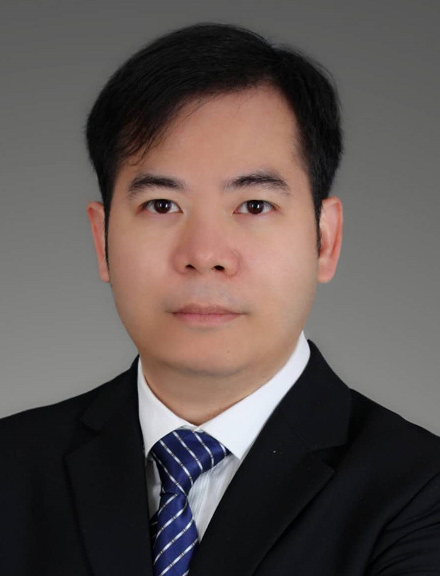}{\footnotesize \textbf{Shiming Ge} (M'13-SM'15) is a professor with the Institute of Information Engineering, Chinese Academy of Sciences. Prior to that, he was a senior researcher and project manager at Shanda Innovations and a researcher at Samsung Electronics and Nokia Research Center. He received B.S. and Ph.D. degrees both in Electronic Engineering from the University of Science and Technology of China (USTC) in 2003 and 2008, respectively. His research mainly focuses on computer vision, data analysis, machine learning, and AI security, especially trustworthy learning solutions towards scalable applications. He is a senior member of IEEE, CSIG, and CCF.}

\bio{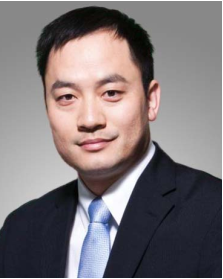}{\footnotesize \textbf{Yanfeng Wang} received the B.S. degree from PLA Information Engineering University, Zhengzhou, China, and the master’s and Ph.D. degrees in business management from Shanghai Jiao Tong University, Shanghai, China. He is currently the Deputy Dean of the School of Artificial Intelligence, Shanghai Jiao Tong University. He has long been focused on scientific research and innovation at the intersection of artificial intelligence with media and healthcare, as well as the translation of research outcomes. Dr. Wang’s achievements have been recognized with the First Prize of the Shanghai Science and Technology Progress Award (twice), the First Prize of the Shanghai Technological Invention Award (once), and the First Prize of the China Institute of Electronics Science and Technology Award (once). He has also been honored as a 2022 Shanghai Outstanding Academic Leader and a recipient of the 2015 Shanghai May Fourth Youth Medal.}

\end{document}